\documentclass{article}

\PassOptionsToPackage{numbers, compress}{natbib}
\usepackage[final]{neurips_2022}
\usepackage{microtype}
\usepackage{graphicx}
\usepackage{subfig}
\usepackage{booktabs} %
\usepackage{multirow}
\usepackage{dirtytalk}
\usepackage{adjustbox}
\usepackage{makecell}
\usepackage{hyperref}

\usepackage[utf8]{inputenc} %
\usepackage[T1]{fontenc}    %
\usepackage{hyperref}       %
\usepackage{url}            %
\usepackage{booktabs}       %
\usepackage{amsfonts}       %
\usepackage{nicefrac}       %
\usepackage{microtype}      %
\usepackage{xcolor}         %
\usepackage{enumitem}

\usepackage{amsmath}
\usepackage{amssymb}
\usepackage{mathtools}
\usepackage{amsthm}
\usepackage{xspace}
\usepackage[hide=true,setmargin=true,marginparwidth=0.75in]{marginalia}
\usepackage{csquotes}

\usepackage[capitalize,noabbrev]{cleveref}

\theoremstyle{plain}

\theoremstyle{definition}

\theoremstyle{remark}

\newcommand{\comment}[1]{}
\newcommand{\training}{training\xspace}
\newcommand{\bettertraining}{better training\xspace}
\newcommand{\capitalizedbettertraining}{Better Training\xspace}

\usepackage[textsize=tiny]{todonotes}

\title{Pruning’s Effect on Generalization Through the Lens of Training and Regularization}

\renewcommand*{\thefootnote}{\fnsymbol{footnote}}
\author{%
Tian Jin$^{1}$\thanks{Correspondence to: \texttt{tianjin@csail.mit.edu}, \texttt{mcarbin@csail.mit.edu}, \texttt{gkdz@google.com} . }\;  
Michael Carbin$^1$ Daniel M. Roy$^{2}$  Jonathan Frankle$^3$  Gintare Karolina Dziugaite$^4$ \\
\\
$^1$MIT \; $^2$University of Toronto, Vector Institute \; $^3$MosaicML \; $^4$Google Research, Brain Team \\
}

\begin{document}

\maketitle

\begin{abstract}
Practitioners frequently observe that pruning improves model generalization.
A long-standing hypothesis based on bias-variance trade-off attributes this generalization improvement to model size reduction.
However, recent studies on \fTBD{Do we need more description for what it is?}{over-parameterization} characterize a new model size regime, in which larger models achieve better generalization.
Pruning models in this over-parameterized regime leads to a contradiction -- while theory predicts that reducing model size harms generalization, pruning to a range of sparsities nonetheless improves it.
Motivated by this contradiction, we re-examine pruning’s effect on generalization empirically.

We show that size reduction cannot fully account for the generalization-improving effect of standard pruning algorithms.
Instead, we find that pruning leads to \bettertraining at specific sparsities, improving the training loss over the dense model.
We find that pruning also leads to additional regularization at other sparsities, reducing the accuracy degradation due to noisy examples over the dense model.
Pruning extends model training time and reduces model size.
These two factors improve \training and add regularization respectively.
We empirically demonstrate that both factors are essential to fully explaining pruning's impact on generalization.

\end{abstract}

\renewcommand*{\thefootnote}{\arabic{footnote}}
\setcounter{footnote}{0}

\section{Introduction}
\label{sec:intro}

%
%
%
%

	%
	%

\iffalse
\begin{figure*}[h!]
\includegraphics[width=\linewidth]{figs/intro-fig.png}
\caption{
}
\label{intro-fig}
\end{figure*}
\fi

Neural network pruning techniques remove unnecessary weights to reduce the memory and computational requirements of a model.
Practitioners can remove a large fraction (often 80-90\%) of weights without harming \emph{generalization}, measured by test error~\citep{NIPS1989_6c9882bb, DBLP:journals/corr/HanPTD15, frankle2018lottery}.
While recent pruning research \citep{DBLP:journals/corr/HanPTD15, frankle2018lottery, Renda2020Comparing, MLSYS2020_d2ddea18,Lebedev_2016_CVPR,molchanov2017pruning,NIPS2017_c5dc3e08,DBLP:journals/corr/abs-1711-05908,baykal2018datadependent,lee2018snip,wang2019eigen,DBLP:journals/corr/abs-2001-00218,Lee2020A,Liebenwein2020Provable,NEURIPS2020_d1ff1ec8,NEURIPS2021_e35d7a57,NEURIPS2021_a376033f,pmlr-v162-yu22f,DBLP:journals/corr/abs-2006-05467}
focuses on reducing model footprint, improving generalization has been a core design objective for earlier work on pruning \citep{NIPS1989_6c9882bb, 298572}; recent pruning literature also frequently notes that it improves generalization \citep{DBLP:journals/corr/HanPTD15, frankle2018lottery}.

How does pruning affect generalization?
An enduring hypothesis claims that pruning may benefit generalization by reducing \emph{model size}, defined as the number of weights in a model\fPROBLEM{TJ: reverted model capacity back to model size, because we don't want to define model capacity, and I haven't used it elsewhere for this reason.}.\footnote{
This hypothesis traces back to seminal work in pruning, such as Optimal Brain Surgeon \citep{298572}:
\say{without such weight elimination, overfitting problems and thus poor generalization will result.}
And again in pioneering work~\citep{DBLP:journals/corr/HanPTD15} applying pruning to deep neural network models:
\say{we believe this accuracy improvement is due to pruning finding the right capacity of the network and hence reducing overfitting.}
Because the latter hypothesis leaves the definition of network capacity unspecified, we examine an instantiation of it, measuring network capacity with model size.
}
We refer to this as the \emph{size-reduction hypothesis}.
\NA{However, algorithms for pruning neural networks and our understanding of the impact of model size on generalization have changed.}

First, pruning algorithms have grown increasingly complex.
Learning rate rewinding \citep{Renda2020Comparing}, a state-of-the-art algorithm prunes weights iteratively.
An iteration consists of \emph{weight removal}, where the algorithm removes a subset of remaining weights, followed by \emph{retraining}, where the algorithm continues to train the model, replaying the original learning rate schedule.
Across iterations, this retraining scheme effectively adopts a cyclic learning rate schedule \citep{7926641}, which may benefit generalization.

Second, emerging empirical and theoretical findings challenge our understanding of generalization for deep neural network models with an ever-increasing number of weights \citep{neyshabur2014search,dziugaite2017computing,nagarajan2019uniform,zhang2021understanding}.
In particular, the size-reduction hypothesis builds on the classical generalization theory on \emph{bias-variance trade-off}, which predicts that reducing model size improves the generalization of an overfitted model \citep{hastie_friedman_tisbshirani_2017}.
However, recent work \citep{Belkin15849, DBLP:journals/corr/abs-1912-02292} reveals that beyond this classical bias-variance trade-off regime lies an over-parameterized regime where models are large enough to achieve near-zero training error.
In this regime, bigger models often\fPROBLEM{GKD: added may, since there are algorithms that hit the maximal error in the second descent, and for which classical theory does apply. TJ: Can I change it to often? It appears so that at least for deep neural network models, bigger models often yield better generalization. GKD: a bit unclear what often means then. maybe we should say achieve better generalization on many standard benchmarks. TJ: I think often can mean that it occurs in the numerous citations linked here.} achieve better generalization \citep{Belkin15849, DBLP:journals/corr/abs-1912-02292, 10.1145/3065386, NEURIPS2019_093f65e0, 7298594}.
Practitioners routinely apply modern pruning algorithms to models in this over-parameterized regime. 
While our renewed understanding of generalization predicts that reducing the size of an over-parameterized model should harm generalization, pruning to a range of sparsities nevertheless improves it.
The size-reduction hypothesis is therefore inconsistent with our empirical observation about pruning.

\fPROBLEM{GKD: i think you should remove CIFAR-10 VGG-16 from the image title, it's wasting a lot of space, just add it in the figure caption.}
\begin{figure*}[t]
    \centering
    \subfloat[\label{fig:intro-fig}]{{
\includegraphics[width=4.5cm]{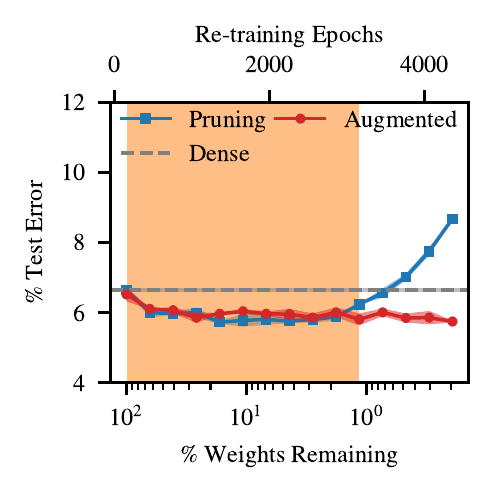}
    }}%
    \enskip
    \subfloat[\label{fig:proposed-explanation}
    ]{{
       	\includegraphics[width=8.4cm]{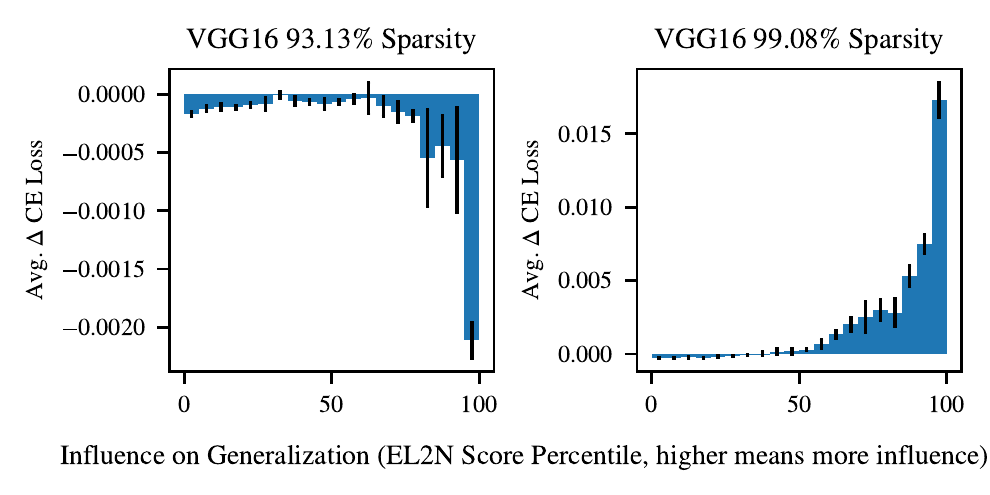} }
       }
    \caption{
(a) %
Retraining the VGG-16 model on CIFAR-10 exactly as done in pruning, but without removing any weights produces models with similar generalization as pruning to the shaded range of weights remaining.
We show the x-axis for pruning/augmented training below/above the plot respectively.
(b) We show the loss difference between the sparse/dense model produced by pruning/standard training respectively for training examples grouped by EL2N score percentiles.
Negative value indicates that the sparse model attains better loss.
Left:
 We find that pruning to a relatively low sparsity may improve generalization by decreasing training loss, with a particular decrease on influential examples.
Right: We find that pruning to a relatively high sparsity may also improve generalization by increasing training  loss, also with a particular increase on influential examples.
}
\label{fig:intro}%
\end{figure*}

\paragraph{A new direction.}
We first confirm that the size-reduction hypothesis does not fully explain pruning’s effect on generalization.
\Cref{fig:intro-fig} compares the generalization of a family of sparse models of different sizes generated by a state-of-the-art pruning algorithm \citep{Renda2020Comparing}, and the generalization of a family of models generated by the same algorithm, except modified to no longer remove weights.
This algorithm, which we refer to as the \emph{augmented training algorithm}, differs from standard training in two ways: (1) the number of training epochs is larger; (2) the learning rate schedule is cyclic.

The results show that for any sparse VGG-16 model generated by the pruning algorithm with more than 1\% of its original weights, its generalization is indistinguishable from that of the model generated by the augmented training algorithm that does not remove weights but trains the model for the same number of epochs and with the same learning rate schedule as the pruning algorithm.\footnote{In \Cref{appx:more-parameter-removal-ablation} we show this phenomenon for 4 additional benchmarks, and other pruning algorithms.}
For a range of sparsities, pruning's effect on generalization remains unchanged in the absence of weight removal, illustrating that the size-reduction hypothesis cannot fully explain pruning's effect on generalization.

\comment{
the generalization of the

We ablate weights removal from a  leaving retraining as the only action pruning performs to the model.
As the original learning rate schedule is replayed during each pruning iteration, this ablated algorithm trains the model with a cyclic learning rate schedule.
We plot the generalization of models we obtain, at the end of each pruning iteration, using pruning and this ablated algorithm in .
}

\paragraph{Our approach.} We instead develop an explanation for pruning's impact on generalization through an analysis of its effect on each training example's loss -- the difference in the example's training loss between the sparse model produced by pruning and the dense model produced by standard training.

We then interpret pruning's effect on each example in relation to the example's influence on generalization.
In particular, an example's influence on generalization is the absolute change to generalization due to leaving  this example out, which \citet{DBLP:journals/corr/abs-2107-07075} propose to approximate with the L2 distance between the predicted probabilities and one-hot labels early in training, called the EL2N (\textbf{E}rror \textbf{L2} \textbf{N}orm) score.
\citet{DBLP:journals/corr/abs-2107-07075} show that an example with a high EL2N score may cause a large weight update, thereby substantially changing the predictions and loss of other examples.
We thus refer to high EL2N examples as \emph{influential examples}.
Our analysis shows the following results.

\paragraph{Better training.} 
We prune a VGG-16 model on the CIFAR-10 dataset to the \emph{optimal sparsity}, the sparsity with the best validation error.
At this sparsity, we find that the pruned model achieves better training loss than the dense model.
In the left plot in \Cref{fig:proposed-explanation}, we present the change in training loss due to pruning for the training examples grouped by their EL2N score percentiles.
We observe that influential examples show the greatest training loss improvement.
Pruning may therefore lead to \emph{\bettertraining}, improving the training loss of the sparse model over the dense model, particularly on examples with a large influence on generalization.

\paragraph{Additional regularization.}
We prune the same VGG-16 model on CIFAR-10 dataset with the pruning algorithm configured to produce the sparsest model that still improves generalization over the dense model.
At this sparsity, we find that the pruned model achieves worse training loss than the dense model.
In the right plot in \Cref{fig:proposed-explanation}, we present the change in training loss due to pruning for the training examples grouped by their \NA{EL2N score percentiles}.
\fTBD{GKD: "grouped by example difficulty"? TJ: I think to maintain consistency with the previous paragraph we should use EL2N score percentiles. And we should probably use EL2N score percentiles for the previous paragraph, as this is the technically precise term. I wish to only replace high EL2N examples with difficult examples, as the mapping from EL2N to difficulty level is potentially cognitively demanding.} 
The results show that pruning may also improve generalization while increasing the training loss of the sparse model over the dense model, particularly on examples with a large influence on generalization.

While it is perhaps counter-intuitive that increasing training loss, particularly on the most influential examples %
may improve generalization, \NA{examining these pruning-affected examples} leads us to hypothesize that it is due to an observation that \citet{DBLP:journals/corr/abs-2107-07075} made: in image classification datasets, a small fraction of influential training examples have ambiguous or erroneous labels, training on which impairs generalization. Therefore, it is instead better for the model to not fit them.\fTBD{MC: this narrative here will prompt readers}

We examine this hypothesis by introducing random label noise into the training dataset.
We find that on such datasets, pruning may improve generalization by increasing the training loss on the noisy examples, while still preserving the training loss of examples from the original training dataset.
The result is an overall generalization improvement for the sparse model.
\NA{Notably, reducing model width has a similar effect on training loss and generalization, suggesting a reduced number of weights in the model as the underlying cause for generalization improvement in the presence of noisy examples.}
\fTBD{Do we really want to surface this results early on?}
We refer to the effect of removing weights as \emph{additional regularization}, which reduces the accuracy degradation due to noisy examples over the original dense model.

\paragraph{Implications.} It has been a long-standing hypothesis that pruning improves generalization by reducing the number of weights (via the Minimum Description Length principle  \citep{grunwald2007minimum} or Vapnik–Chervonenkis theory \citep{vapnik1999nature}).
Our empirical observations suggest that the theory and practice of pruning should instead focus on the two effects of modern pruning algorithms: \bettertraining and additional regularization, both of which are indispensable to explaining pruning's benefits to generalization fully.
Our results also suggest that quantifying pruning's heterogenous effect across training examples is key to understanding pruning's influence on generalization.

\comment{
\paragraph{Pruning's Practical Value.}
We show that to attain the best generalization, the optimal effect pruning exerts on the model fit is neither always to increase or reduce it, but rather depends on the portion of noisy examples in the dataset.
Our finding suggests that the practical value of pruning lies in the range of sparse models produced by the pruning algorithm, each corresponding to a unique balance between \bettertraining and additional regularization, indexed by the sparsity level.
For a given dataset, practitioners can select the optimal balance between these two effects, by picking the sparsity level with the best validation performance.
}

\paragraph{Contributions.}
\fTBD{GKD: can we shift the bullet points to the left?}
We present the following contributions:

\begin{enumerate}[leftmargin=1.5em,itemsep=0.5ex]
\item We find that pruning may lead to \emph{\bettertraining}, improving the training loss, as well as the generalization over the dense model at optimal sparsity.

\item We find that pruning may also lead to \emph{additional regularization}, reducing the accuracy degradation due to noisy examples, and improving the generalization over the dense model at optimal sparsity.

\item We demonstrate that our deconstruction of pruning’s effect into \training and regularization cannot be further simplified — alternative explanations such as extended training time and reduced model size only partially explain the effects of pruning on generalization.
\end{enumerate}

\section{Preliminaries}
\label{sec:prelim}

We study \emph{Iterative Magnitude Pruning (IMP) with learning-rate rewinding} \citep{Renda2020Comparing}.
This algorithm consists of four steps:
(1) Train a neural network to completion.
(2) Remove a specified fraction of the smallest-magnitude remaining weights.
(3) Reset the learning rate to its value in an earlier epoch $t$ and
(4) Repeat steps 1-3 iteratively until the model reaches the desired overall sparsity.

\textbf{Terminology.}
\emph{Dense models} refer to models without any removed weights.
\emph{Sparse models} refer to models with removed weights as a result of pruning.
The \emph{optimally sparse model} is the model pruned to the sparsity at which it achieves the best validation error. 
The \emph{optimal sparsity} is the sparsity of the optimally sparse model.
\emph{Generalization} refers to the classification error on the test set.

\textbf{Experimental methodology.}
We use standard architectures: LeNet \citep{lenet}, VGG-16 \citep{Simonyan15}, ResNet-20, ResNet-32 and ResNet-50 \citep{7780459}, and train on benchmarks (MNIST, CIFAR-10, CIFAR-100, ImageNet) using standard hyperparameter settings and standard cross-entropy loss function \citep{frankle2018lottery, DBLP:journals/corr/abs-1912-05671, DBLP:journals/corr/abs-2002-07376}. 
Following \citet{frankle2018lottery, DBLP:journals/corr/abs-1912-05671}, we set the $t$ in IMP to $t=0$ for MNIST-LeNet benchmark and $t=10$ for the others. \cref{sec:expdetails} shows further details.

We report all results by running the same experiment 3 times with distinct random seeds.
We conclude that any real-valued results of two experiments differ if and only if the mean difference between three independent runs of respective experiments is at least one standard deviation away from zero, otherwise we say that the results \emph{match}.
We use PyTorch \citep{NEURIPS2019_9015} on TPUs with OpenLTH library \citep{frankle2018lottery}.

\textbf{Limitations.} Our work is empirical in nature. Though we validate each of our claims with extensive experimental results on benchmarks with different architectures and datasets, we recognize that our claims may still not generalize due to the empirical nature of our study.

\section{Generalization Improvement from \capitalizedbettertraining}
\label{subsec:aggregate-effect}

Not all training examples have the same effect on training and generalization \citep{toneva2018an,DBLP:journals/corr/abs-2107-07075, 5360534, NIPS2016_9d268236, DBLP:journals/corr/abs-1911-05248}.
In this section, we measure pruning's effect on each training example by examining the difference in its training loss between the dense model produced by standard training and the sparse model produced by pruning.
Our results show that pruning most significantly affects training examples that are most influential to model generalization.

\textbf{Method.}
\citet{DBLP:journals/corr/abs-2107-07075} propose to estimate an example's influence on model generalization with the EL2N (\textbf{E}rror \textbf{L2} \textbf{N}orm) score -- the L2 distance between the predicted probability and the one hot label of the example early in training.
Details for computing the EL2N score is available in \Cref{sec:calc-el2n}.
They show that an example with a high EL2N score may cause a large weight update, thereby substantially changing the predictions and loss of other examples.
\citet{DBLP:journals/corr/abs-2107-07075} rely on this relationship between EL2N score and generalization to remove examples with little influence on generalization from the dataset to accelerate training without affecting model generalization.
We similarly use the EL2N score to assess the impact of pruning on generalization.

To measure the difference in training loss on examples due to pruning, we partition the training set into $M$ subgroups, each with a different range of EL2N score percentiles.
We then compute the average training loss difference on examples in each subgroup.
We compute this difference by subtracting the training loss of the dense model produced by standard training from that of the sparse model produced by pruning.
A negative value indicates that the sparse model has better training loss than the dense model.
We pick $M=20$ because it is the largest value of $M$ that enables us to clearly present the per-subgroup training loss difference.

We measure training loss at two sparsities of interest: the sparsity that achieves the best generalization, and the highest sparsity that still attains better generalization than the dense model.
In \Cref{appx:aggregate-effect}, we present pruning's effect on subgroup training loss at a wider range of sparsities.

\begin{figure*}[h]
\centering
\subfloat[At sparsity achieving the best generalization]{%
   \includegraphics[width=\linewidth]{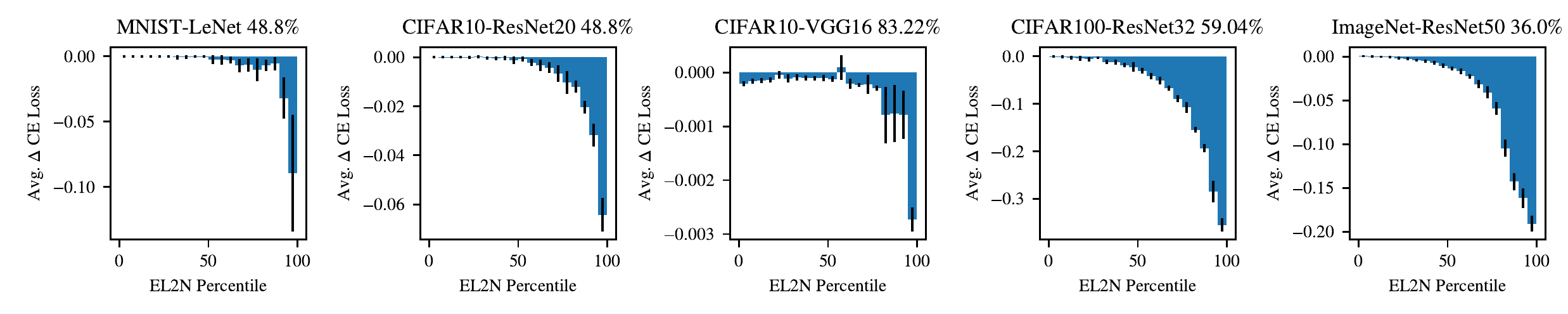}
   \label{fig:max-generalization}
}

\subfloat[At highest sparsity with generalization exceeding that of the dense counterpart]{%
   \includegraphics[width=\linewidth]{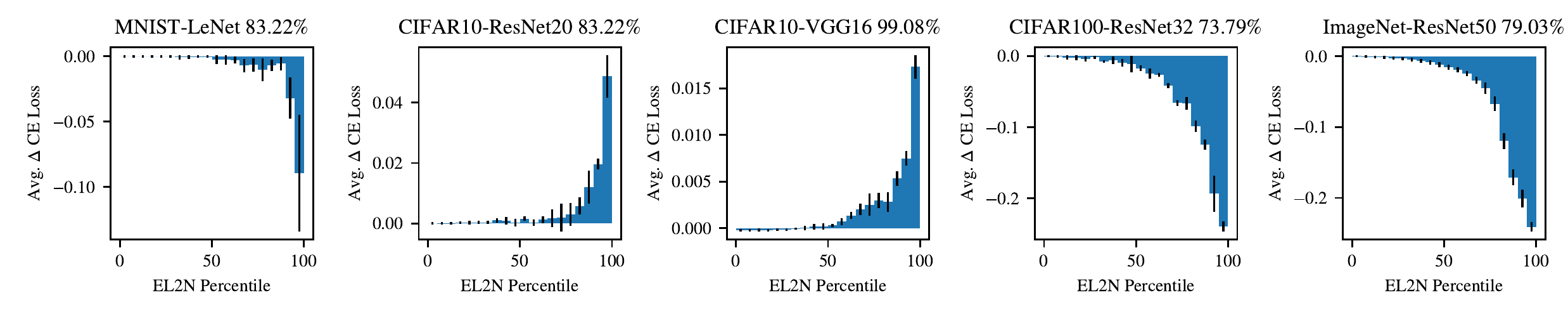}
   \label{fig:max-sparsity}
}
\caption{Difference between the average training loss of the dense model produced by standard training and the sparse model produced by pruning on example subgroups with distinct EL2N percentile range.
Negative values indicate that the sparse model has better loss than the dense model.
}
    \label{fig:pruning-affect-examples-with-high-el2n}%
\end{figure*}

\textbf{Results.}
\Cref{fig:max-generalization} presents pruning's effect on subgroup training loss.
Our results show that at the sparsity that achieves the best generalization, for all subgroups, the average training loss of the pruned model either matches or improves over the dense model, indicating \bettertraining.

\Cref{fig:max-generalization} also shows that pruning's effect on training example subgroups is nonuniform.
Pruning most significantly affects the most influential examples -- the ones with the highest range of EL2N scores.
Specifically, the average magnitude of pruning's effect on examples with the top 20\% EL2N scores is 3.5-71x that on examples with the bottom 80\% EL2N scores.
Pruning to the said sparsity thus improves generalization while improving the training loss of example subgroups nonuniformly, with an emphasis on influential examples.
In \cref{sec:pruning-affected-examples}, we validate the effect of pruning-improved examples on generalization:
excluding 20\% of pruning-improved examples hurts generalization more than excluding a random subset of the same size.\fTBD{MC: seems like this should be mainline rather than in the appendix}

\textbf{Conclusion.}
\NA{On standard datasets, pruning to the sparsity that achieves the best generalization leads to better training -- the training loss of most example subgroups improves over the dense model.
The subgroup with the most influence on generalization sees the largest training loss improvement.}

\section{Generalization Improvement from Additional Regularization}
\label{sec:dist-dependence}

Pruning to the highest sparsity that improves generalization (\Cref{fig:max-sparsity}) displays a similar effect as pruning to the sparsity that attains the best generalization (\Cref{fig:max-generalization}) -- generalization improves with an overall improved training loss over the dense model, except for CIFAR-10, where pruning improves generalization while worsening training loss on most subgroups of examples.
Specifically, for ResNet20 on CIFAR-10 at 83.22\% sparsity (\Cref{fig:max-sparsity}, second from left) and VGG-16 on CIFAR-10 at 99.08\% sparsity (\Cref{fig:max-sparsity}, third from left), the training loss of the sparse model is worse than the dense model.
While pruning always increases training loss when it removes a large enough fraction of weights, the accompanied generalization improvement does not always occur in general.
The second and third plots of \Cref{fig:max-sparsity} show that the loss increase is especially pronounced on particular subgroups of influential examples.
Examining these pruning-affected training examples in \Cref{appx:visualizing-pruning-affected-examples}  leads us to hypothesize that the generalization improvement is due to an observation \citet{DBLP:journals/corr/abs-2107-07075} made: although influential examples are predominately beneficial for generalization, there exist a small fraction of noisy examples, such as ones with ambiguous or erroneous labels, that exert a significant influence on generalization, albeit in a harmful way.\footnote{We present leave-subgroup-out retraining experiments for these examples in \Cref{appx:effect-of-avoided-examples}.}

To precisely characterize pruning's effect on noisy examples, we inject random label noise into the training dataset.
Doing so enables a comparison of pruning's effect on noisy data and original data.
%

%
%
%
%

\iffalse
We describe pruning's effect as a function of sparsity level -- at distinct sparsity levels, pruning increases or decreases model fit to certain examples, especially those with high extent of influence on generalization.
Applied to a particular dataset, when pruning improves generalization, its effect is to either improve model fit to examples beneficial to generalization or by avoiding fitting examples harmful to generalization.
Pruning’s effect must therefore depend on the portion of noisy examples in the data distribution.
%
In this section, we demonstrate the implication of this distribution dependence property, and reflect on the value of pruning to improving model generalization.
\fi

\begin{figure*}[t!]
\includegraphics[width=\linewidth]{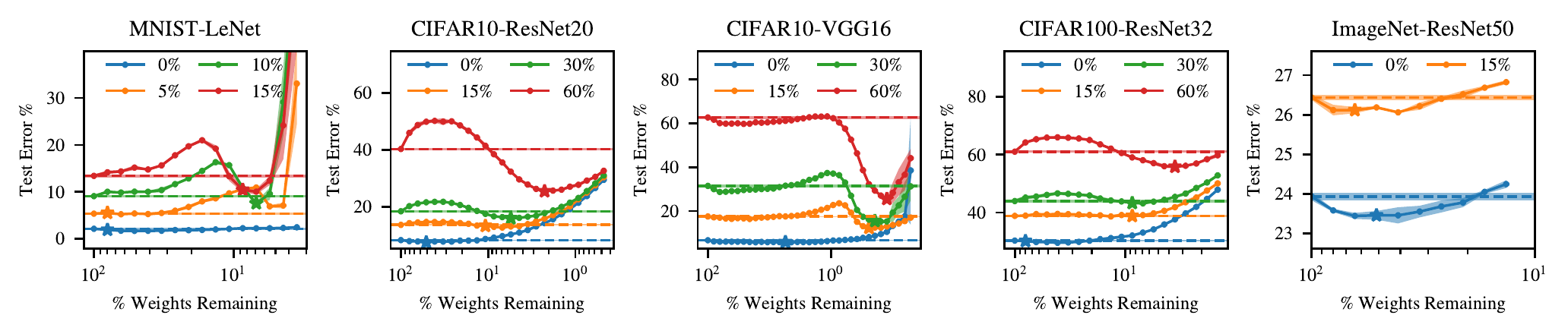}
\caption{Pruning improves generalization over the dense model in the presence of random label noise. 
We show the generalization of the sparse models as a function of the \% weights remaining.
Horizontal dashed lines indicate the baseline generalization of the dense models.
We mark the optimal sparsity with a star.
Legends show the fraction of the training examples with random label noise.}%
\label{fig:pruning-effect-test-error}%
\end{figure*}

\begin{figure*}[h!]
\includegraphics[width=\linewidth]{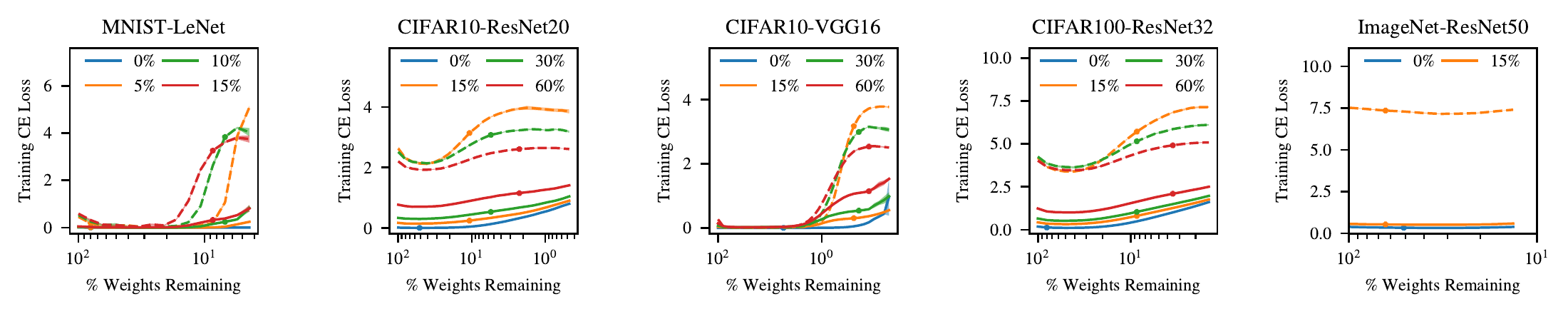}
\caption{Pruning to the optimal sparsity increases training loss over the dense model in the presence of random label noise.
Noisy examples see a particular large loss increase.
We show the training loss on the original and noisy data with dashed and solid lines respectively, as functions of \% weights remaining.
We mark the optimal sparsity with a star.
Legends show the random label noise level.
}%
\label{fig:pruning-effect-on-memorization}
\end{figure*}

\begin{figure*}[t!]
\includegraphics[width=\linewidth]{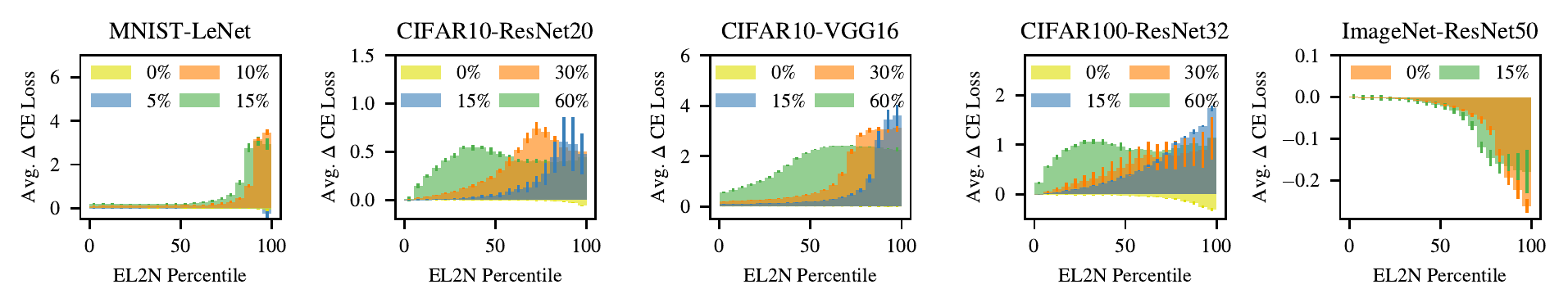}
\caption{
In the presence of random label noise, pruning to the optimal sparsity often increases training loss (except for ResNet50) over the dense model, especially to highly influential examples.
For example subgroups with distinct EL2N percentile ranges, we show the difference in average training loss between dense and optimally sparse models.
Legend indicates label noise level.
We do not show LeNet with 5\% noise because the dense model achieves the best generalization.
For ResNet20 and ResNet32 with 60\% noise, the training loss for high EL2N examples are already very high for dense models to begin with, leaving little room for their loss to increase further by pruning.}%
\label{fig:pruning-el2n-loss-change-noisy}%
\end{figure*}

\textbf{Method.}
We inject $p \%$ random label noise by selecting $p\%$ examples uniformly at random and changing the label of each example to one of the other labels in the dataset sampled uniformly at random. We refer to the data that is not affected as original data.
We sample and fix the random labels before each independent run of an experiment.
We report the following results for models trained on datasets with and without random label noise:
(1) pruning's impact on generalization versus sparsity (\Cref{fig:pruning-effect-test-error});
(2) pruning's impact on training loss for the original data and random label data versus sparsity (\Cref{fig:pruning-effect-on-memorization});\footnote{Numerical results available in \cref{tbl:pruning-impact-on-noisy-clean-memorization} in \cref{appx:raw-data}}
(3) pruning's impact on training loss for subgroups of examples, each with a different range of EL2N scores (\Cref{fig:pruning-el2n-loss-change-noisy}).

\textbf{Generalization results.}
\Cref{fig:pruning-effect-test-error} shows that, in the presence of random label noise, the generalization of the optimally sparse models is better than that of the dense models on 11 out of 13 benchmarks and matches that of the dense models on the rest of the benchmarks.
Our results provide evidence that pruning to the optimal sparsity improves generalization in the presence of random label noise.

\NA{The extent of pruning's generalization improvement at the optimal sparsity grows as the portion of label noise increases.}\fTBD{I think this is an important observation, but can potentially be further distilled so that it better serves our conclusion.}
For example, for VGG-16 on CIFAR-10 benchmark, pruning to the optimal sparsity improves generalization by 5.2\%, 16.1\% and 34.6\% on 15\%, 30\% and 60\% label noise.

\textbf{Original versus noisy data.}
\Cref{fig:pruning-effect-on-memorization} shows that pruning initially reduces the average training loss on noisy examples: for instance, training loss on noisy examples is slightly lower for the sparse ResNet20 models on CIFAR-10 with many weights remaining (i.e., more than 10\% ) than for the dense model.
Since noisy examples are influential examples \citep{DBLP:journals/corr/abs-2107-07075}, these results indicate \bettertraining, which improves model training loss with a particular emphasis on influential examples.

As the fraction of weights remaining drops further, the loss of the sparse model on noisy data grows.
For instance, the training loss on noisy examples increases for the sparse ResNet20 models on CIFAR-10 with few weights remaining (i.e., less than 10\%).
Moreover, the gap between the average training loss of the sparse model on original versus noisy examples widens beyond the dense model, indicating that pruning leads to additional regularization.
In particular, for a range of sparsities where generalization improves over the dense model, the sparse model fits original examples significantly better than those with randomized labels.
Taking the dense VGG-16 model trained on the CIFAR-10 dataset with 15\% -- 60\% random label noise as an example, we observe that this model incurs 0.03 -- 0.13 higher training loss on the partition with injected random label noise than on the partition without.
Pruning to the optimal sparsity widens this training loss gap to 1.38 -- 2.97.
Furthermore, the gap between the average training loss of the sparse models on original versus noisy examples eventually diminishes as the sparsity increases even further.

\textbf{Difference in loss by subgroups.}
\Cref{fig:pruning-el2n-loss-change-noisy} shows the difference between the loss of the dense and optimally sparse model on subgroups of examples with different EL2N scores, but this time in the presence of random label noise.
The results show that pruning to the optimal sparsity often increases the loss of influential examples, contrary to what \cref{fig:pruning-affect-examples-with-high-el2n} shows when training on original data only where pruning decreases the training loss of influential examples.

As random label examples are influential and harm generalization \citep{DBLP:journals/corr/abs-2107-07075}, increasing the training loss on influential examples may therefore improve generalization.
\NA{In \Cref{sec:make-dense-model-ignore-noisy-data}, we validate that excluding examples misclassified by the optimally sparse models improves generalization similar to pruning.}\fTBD{The motivation for this validation experiment is unclear.}

\textbf{Conclusion.}
We study pruning's effects on training and regularization by connecting them to noisy examples. 
By introducing random label noise, we show the two effects of pruning more clearly.
At relatively low sparsities, we observe \bettertraining because the training loss of sparse models is lower than the dense model, especially on influential examples highly influential to generalization.
As sparsity increases, regularization effects dominate, because the sparse models see a larger gap between training loss on the noisy versus original data than the dense model.
We note that across all sparsities, pruning has a highly nonuniform effect across example subgroups with different influences on generalization (EL2N scores) on both standard datasets and datasets with random label noise.

Crucially, the effect of pruning at the optimal sparsity -- be it \bettertraining or additional regularization -- depends on the noise levels in the dataset.
On datasets with few noisy examples, \Cref{fig:pruning-affect-examples-with-high-el2n} shows that at the optimal sparsity, \bettertraining is the key to good generalization, because training loss reduces.
On datasets with injected random label noise, \Cref{fig:pruning-el2n-loss-change-noisy} shows that the optimal sparsity is the one at which pruning-induced regularization effects increase noisy data loss.

\section{Isolating the Effects of Extending Training Time and Reducing Model Size}
\label{sec:pruning-inspired-dense-training}

We identified two effects of pruning on generalization: \bettertraining and additional regularization.
They intersect with two components of the pruning algorithm: extended training time and model size reduction, respectively.
In this section, we study these two effects in isolation by extending dense model training time and reducing the dense model size by scaling down its width.\footnote{To isolate other effects of pruning, we also evaluate the contribution of design choices of pruning algorithms, such as the heuristic selecting which weights to prune, to generalization via ablation studies in \Cref{sec:ablation}}.
We confirm that both effects of pruning are necessary to fully explain pruning's impact on generalization.
Thus our deconstruction of pruning's effect in terms of training and regularization cannot be further simplified.

\fTBD{MC: such as what? Or is this specifically vague so we can incorporate other things later}

\subsection{Extended Training Time}
\label{subsec:edt}

What happens to generalization if the weight removal part is removed from pruning, and only the retraining part remains?
To answer, we train dense models exactly as done in the pruning algorithm, for the same number of gradient steps and learning rate schedule, but without removing weights.

\textbf{Method.}
We study the effect of training dense models for extra gradient steps on generalization, without removing any weights.
When training the dense models, we use the same learning rate schedule as pruning, replaying the original learning rate schedule designed for the dense model multiple times, effectively adopting a cyclic learning rate schedule.
We refer to this training algorithm as \emph{extended dense training}.
We report and compare the generalization of models that pruning and extended dense training algorithm produce.

\begin{figure*}[h!]
\includegraphics[width=\linewidth]{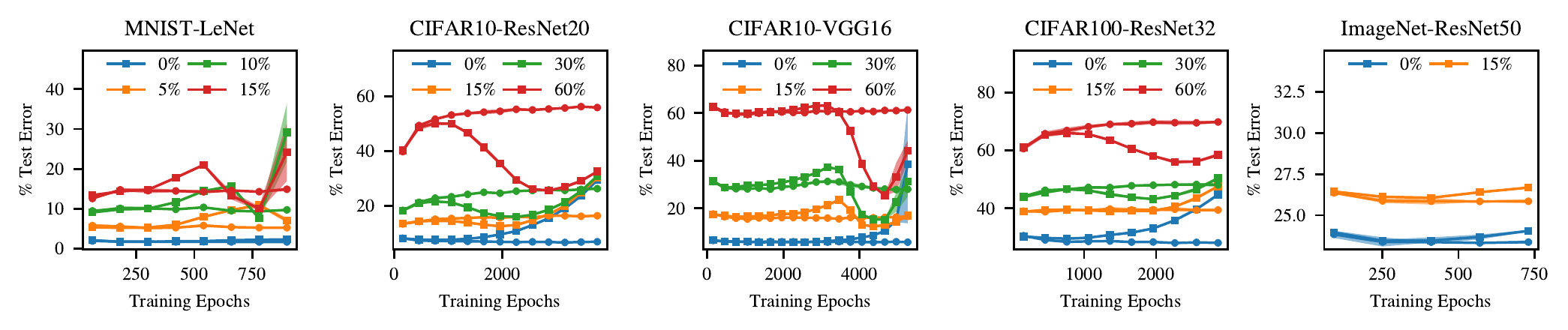}
\caption{
Extended training time does not explain the full extent of pruning's benefits to generalization.
We show the test errors of the dense and sparse models with circle and square markers respectively as a function of the training epochs.
Legends show random label noise level.}
\label{fig:pruning-vs-cyclic-lr}%
\end{figure*}

\textbf{Results.}
\Cref{fig:pruning-vs-cyclic-lr} shows test errors of models that extended dense training and pruning produce as a function of training epochs.\footnote{Numerical results available in \Cref{tbl:pruning-vs-dense-training} in \Cref{appx:raw-data}.}
The two algorithms produce models with similar generalization on standard datasets without random label noise, where pruning improves \training at optimal sparsity, as shown in \Cref{subsec:aggregate-effect}.
Indeed, in \Cref{appx:aggregate-effect}, we show that longer training time similarly leads to \bettertraining.
However, on benchmarks with random label noise, extended dense training underperforms pruning on 9 out of the 13 benchmarks.
On LeNet, ResNet20, VGG-16, ResNet32 and ResNet50 benchmarks with noise, extended dense training achieves test errors that are worse than pruning by 0  (matching) to 2.6\%, 0.4 to 14.5\%, 3.6 to 34.2\%, 0 (matching) to 4.8\% and -0.3\%, respectively.

\textbf{Conclusion.}
On standard datasets without random label noise, the extended dense training algorithm produces models with generalization that matches or exceeds the optimally sparse model.
However, with random label noise, the extended dense training algorithm can no longer produce models that match the generalization of the optimally sparse model.

\subsection{Size Reduction}
\label{sec:size-reduction}
What happens to generalization if the retraining part is removed from pruning, and only the weight removal part remains?
To answer, we compare pruning with model width down-scaling, since both techniques reduce the number of weights in a model.

At optimal sparsity, we show that size reduction is not necessary to \NA{replicate generalization improvements with dense models} in \cref{sec:intro};
in this section, we show that size reduction is also
not sufficient to \NA{replicate generalization improvement with dense models}.

\textbf{Method.}
We reduce the model size by down-scaling the width of a dense model: we train a sequence of dense models using the standard number of training epochs, where the next model in the sequence has 80\% of the width of the model preceding it.
This width scaling ratio mimics our sparsity schedule, where the pruned model always has 80\% of the remaining weights in the previous iteration.
We then compare the generalization of models that pruning and width down-scaling produce.

\begin{figure*}[h!]
\includegraphics[width=\linewidth]{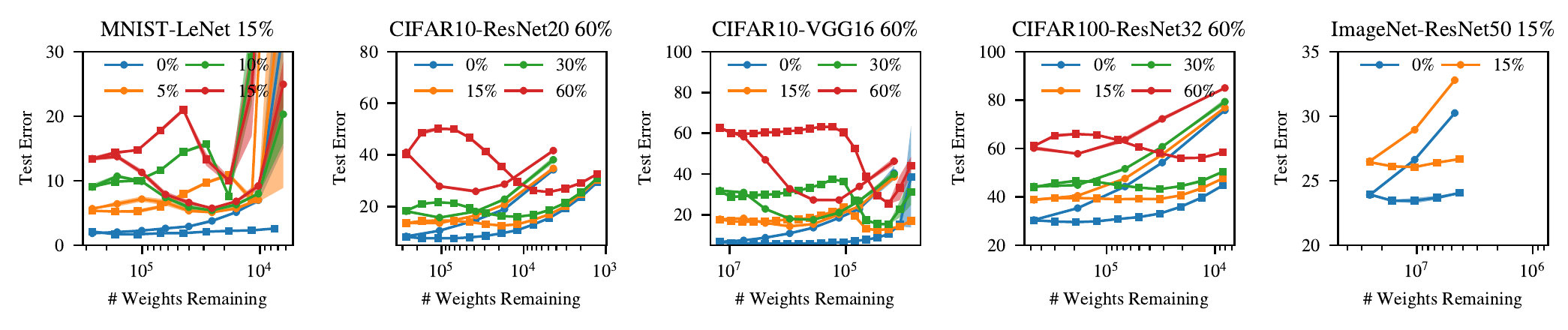}
\caption{
Size reduction does not explain the full extent of pruning's benefits to generalization.
We show the test errors of the dense/sparse models with round/square markers respectively as a function of the weights remaining.
Legends show label noise level.
\Cref{appx:pruning-vs-width-scaling} includes detailed views.}
\label{fig:pruning-vs-width-scaling-main-body}%
\end{figure*}

\textbf{Results.}
\Cref{fig:pruning-vs-width-scaling-main-body} presents the test errors of models that pruning and width down-scaling produce, using square and circle markers respectively.
Width down-scaling under-performs pruning on 4 out of 5 benchmarks on datasets without random label noise.
When training with random label noise, width down-scaling has notable regularization effects and significantly improves generalization, similar to pruning.
In \Cref{sec:similarity-between-pruning-and-width-scaling}, we show that the regularization effects of both algorithms manifest similarly across example subgroups, suggesting that pruning's regularization effect is a consequence of model size reduction in general.
However, despite this improvement, a gap remains between the minimum test error of models that pruning and width down-scaling produce achieve:
pruning does better on LeNet, ResNet20, VGG-16, ResNet32 and ResNet50 benchmarks by -4 (pruning does worse) to 0.4\%, 0\% (matching), 1.5 to 1.8\%, 0 (matching) to 2.1\%, and 0.36\% in test error, respectively.

\textbf{Conclusion.}
On standard datasets with random label noise, width down-scaling produces models with similar generalization (no worse by more than 2.1\%) as the optimally sparse models.
However, without random label noise, width down-scaling under-performs pruning.

\subsection{Conclusion}
By removing the weight removal part from pruning in \Cref{subsec:edt}, we show the importance of pruning-induced additional regularization for achieving good generalization in the presence of label noise more clearly than \Cref{sec:dist-dependence}.
Similarly, by removing the retraining part from pruning in \Cref{sec:size-reduction}, we show the importance of pruning-induced  better training for achieving good generalization using the original dataset without random label noise more clearly than \Cref{subsec:aggregate-effect}.

We confirm that both effects of pruning are required to fully account for its impact on generalization.
Our deconstruction of pruning's effect in terms of training and regularization is therefore minimal.

\section{Related Work}
Pruning has an extended history: early work starting from the 1980s found that pruning enhances model interpretability and generalization  \citep{10.5555/2969735.2969748, NIPS1989_6c9882bb, 298572}. 
More recently, the advent of deep neural network models motivates the adoption of pruning to reduce the storage and computational demand of deep models \citep{DBLP:journals/corr/HanPTD15, DBLP:journals/corr/HanMD15, Lebedev_2016_CVPR, liu2017learning, MLSYS2020_d2ddea18,molchanov2017pruning,NIPS2017_c5dc3e08,DBLP:journals/corr/abs-1711-05908,baykal2018datadependent,lee2018snip,wang2019eigen,DBLP:journals/corr/abs-2001-00218,Lee2020A,Liebenwein2020Provable,NEURIPS2020_d1ff1ec8,NEURIPS2021_e35d7a57,NEURIPS2021_a376033f,pmlr-v162-yu22f,DBLP:journals/corr/abs-2006-05467}.
In this section, we describe several branches of pruning research pertinent to our work.

\textbf{Pruning algorithm design.}
Recent pruning research often focuses on improving pruning algorithm design.
An extended line of research studies the heuristics that determine which weights to remove \citep{molchanov2017pruning, NIPS2017_c5dc3e08, lee2021layeradaptive} -- the simplest heuristic is to remove weights with the smallest magnitude.
For example, \citet{molchanov2017variational,louizos2018learning} propose to learn which weights to prune as part of the optimization process.
Another line of research concerns the structure of the weights to remove to achieve computational speedup without significantly hurting model generalization.
For example, \citet{li2017pruning, liu2018rethinking} propose to prune weights in groups or to prune neurons, convolutional filters and channels.
Our work does not produce a new pruning algorithm design.
Rather, we study the existing pruning algorithms to understand their effect on model generalization.

\textbf{Pruning's effect on generalization.}
There are many reported instances in which pruning benefits model generalization \citep{NIPS1989_6c9882bb,DBLP:journals/corr/HanPTD15, frankle2018lottery}.
This effect has sparked interest in examining pruning as a technique to gain additional generalization benefits beyond common regularization techniques.
\emph{Weight decay} is a typical regularization technique that encourages model weights to have a small $L_2$ norm as part of the training objective.
\citet{317740} demonstrated that pruning can improve the generalization of recurrent neural network models better than weight decay.
\citet{Thimm95evaluatingpruning, Augasta2013PruningAO} compared the effect on the generalization of several contemporary pruning techniques.
\citet{DBLP:journals/corr/abs-1906-03728} discovered a positive correlation between \emph{pruning instability}, defined as the drop in test accuracy immediately following weights removal, and model generalization.
The authors found that higher pruning-induced instability leads to increased flatness of minima, which in turn improves generalization.
\NA{Our work differs from prior work as we study pruning's effects on generalization through a novel perspective, examining pruning's impact on training examples.}

\textbf{Pruning's effect beyond generalization.} Our work contributes to a growing line of work \citep{DBLP:journals/corr/abs-1911-05248, MLSYS2021_2a79ea27} investigating pruning's effect on examples beyond test error.
\citet{DBLP:journals/corr/abs-1911-05248} discovered that model compression can have a disproportionately large impact on predicting the under-represented long-tail of the data distribution.
\citet{MLSYS2021_2a79ea27} studied the performance of pruning using metrics such as out-of-distribution generalization and resilience to noise, and found that pruning may not preserve these alternative performance metrics even when it preserves test accuracy.
To the best of our knowledge, our work is the first to examine pruning's impact on examples in the training set.

\section{Closing Discussion}
\label{sec:conclusion}

\textbf{Importance of training improvement.}
Our work sheds light on an overlooked and underestimated effect of pruning that boosts generalization -- training improvement.
Seminal work on pruning \citep{298572, DBLP:journals/corr/HanPTD15} focuses on its regularization effect and attributes pruning's beneficial influence on generalization to regularization that comes with model size reduction.
Instead, our work discovers that, on standard image classification benchmarks, the state-of-the-art pruning algorithm attains the optimal generalization consistently by improving model training.

\textbf{How does model size reduction impact generalization?}
The relationship between \emph{model complexity} and generalization has long been a subject of immense interest \citep{NIPS2000_0950ca92, hastie_friedman_tisbshirani_2017, Belkin15849, DBLP:journals/corr/abs-1912-02292}.
\NA{Informally, model complexity refers to the model's ability to fit a wide variety of functions \citep{Goodfellow-et-al-2016}.}\fTBD{Can we give any better definition of model complexity?}
For example, model size \citep{Belkin15849} and the norm of model weights \citep{pmlr-v40-Neyshabur15} both provide meaningful measures of model complexity for the study of generalization.
Classical theories (e.g., on VC dimension \citep{vapnik:264} and Rademacher complexity \citep{10.5555/944919.944944}) show that reducing model complexity decreases the likelihood for the learning algorithm to produce trained models with a large gap between training and test loss, and may therefore improve generalization,
We instead focus on learning dynamics of training example subgroups and show that reducing model size improves generalization by mitigating the accuracy degradation due to noisy examples.
Therefore, our work contributes a complementary perspective to the understanding of the relationship between model complexity and generalization.

\textbf{Societal impact.} Our work examines pruning's effect on generalization. 
While we measure generalization using test error, \citet{DBLP:journals/corr/abs-1911-05248} show that test error may be an imperfect characterization of model prediction quality due to a lack of consideration of fairness and equity. 
Concretely, our work does not address the potential disproportionate impact of model pruning on label categories.
Our work must therefore be interpreted within the context of the community's current and future understanding of pruning's potential contribution to systemic bias, especially against minority groups that may be underrepresented in existing training datasets.

\textbf{Implication for fairness.}
A growing line of work concerns the adverse impact of pruning on model fairness \citep{DBLP:journals/corr/abs-1911-05248, DBLP:journals/corr/abs-2010-03058, DBLP:journals/corr/abs-2106-07849, DBLP:journals/corr/abs-2106-07849}  -- they show that pruning disproportionately worsens the prediction accuracy on a small subgroup of examples.
However, our results show that their work does not completely characterize pruning's effects pertinent to fairness:
(1) We show that only at relatively high sparsities can pruning harm the prediction accuracy of any example subgroup, where pruning's regularization effects dominate.
(2) We show that size reduction in general, rather than pruning in particular, underpins the disproportionately worse accuracy of the pruned model on certain subgroups of examples.
Our results call for further assessment and discussion of pruning's risk to model fairness.

\textbf{Conclusion.} %
We show that the long-standing size-reduction hypothesis attributing pruning's beneficial effect on generalization to its reduction of model size does not fully explain pruning's impact on generalization.
Inspired by studies \citep{toneva2018an,DBLP:journals/corr/abs-2107-07075} that show nonuniform effects of training examples on generalization, we develop an analysis to study pruning's impact on generalization by interpreting its effect on an example's training loss in relation to the example's influence on generalization.

With this novel analysis, we find that, at the optimal sparsity, pruning leads to either \bettertraining or additional regularization, which improves training loss over the dense model, and reduces the accuracy degradation due to noisy examples over the dense model, respectively.
Both effects contribute to improving model generalization.

Our novel analysis adds to our empirical toolkit for studying the effect of a learning algorithm on generalization through its effect on the loss of training examples.
Using our novel analysis, we derive findings that advance our empirical understanding of pruning as a learning algorithm.

\bibliography{paper}
\bibliographystyle{abbrvnat}

\clearpage
\section*{Checklist}

The checklist follows the references.  Please
read the checklist guidelines carefully for information on how to answer these
questions.  For each question, change the default \answerTODO{} to \answerYes{},
\answerNo{}, or \answerNA{}.  You are strongly encouraged to include a {\bf
justification to your answer}, either by referencing the appropriate section of
your paper or providing a brief inline description.  For example:
\begin{itemize}
  \item Did you include the license to the code and datasets? \answerYes{See Section.}
  \item Did you include the license to the code and datasets? \answerNo{The code and the data are proprietary.}
  \item Did you include the license to the code and datasets? \answerNA{}
\end{itemize}
Please do not modify the questions and only use the provided macros for your
answers.  Note that the Checklist section does not count towards the page
limit.  In your paper, please delete this instructions block and only keep the
Checklist section heading above along with the questions/answers below.

\begin{enumerate}

\item For all authors...
\begin{enumerate}
  \item Do the main claims made in the abstract and introduction accurately reflect the paper's contributions and scope?
    \answerYes
  \item Did you describe the limitations of your work?
    \answerYes{\Cref{sec:conclusion}}
  \item Did you discuss any potential negative societal impacts of your work?
    \answerYes{\Cref{sec:conclusion}}
  \item Have you read the ethics review guidelines and ensured that your paper conforms to them?
    \answerYes{}
\end{enumerate}

\item If you are including theoretical results...
\begin{enumerate}
  \item Did you state the full set of assumptions of all theoretical results?
    \answerNA{}
        \item Did you include complete proofs of all theoretical results?
    \answerNA{}
\end{enumerate}

\item If you ran experiments...
\begin{enumerate}
  \item Did you include the code, data, and instructions needed to reproduce the main experimental results (either in the supplemental material or as a URL)?
    \answerNo{}
  \item Did you specify all the training details (e.g., data splits, hyperparameters, how they were chosen)?
    \answerYes{\Cref{sec:expdetails}}
        \item Did you report error bars (e.g., with respect to the random seed after running experiments multiple times)?
    \answerYes{\Cref{sec:prelim}}
        \item Did you include the total amount of compute and the type of resources used (e.g., type of GPUs, internal cluster, or cloud provider)?
    \answerYes{\Cref{sec:expdetails}}
\end{enumerate}

\item If you are using existing assets (e.g., code, data, models) or curating/releasing new assets...
\begin{enumerate}
  \item If your work uses existing assets, did you cite the creators?
    \answerYes{\Cref{sec:prelim}}
  \item Did you mention the license of the assets?
    \answerNA{}
  \item Did you include any new assets either in the supplemental material or as a URL?
    \answerNo{}
  \item Did you discuss whether and how consent was obtained from people whose data you're using/curating?
    \answerNA{}
  \item Did you discuss whether the data you are using/curating contains personally identifiable information or offensive content?
    \answerNA{}
\end{enumerate}

\item If you used crowdsourcing or conducted research with human subjects...
\begin{enumerate}
  \item Did you include the full text of instructions given to participants and screenshots, if applicable?
    \answerNA{}
  \item Did you describe any potential participant risks, with links to Institutional Review Board (IRB) approvals, if applicable?
    \answerNA{}
  \item Did you include the estimated hourly wage paid to participants and the total amount spent on participant compensation?
    \answerNA{}
\end{enumerate}

\end{enumerate}

\clearpage

\appendix

\section{Acknowledgement.}
We thank Zack Ankner, Xin Dong, Zhun Liu, Jesse Michel, Alex Renda, Cambridge Yang, and Charles Yuan for their helpful discussion and feedback to this project. 
This work was supported in part by a Facebook Research Award, the MIT-IBM Watson AI-LAB, Google’s Tensorflow Research Cloud, and the Office of Naval Research (ONR N00014-17-1-2699).
Daniel M. Roy is supported in part by an NSERC Discovery Grant and Canada CIFAR AI Chair funding through the Vector Institute.
Part of this work was done while Gintare Karolina Dziugaite and Daniel M. Roy were visiting the Simons Institute for the Theory of Computing.

\section{Additional Experimental Details}
\label{sec:expdetails}

We complement the description of experimental methods in \Cref{sec:prelim} with additional details.

\subsection{Models and Datasets}

We study pruning's effect on generalization using LeNet \citep{lenet}, VGG-16 \citep{Simonyan15}, ResNet-20, ResNet-32 and ResNet-50 \citep{7780459}.
We use the MNIST dataset, consisting of 60,000 images of handwritten digits whose labels correspond to 10 integers between 0 and 9.
We also use the CIFAR-10 and -100 \citep{CIFAR-10} datasets, which consists of 60,000 images in 10 and 100 classes.
For both datasets, we draw 2,000 of the original training images randomly as validation set; we continue to use the remaining 48,000 of the training images as training set.
We use all 10,000 original test images as our test set.
We use the ImageNet dataset \citep{deng2009imagenet} as well, which contains 1,281,167 images in 1,000 classes.
We again randomly draw 50,000 images as validation set and use the remaining 1,231,167 training images as our training set.
We use all 50,000 original test images as our test set.

\subsection{Training Hyperparameters}
\label{subsec:training-hyperparameters}

Our training hyper-parameter configuration follows the precedent of \citet{frankle2018lottery, DBLP:journals/corr/abs-1912-05671, DBLP:journals/corr/abs-2002-07376}: details are available in \Cref{tbl:hyper-parameter}.

\begin{table}[]
\centering
\resizebox{\textwidth}{!}{
\begin{tabular}{@{}cccccccccc@{}}
\toprule
Model    & Dataset  & Epochs & Batch & Opt.                                                                                           & LR     & LR Drop                 & Weight Decay & Initialization & Rewind Epoch \\ \midrule
LeNet    & MNIST    & 60     & 128   & \begin{tabular}[c]{@{}c@{}}Adam($\beta_1=0.9$ \\ $\beta_2=0.999$, \\ $\epsilon=1e-8$)\end{tabular} & 1.2e-3 & -                       & -            & Kaiming Normal & 0            \\
ResNet20 & CIFAR-10  & 160    & 128   & SGD($\mu=0.9$)                                                                                 & 0.1    & 10x at epoch 80, 120    & 1e-4         & Kaiming Normal & 10           \\
VGG-16    & CIFAR-10  & 160    & 128   & SGD($\mu=0.9$)                                                                                 & 0.1    & 10x at epoch 80, 120    & 1e-4         & Kaiming Normal & 10           \\
ResNet32 & CIFAR-100 & 160    & 128   & SGD($\mu=0.9$)                                                                                 & 0.1    & 10x at epoch 80, 120    & 1e-4         & Kaiming Normal & 10           \\
ResNet50 & ImageNet & 90     & 768   & SGD($\mu=0.9$)                                                                                 & 0.3    & 10x at epoch 30, 60, 80 & 1e-4         & Kaiming Normal & 10           \\ \bottomrule
\end{tabular}}
\caption{We use standard hyperparameters following the precedent of \citet{frankle2018lottery, DBLP:journals/corr/abs-1912-05671, DBLP:journals/corr/abs-2002-07376}. $\mu$ in SGD configuration parameter denotes momentum.}
\label{tbl:hyper-parameter}
\end{table}

\subsection{EL2N Score Calculation}
\label{sec:calc-el2n}
We follow the method that \citet{DBLP:journals/corr/abs-2107-07075} describes to compute the EL2N scores of examples in the training dataset.
For a model architecture $f$ and dataset $S$, we first train $N$ of a model for $K\%$ of the total training time to obtain partially trained weights $\theta_n, n=1, \cdots, N$ for each model.
Subsequently, we compute the EL2N score for each image-label pair $(x, y)$, as $\frac{1}{N} \sum_n \|p_{\theta_n}(x) - y \|_2$, where $y$ is one-hot label, and $p_{\theta_n}(x)$ is the softmax output of the model.
Following the precedent of  \citep{DBLP:journals/corr/abs-2107-07075}, we take $N=10$ and $K=10$.
There is no precedent for choosing the appropriate value for $N, K$ for the ImageNet benchmark, and we find setting $N=22$, which corresponds to measuring EL2N scores at the $20$th epoch of training, and $K=10$ to be a reasonable choice.

\section{Size-Reduction Does Not Explain Pruning's Benefits to Generalization}
\label{appx:more-parameter-removal-ablation}

In \Cref{sec:intro}, we develop an augmented version of the pruning algorithm, by modifying the pruning algorithm to no longer remove weights.
This augmented training algorithm corresponds to extended dense training with a cyclic learning rate schedule (EDT) we examine in \Cref{subsec:edt} -- it trains the model for the same number of epochs and with the same cyclic learning rate schedule as pruning.

In \Cref{sec:intro}, we highlight the similarity between the generalization of models that this EDT algorithm and pruning produce.
This similarity shows that the size-reduction hypothesis does not fully explain pruning's effect on the generalization of VGG-16 model.
In this section, we demonstrate that, with the exception of LeNet on MNIST, this phenomenon generalizes to other model architectures and pruning algorithms, thereby showing that size-reduction does not fully explain pruning's benefits to generalization in general.

\paragraph{Method.}
For 5 model architecture and dataset combinations, we generate a family of models with different sparsities (or training epochs, in the case of the augmented training algorithm), using the following algorithms including 5 variants of pruning algorithms and the EDT algorithm:
\begin{enumerate}
\item Learning rate rewinding, as described in \Cref{sec:prelim}.
\item A variant of iterative magnitude pruning called \textit{weight rewinding} \citep{frankle2018lottery, Renda2020Comparing}. 
At the end of each pruning iteration, this algorithm rewinds not only the learning rate, but also values of remaining weights by resetting their values to the values they had had earlier in training. 
\item Learning rate rewinding, but modified to remove weights according to a gradient-based weight selection criterion called \textit{SynFlow}  \citep{DBLP:journals/corr/abs-2006-05467,frankle2021pruning}.
    This pruning algorithm uses the following heuristic to remove weights:
    the pruning algorithm first replaces each weight $w_l$ in the model with $\|w_l\|$;
    then, it feeds an input tensor filled with all 1's to this instrumented model, and the sum of the output logits is computed as $R$.
    This pruning algorithm then assigns an importance score $\| \frac{dR}{dw_l} \odot w_l  \|$ to each weight, and remove the weights receiving the lowest such scores.
  \citet{DBLP:journals/corr/abs-2006-05467} designed SynFlow to mitigate \textit{layer collapse}, a phenomenon associated with ordinary magnitude-based pruning algorithm where weight removal concentrates on certain layers, effectively disconnecting the sparse model.
\item Learning rate rewinding, but modified to remove weights according to a gradient-based selection criterion called \textit{SNIP}  \citep{lee2018snip}.
    This pruning algorithm computes gradient $g_l$ for each layer $l$ using samples of training data.
    It then assigns an importance score $|w_l \odot g_l|$ to each weight, and removes the weights receiving the lowest such scores.
    The intuition behind this importance score is that it prevents pruning weights with the highest "effect on the loss (either positive or negative)" \citep{frankle2021pruning, lee2018snip}.
\item \textit{Iterative random} pruning, where a random set of weights are removed at each pruning iteration. The algorithm otherwise behaves like learning rate rewinding.
\item The \textit{augmented} algorithm, which trains the model for the same number of epochs and with the same cyclic learning rate schedule as pruning, without removing weights.
\end{enumerate}

\begin{figure}[h!]
\includegraphics[width=\linewidth]{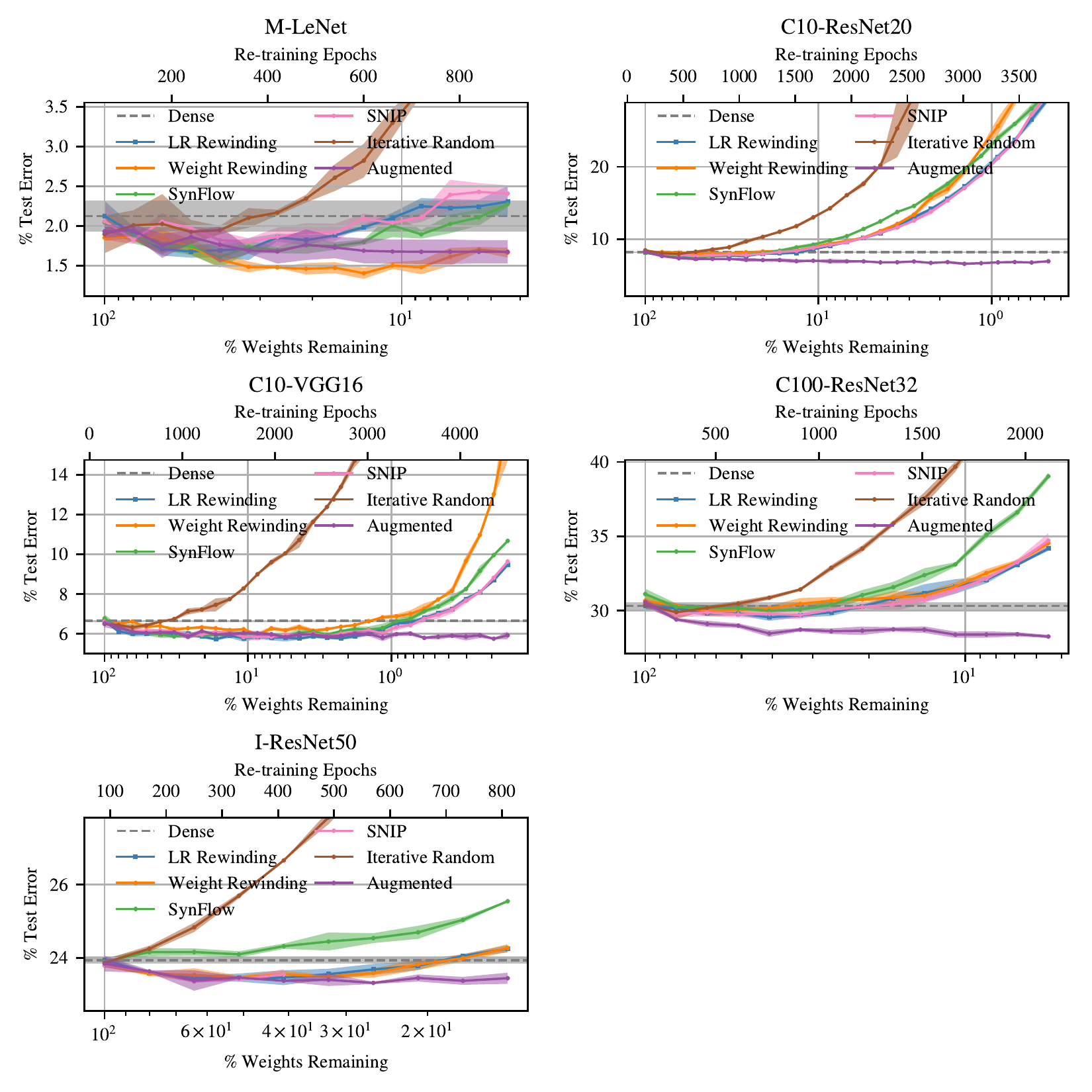}
\caption{Comparing variants of pruning algorithm.
With the exception of MNIST-LeNet benchmark, test errors of models we generate using the augmented pruning algorithm modified to no longer remove weights matches or exceeds the test errors of models we generate using all other variants of pruning algorithms.}
\label{fig:all-pruning-variants}
\end{figure}

In \Cref{tbl:pruning-ablation-study}, for each benchmark and algorithm, we tabulate the test error of the model with the best validation error, selected from the family of models that each aforementioned algorithm generates.
In \Cref{fig:all-pruning-variants}, we plot generalization of all models within the family of models each aforementioned algorithm generates as a function of sparsities and training time.

\begin{table}[h!]
\centering
\begin{tabular}{@{}cccccc@{}}
\toprule
                                                            & M-LeNet & C10-ResNet20 & C10-VGG-16 & C100-ResNet32 & I-ResNet50 \\ \midrule
\begin{tabular}[c]{@{}c@{}}LR \\ Rewinding\end{tabular}     & 1.8±0.1 & 7.7±0.1      & 5.9±0.1   & 29.9±0.3      & 23.4±0.1   \\ \midrule
\begin{tabular}[c]{@{}c@{}}Weight \\ Rewinding\end{tabular} & 1.5±0.1 & 8.1±0.3      & 6.2±0.1   & 30.6±0.4      & 23.5±0.1   \\ \midrule
SynFlow                                                     & 1.7±0.1 & 7.8±0.3      & 6.1±0.1   & 30.2±0.2      & 23.9±0.1   \\ \midrule
SNIP                                                        & 1.9±0.1 & 7.7±0.1      & 6.1±0.1   & 29.6±0.2      & 23.4±0.0   \\ \midrule
\begin{tabular}[c]{@{}c@{}}Iterative \\ Random\end{tabular} & 1.9±0.2 & 8.1±0.3      & 6.4±0.2   & 30.2±0.1      & 23.9±0.1   \\ \midrule
EDT                                                   & 1.8±0.1 & 7.6±0.3      & 6.0±0.1   & 29.0±0.1      & 23.4±0.3   \\ \bottomrule
\end{tabular}
\caption{We tabulate test errors of the model with the minimum validation error that pruning and the EDT algorithm generate.
With the exception of MNIST-LeNet, the EDT algorithm matches or exceeds the generalization of models that all pruning algorithms we test generate.}
\label{tbl:pruning-ablation-study}
\end{table}

\paragraph{Results.}
In \Cref{tbl:pruning-ablation-study}, for each benchmark and algorithm, we select, from the family of models that each aforementioned algorithm generates, the model with the best validation error and tabulate its test error.
In \Cref{fig:all-pruning-variants}, we plot the generalization of the family of models each aforementioned algorithm generates as a function of sparsities and training time in epochs.
We observe that with the exception of LeNet on MNIST, the generalization of dense models that the augmented training algorithm produces matches or exceeds the generalization of sparse models that pruning algorithms produce.
For LeNet on MNIST, however, the augmented training algorithm under-performs weight rewinding.
We conjecture, without testing, that for this benchmark, the lack of any explicit form of regularization in the training process makes pruning's regularization effect uniquely important for generalization.

\paragraph{Conclusion.}
In \Cref{sec:intro}, We show that the augmented training algorithm produces VGG-16 models with generalization that is indistinguishable from that of models that pruning with learning rate rewinding produces.
In this section, We further show that this phenomenon shows up even if we apply other iterative pruning algorithms to additional model-dataset combinations.
We therefore conclude that in general, size reduction does not fully account for pruning's benefits to generalization.

\section{Effects of Leaving Out Pruning-Affected Examples}
\label{sec:pruning-affected-examples}

In this section, we focus on the subgroups of examples whose training loss improves when pruning to a range of generalization-improving sparsities.
By excluding them from the training dataset, we empirically examine their influence on the generalization of the dense models.

\paragraph{Method.}
We refer to the top $K\%$ of training examples whose training loss improves the most during pruning as the \textit{top-improved examples}.
To examine the influence of these top-improved examples on generalization, for each sparsity pruning reaches, we train two dense models on two datasets respectively: a). the original training dataset excluding the top-improved examples at the specified sparsity, which we denote as TIE (\textbf{T}op-\textbf{I}mproved \textbf{E}xamples); b). a dataset of the same size as a)., but consisting of randomly drawn examples, which we denote as RND.
We then compare their resulting generalization.
If the top-improved examples are indeed the ones with the largest influence on generalization, excluding them from the training dataset should affect generalization more so than excluding a random subset of the same size.
We set $K=20$ in our experiments, we also tried setting $K=10$, but the resulting generalization difference between models we train on two datasets is negligible.

\begin{figure}[h!]
\includegraphics[width=\linewidth]{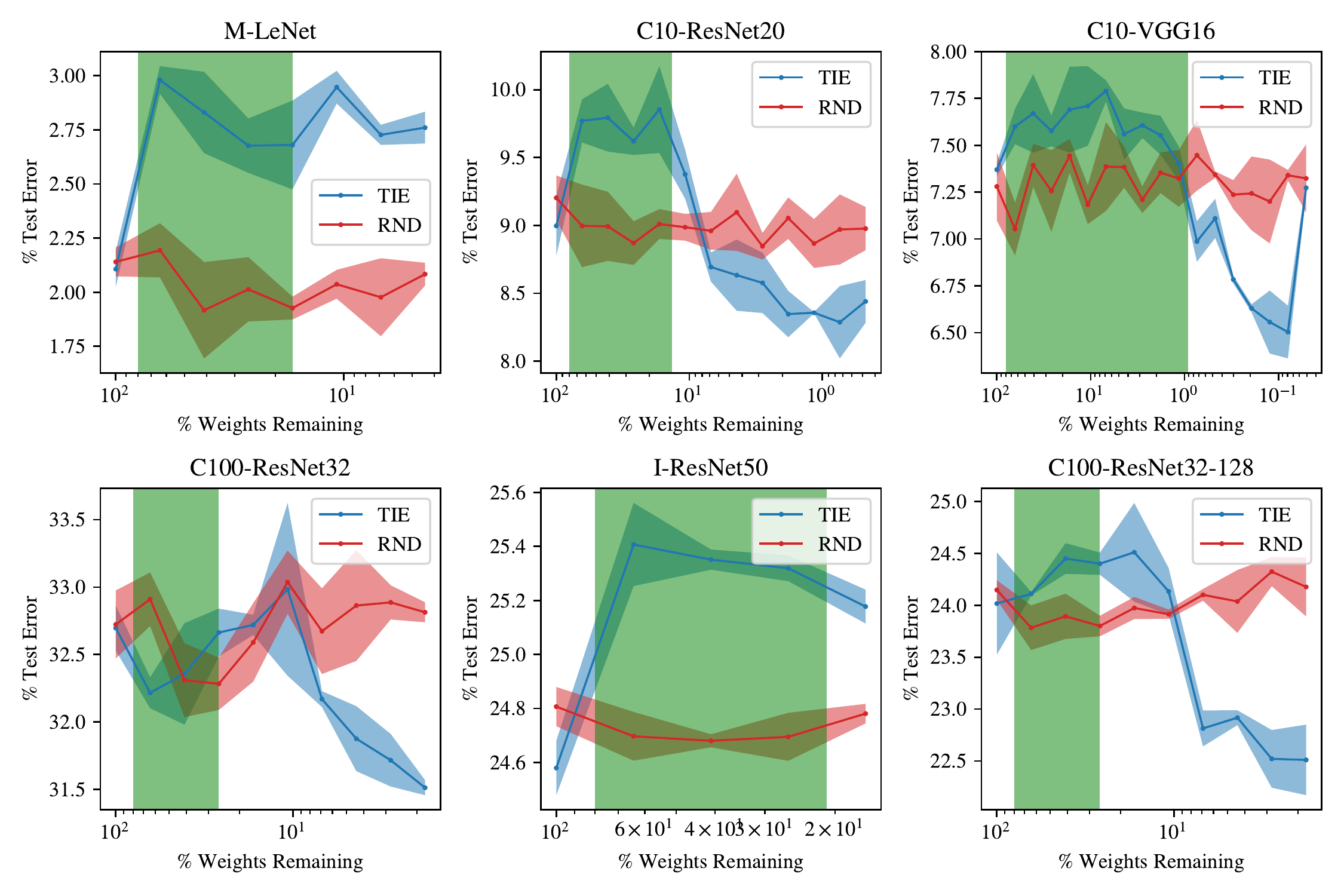}
\caption{Excluding top-improved examples hurt generalization more than excluding a random subset of the same size on 4/5 benchmarks.
We show test errors of dense models we train on the dataset excluding top-improved examples (denoted as TIE), and on the dataset excluding a random subset of the same size (denoted as RND) as a function of sparsity at which the top-improved examples are selected.
The range of sparsities where generalization is improved by pruning is shaded in green.}
\label{fig:excluding-tie}
\end{figure}

\paragraph{Result.}
We show in \Cref{fig:excluding-tie} the generalization of dense models on the dataset excluding top-improved examples (TIE) and randomly drawn dataset of the same size (denoted RND) as a function of the sparsity at which the top-improved examples are selected.
\Cref{fig:excluding-tie} shows that, averaging within the sparsities where generalization improves (the range of sparsities in green), excluding top-improved examples hurts generalization more than excluding a random subset of examples of the same size by 0.78\%, 0.77\%, 0.34\%, 0.69\% for LeNet on MNIST, ResNet20 on CIFAR-10, VGG-16 on CIFAR-10, ResNet50 on ImageNet benchmarks.
For ResNet32 on CIFAR-100 benchmark, we do not observe a significant difference between them.
However, increasing the width of the ResNet32 model on CIFAR-100 from 16 to 128, we find that excluding the top-improved examples hurts generalization more than excluding a random subset of examples of the same size by 0.44\% (c.f. the 1st image from the right on the 2nd row), similar to our finding on the remaining benchmarks.

\paragraph{Conclusion}
The top-improved examples affect generalization to a greater extent than a randomly chosen set of examples of the same size.
Moreover, on standard image classification benchmarks, they are more beneficial to generalization than randomly chosen examples.

\section{Leaving Out Noisy Examples Improves Generalization}
\label{sec:make-dense-model-ignore-noisy-data}

\Cref{sec:dist-dependence} demonstrates that, when pruning to the optimal sparsity in the presence of random label noise, its effect is to increase training loss on noisy examples.
In this section, we show that increasing training loss on the same set of noisy examples improves the generalization of dense models to the same, if not larger, extent as pruning.
Our results validate the connection between increased training loss on noisy examples to generalization improvement.

\begin{figure*}[h!]
\centering
\includegraphics[width=\linewidth]{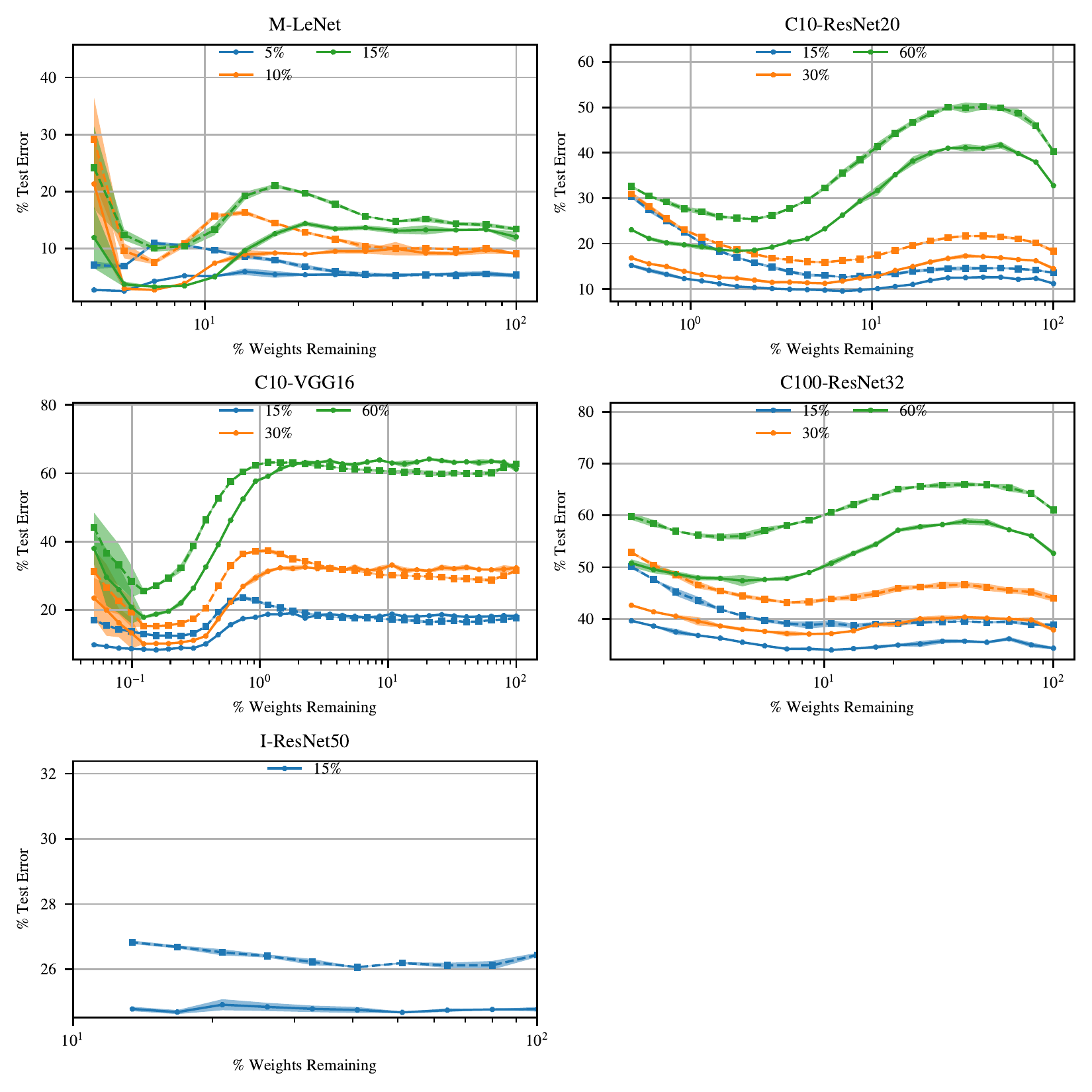}
\caption{Excluding examples that the optimally sparse models misclassify improves dense model generalization to the same if not greater extent as pruning does.
Test errors of dense-subset models are shown using circular markers with solid lines.
Test errors of sparse models are shown using square markers with dashed lines.
Legend shows levels of random label noise.
    }%
    \label{fig:pruning-vs-subset-dense}%
\end{figure*}

\paragraph{Method.}
Pruning to the optimal sparsity increases the training loss especially on a particular subgroup of examples consisting primarily of noisy examples.
A simple and hyperparameter-free method to increase training loss of dense models on the same set of examples is to exclude  from the training dataset the set of examples that the sparse models misclassify.
Since training only on a dataset subset changes the number of gradient steps an epoch takes, we increase the total number of epochs so that the total number of gradient steps taken remains the same as training on the original dataset.
We compare pruning and training dense models exclusively on the subset of examples that the sparse models correctly predict; we refer to such dense models as \textit{dense-subset} models.
In \Cref{fig:pruning-vs-subset-dense}, plot the test errors of the dense-subset models as a function of the sparsity of the pruned model whose prediction determines which examples to exclude when training the said dense-subset model.
For comparison, we also plot the test errors of sparse models as a function of its sparsity.
We summarize the numerical results in \Cref{tbl:cmp-pruning-with-dense-subset}.

\paragraph{Results.}
Variations in the generalization of the dense-subset models track variations in the generalization of the sparse models with respect to the fractions of weights remaining.
Moreover, the generalization of the optimal dense-subset models matches or exceeds the generalization of the optimally sparse models.

\paragraph{Conclusion.}
\Cref{sec:dist-dependence} shows that, in the presence of random label noise, pruning to the optimal sparsity has regularization effects: pruning increases the loss on a select subgroup of training examples consisting predominantly of noisy examples.
In this section, we demonstrate that increasing training loss on the same set of examples benefits dense model generalization as well.
Our results establish a connection between pruning's regularization effect and generalization improvement.

\section{Comparing Pruning with Width Down-scaling}
\label{appx:pruning-vs-width-scaling}

In this section, we present a less-cluttered version of the images we show in \Cref{sec:size-reduction}.

\begin{figure*}[h!]
\includegraphics[width=\linewidth]{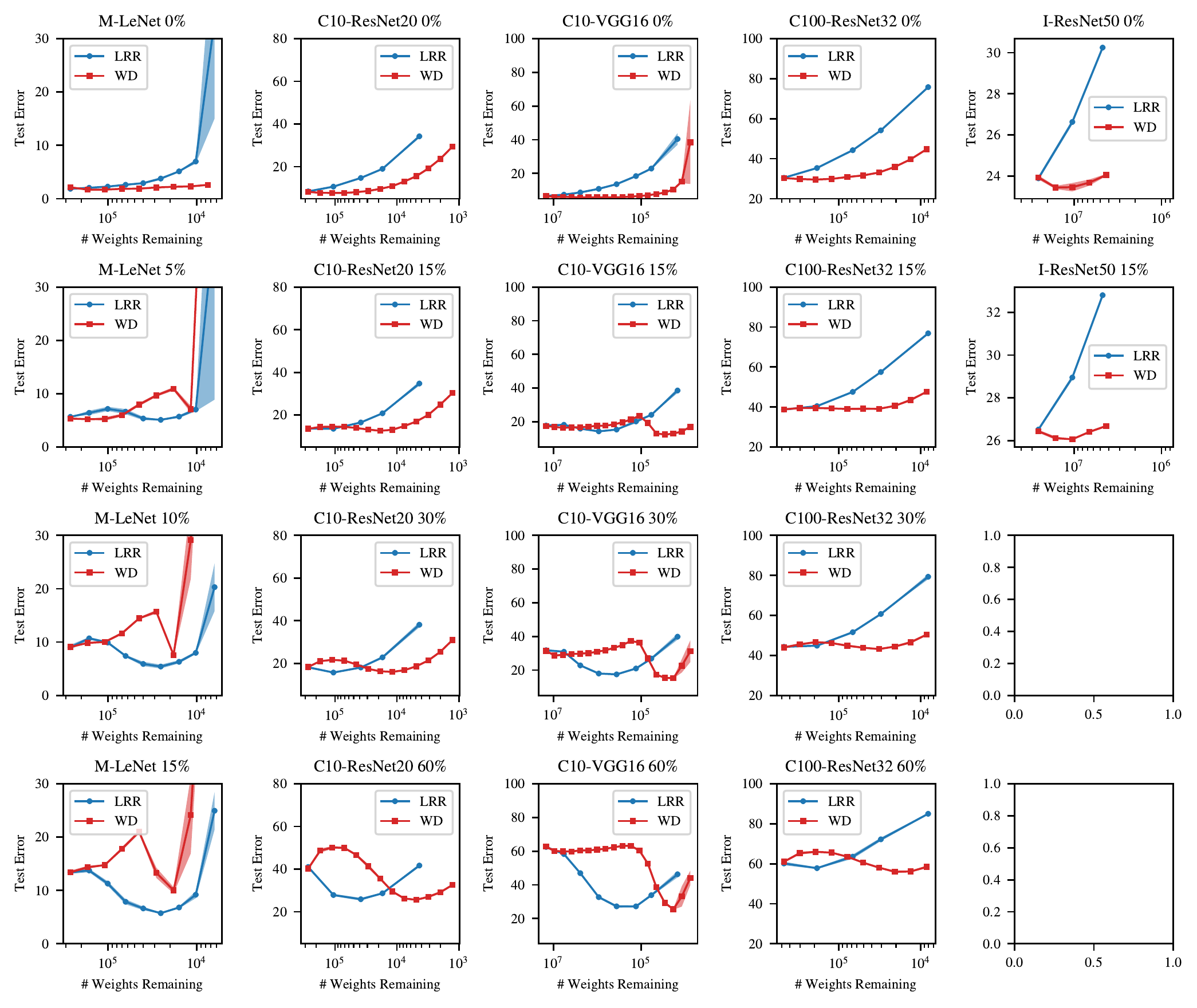}
	\caption{
	Comparing Pruning with Model Width Reduction.
	The percentage of random label noise injected is shown in each subplot title.
	The test errors of sparse models generated by learning rate rewinding algorithm (denoted LRR) and width-down-scaled models (denoted WD) as a function of their number of weights are shown using orange and blue lines, respectively.
	}%
    \label{fig:pruning-vs-width-scaling}%
\end{figure*}

\stepcounter{section}

\stepcounter{section}

\section{Similarities between Pruning and Width Down-scaling}
\label{sec:similarity-between-pruning-and-width-scaling}

In \Cref{sec:size-reduction}, we observe that the pruned models and width down-scaled models attain similar generalization at the optimal sparsity and model width on benchmarks with random label noise, where pruning's effect at the optimal sparsity is to improve generalization by strengthening regularization (\Cref{sec:dist-dependence}).
Their similarity is surprising because pruning and down-scaling model width are two distinct methods to reduce model size.
In this section, we further compare pruning and width down-scaling with equalized training time and quantify the similarities between their effects on training examples.

\paragraph{Method.}
Width down-scaling differs from pruning because a pruned model receives extra training, due to the retraining step.
To compare fairly, we modify the pruning and width down-scaling algorithms as such to equalize the amount of training:\footnote{Alternatively, one can equalize training by increasing the amount of training of width down-scaled models.
However this increases the amount of compute required by an order of magnitude, and beyond reasonable budget.
This is because unlike iterative pruning, a trained model with a larger width does not become the starting point for training a model with a smaller width, and therefore no amount of training is shared between models with different sizes.}

\begin{enumerate}
	\item For pruning, we adopt a variant of the magnitude pruning algorithm, called \emph{one-shot pruning} \citep{Renda2020Comparing}.
		While iterative pruning (the pruning algorithm we introduce in \Cref{sec:prelim}) removes a fixed fraction of the remaining weights per iteration until achieving the desired sparsity, one-shot pruning removes all weights at once to reach the desired sparsity.
		Like iterative pruning, one-shot pruning retrains the model after weights removal.
	\item For width down-scaling, we equalize the amount of training with one-shot pruning by training the width down-scaled model for the same number of gradient steps and using the same learning rate schedule as one-shot pruning.
\end{enumerate}

We then quantify the similarities between two algorithms by comparing (1). the generalization of their produced models; (2). average loss change, relative to the standard dense model, on training example subgroups with distinct EL2N percentile ranges due to respective algorithms; (3). overlap of the 10\% examples whose training loss changes the most due to two algorithms, relative to the training loss of the original dense model, measured in terms of \emph{Jiccard similarity index}. The Jiccard similarity index $J(A, B)$ measures the similarity between the content of two sets $A$ and $B$, with the following formula:

$$J(A, B) = \frac{|A \cap B|}{|A \cup B|}$$

\begin{figure*}[h!]
	\includegraphics[width=\textwidth]{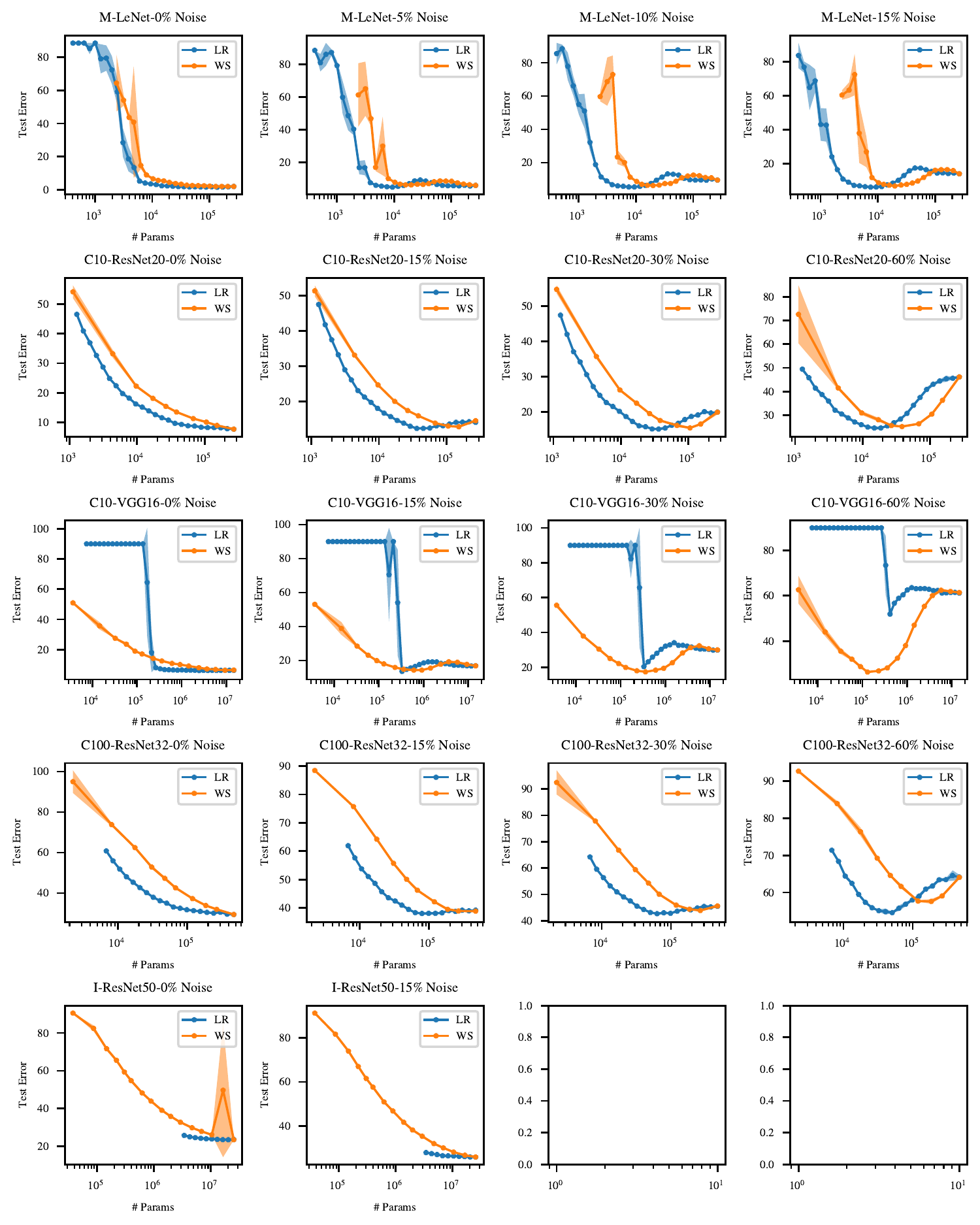}
	\caption{Generalization of pruned models (denoted LR) is similar to generalization of width down-scaled models (denoted WS).
	We equalize training time between pruning and width down-scaling.
	}
	\label{fig:width-scaling-vs-pruning-generalization}
\end{figure*}

\paragraph{Generalization.}
\Cref{fig:width-scaling-vs-pruning-generalization} shows that generalization of models that both algorithms produce are qualitatively similar on benchmarks with injected random label noise -- generalization of models that both algorithms generate improves when they only remove a relatively small fraction of weights.
As both algorithms remove more weights, generalization of produced models begins to suffer.

\begin{figure*}
	\includegraphics[width=\textwidth]{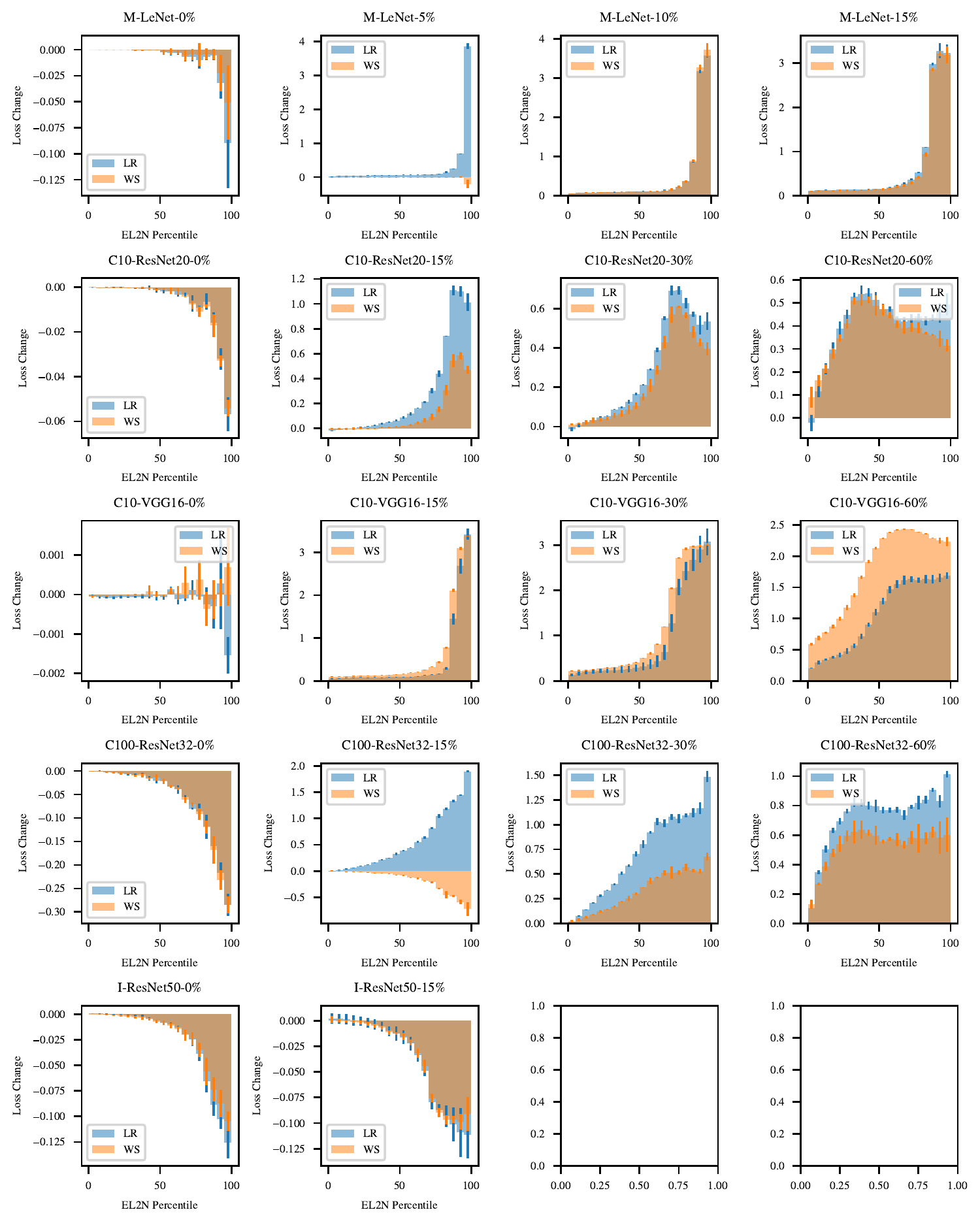}
	\caption{Pruning (denoted LR) and width down-scaling (denoted WS) has similar effect on the average training loss of example subgroups with distinct EL2N score percentile ranges at optimal sparsity and model width.}
	\label{fig:width-scaling-vs-pruning-el2n}
\end{figure*}

\paragraph{Change in loss by subgroups.}
\Cref{fig:width-scaling-vs-pruning-el2n} shows that on benchmarks with random label noise, pruning and width down-scaling has similar effects on example subgroups with distinct EL2N score percentile ranges at optimal sparsity and model width respectively -- training loss increase for both algorithms tends to concentrate on examples with high EL2N score percentile ranges.

\begin{table}[]
\centering
\begin{tabular}{@{}ccc@{}}
\toprule
\textbf{Model} & \textbf{Noise Level} & \textbf{Jaccard Similarity}                     \\ \midrule
M-LeNet        & 0\%/5\%/10\%/15\%    & 0.60 / 0.11 / 0.74 / 0.62 \\
C10-ResNet20   & 0\%/15\%/30\%/60\%   & 0.20 / 0.37 / 0.35 / 0.42 \\
C10-VGG-16      & 0\%/15\%/30\%/60\%   & 0.14 / 0.57 / 0.27 / 0.12 \\
C100-ResNet32  & 0\%/15\%/30\%/60\%   & 0.17 / 0.23 / 0.27 / 0.28 \\
I-ResNet50  & 0\%/15\%/   & 0.26 / 0.26 \\ \bottomrule
\end{tabular}
\caption{Jaccard similarity of top 10\% of examples most affected by pruning and width down-scaling.
	A randomly chosen two sets of 10\% training examples has a baseline similarity index of 0.05.
	All standard deviations are less than or equal to 0.03.}
\label{tbl:jaccard-similarity}
\end{table}

\paragraph{Affected examples overlap.}
\Cref{tbl:jaccard-similarity} shows that the 10\% of training examples whose training loss changes the most due to pruning and width down-scaling overlap significantly more than random baseline.

\paragraph{Conclusion.}
Pruning and width down-scaling have similar regularization effects: they nonuniformly increase the average training loss of example subgroups, which we show is key to improving generalization in the presence of random label noise (\Cref{sec:dist-dependence}).
\NA{Therefore, the regularization effect of pruning may be a consequence of model size reduction in general.}\fTBD{This appear to be the strongest statement we can say about the origin of pruning's regularization effect.}

\section{Ablation Studies}
\label{sec:ablation}
Pruning algorithms are complex aggregation of individual design choices.
In this section, we examine whether specific design choices of pruning algorithms including weight resetting and weight selection heuristics contributes to the generalization-improving effect of pruning.

\subsection{Rewinding Weights}
The pruning algorithm we use in this study, namely, learning rate rewinding, rewinds learning rate to their value early in training after each pruning iteration.
Its predecessor, an iterative pruning algorithm called weight rewinding rewinds not just learning rate, but also weights to their values early in training after each pruning iteration.
\citet{frankle2018lottery} proposed weight rewinding and later, \citet{Renda2020Comparing} found that rewinding weight is unnecessary if one's goal is to produce a family of models achieving the optimal parameter count - generalization trade-off.
In this section, we expand the comparison weight rewinding \citep{frankle2018lottery} and learning rate rewinding \citep{Renda2020Comparing} to datasets with random label noise, and show that rewinding weights is actually necessary for achieving optimal generalization on benchmarks with random label noise.

\paragraph{Method.}
To examine whether rewinding weights contributes to the generalization-improving effect of pruning,
we compare the generalization of models that weight rewinding produces with generalization of models that learning rate rewinding produces.
The former pruning algorithm rewinds weights and the latter does not.

\begin{figure*}[h!]
	\includegraphics[width=\linewidth]{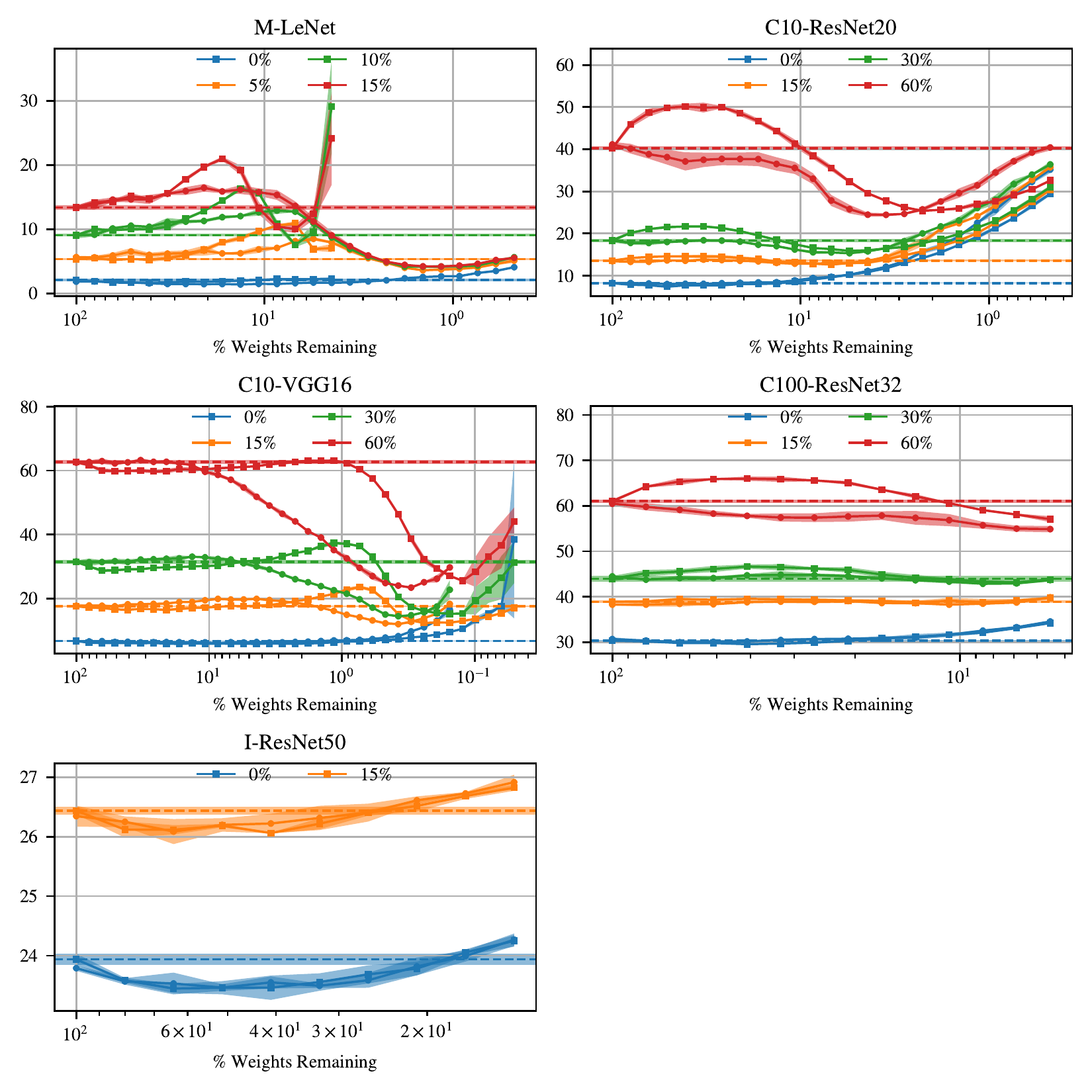}
	\caption{Rewinding weights helps improve generalization in the presence of random label noise.
	We plot test errors of models that weight rewinding and learning rate rewinding produce in lines with circle and square markers, respectively.
	We plot test errors of dense models with horizontal dashed lines.
	Legend shows noise levels.}
	\label{fig:weight-resetting}
\end{figure*}

\paragraph{Results.}
We plot our results in \Cref{fig:weight-resetting}.
Numerical results are available in \Cref{tbl:effect-of-weight-resetting} in \Cref{appx:raw-data}.
We observe that across 18 model, dataset, noise level combinations, weight rewinding out-performs learning rate rewinding on 10 benchmarks, whereas learning rate rewinding only out-performs weight rewinding on 2.
For the remaining 6 benchmarks, two algorithms produce models with matching generalization.
Notably, weight rewinding is better at mitigating random label noise than learning rate rewinding.
Across all model architectures, on datasets with and without random label noise, the optimal test error of models that weight rewinding produces is lower than the optimal test error of models that learning rate rewinding produces by -0.7\% to 0.3\%, 0\% (matching) to 5.5\%.

\paragraph{Conclusion.} Weight resetting contributes to the generalization-improving effect of pruning.

\subsection{Weight Selection Heuristics}
The pruning algorithm we study, namely learning rate rewinding \citep{Renda2020Comparing}, removes weights based on the simple heuristic that weights with low magnitude are less important for the given task.
We study whether pruning's generalization improving effect is affected by the choice of weight selection heuristic.

\paragraph{Method.}
We modify learning rate rewinding to adopt the following alternative weight selection heuristics, and compare the generalization of models that each modified pruning algorithm generates.

\begin{enumerate}
  \item Random selection. Weights are removed at random without considering their values.
  \item Synaptic flow preserving weight selection \citep{DBLP:journals/corr/abs-2006-05467}.
  This weight selection heuristic works as follow.
  The heuristic algorithm first replaces each weight $w_l$ in the model with $\|w_l\|$;
  then, the algorithm feed an input tensor filled with all 1's to this instrumented model, and the sum of the output logits is computed as $R$.
  The heuristic algorithm then assigns an importance score $\| \frac{dR}{dw_l} \odot w_l  \|$ to each weight, and the weights receiving the lowest such scores are removed.
  \citet{DBLP:journals/corr/abs-2006-05467} designed SynFlow to mitigate \textit{layer collapse}, a phenomenon associated with ordinary magnitude-based pruning method where weight removal concentrates on certain layers, effectively disconnecting the sparse model.
\end{enumerate}

\begin{figure*}[h!]
	\includegraphics[width=\linewidth]{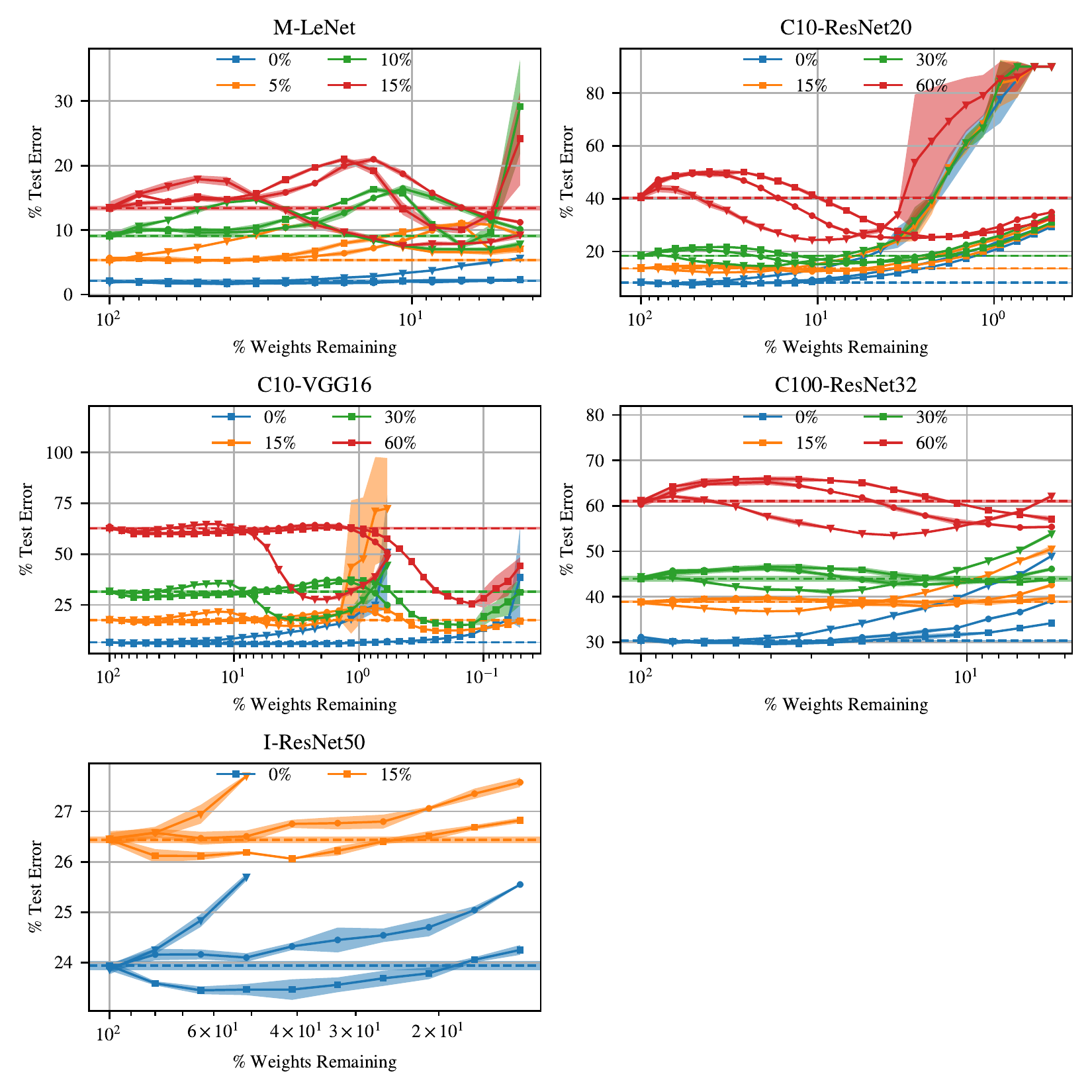}
	\caption{The choice of weight selection heuristic matters to generalization.
	However, among magnitude, synflow and random selection heuristics we test, not one selection strategy stands out as the best heuristic for improving generalization.
	We plot test errors of models that pruning with magnitude, synflow and random selection heuristics produce in lines with square, circle and triangle markers, respectively.
	We plot test errors of dense models with horizontal dashed lines.
	Legend shows noise levels.}
	\label{fig:weight-selection}
\end{figure*}

\paragraph{Results.}
We plot and tabulate the result of comparison in \Cref{fig:weight-selection} and \Cref{tbl:weight-selection-heuristics}.
We observe that SynFlow-based weight selection heuristic beats magnitude-based one on 6 out of 18 benchmarks.
We also observe that random pruning is particularly effective at mitigating random label noise -- on 9 out of 13 benchmarks with random label noise, random pruning out-performs magnitude-based weight selection heuristic.

\paragraph{Conclusion.} The choice of weight selection heuristic plays a significant role in determining pruning's generalization-improving effect.

\section{Closer Look at Examples with Training Loss Most Worsened by Pruning}
\label{appx:effect-of-avoided-examples}

In this section, we take a closer look at the examples in the standard image classification training datasets whose training loss is increased the most by pruning to a range of generalization-improving and generalization-preserving sparsities.

\subsection{Example Images and Labels}
\label{appx:visualizing-pruning-affected-examples}
When pruning to sparsities that improve or preserve (causing <2\% error increase) generalization, what is the nature of examples whose training loss increases or decreases the most?
Here, we present examples whose training loss is affected the most by pruning to the aforementioned sparsities.
We show that while the majority of these examples are atypical representations of their labels, a fraction of these examples contain incorrect or ambiguous labels.

\begin{table}[]
\centering
\begin{tabular}{@{}cccccc@{}}
\toprule
               & \begin{tabular}[c]{@{}c@{}}MNIST\\ LeNet\end{tabular} & \begin{tabular}[c]{@{}c@{}}CIFAR-10\\ ResNet20\end{tabular} & \begin{tabular}[c]{@{}c@{}}CIFAR-10\\ VGG-16\end{tabular} & \begin{tabular}[c]{@{}c@{}}CIFAR-100\\ ResNet32\end{tabular} & \begin{tabular}[c]{@{}c@{}}ImageNet\\ ResNet50\end{tabular} \\ \midrule
Most-worsened  & 1.4\%                                                 & 16.7-13.4\%                                                & 2.3-0.7\%                                               & 16.7-8.6\%                                                  & 13.4\%                                                      \\
Most-improved  & 80-5.5\%                                              & 80-21\%                                                    & 80-2.3\%                                                & 80-21\%                                                     & 80-16.8\%                                                   \\
Least-affected & 100-1.4\%                                             & 100-5.5\%                                                  & 100-0.19\%                                              & 100-8.6\%                                                   & 100-13.4\%                                                  \\ \bottomrule
\end{tabular}
\caption{Sparsities at which we measure training loss after pruning to select training dataset subsets.}
\label{fig:loss-measuring-sparsities}
\end{table}

\paragraph{Method.}
We select three subsets of training examples: the most-worsened examples, the most-improved examples and the least-affected examples as follow.
\begin{enumerate}
	\item \emph{The most-worsened examples:} we first measure the per-example training loss after pruning to the following sparsities.
For ResNet20 on CIFAR-10 and VGG-16 on CIFAR-10 benchmarks, a range of sparsities with an increased overall training loss but improved generalization exist.
We measure per-example training loss after pruning to these sparsities.
For LeNet on MNIST, ResNet32 on CIFAR-100 and ResNet50 on ImageNet benchmarks, however, no such sparsities exist.
We therefore choose to measure per-example training loss after pruning to sparsities with an increased overall training loss and generalization that is no worse than that of the dense models by 2\%.
We then rank examples in the training dataset based on the geometric average of their training loss increase after pruning to aforementioned sparsities, relative to their training loss before pruning.\footnote{Since we are concerned with examples whose training loss increases due to pruning, and in general, one cannot compute geometric average of arrays with negative numbers, we set negative training loss increase due to pruning to a small number $1e-5$, effectively ignoring them.}
We select examples with the most training loss increase as the most-worsened examples.
\item \emph{The most-improved examples:} we first measure the per-example training loss after pruning to sparsities with improved generalization and decreased training loss.
We then rank examples in the training dataset based on the geometric average of their training loss decrease after pruning to aforementioned sparsities, relative to their training loss before pruning.\footnote{Similar to how we select most-worsened examples, since we are concerned with examples whose training loss decreases due to pruning, and in general, one cannot compute geometric average of arrays with negative numbers, we set negative training loss decrease due to pruning to a small number $1e-5$, effectively ignoring them.}
We select examples with the most training loss decrease as the most-improved examples.

\item \emph{The least-affected examples:} we first measure the per-example training loss after pruning to sparsities that attains generalization no worse than that of the dense models by 2\%.
We then rank examples in the training dataset based on geometric average of the absolute value of their training loss change after pruning to aforementioned sparsities, relative to their training loss before pruning.
We select examples with the least absolute value of training loss change as the least-affected examples.
\end{enumerate}

For each type of training example subset, we tabulate the sparsities at which we measure per-example training loss after pruning in \Cref{fig:loss-measuring-sparsities}.
We average the per-example training loss across selected sparsities and across independent runs of the same experiment with distinct random seeds.

On each benchmark, we present the 20 examples from each category \Cref{fig:most-worsened-examples}.
Notably, the set of 20 least-affected examples are not unique, as many examples are practically unaffected by pruning.

\paragraph{Results.}
We describe our observations in the captions of \Cref{fig:most-worsened-examples}.

\paragraph{Conclusion.}
When pruning to sparsities that improves or preserves generalization (<2\% error increase), the set of examples whose training loss is most affected (i.e., either most worsened or most improved) are examples that are atypical representations of their labels.
A fraction of them have ambiguous or erroneous labels.
In contrast, the least affected examples are mostly unambiguous and canonical representations of the labels.

\begin{figure*}[h!]
\centering
    \subfloat[\centering MNIST-LeNet Most-worsened Examples.
    The 3rd image on the 1st row has the incorrect label (should be 5, but labeled 3).
    The 2nd image on the 1st row is labeled 7, but is indistinguishable from a 2.
    The 5th image on the 1st row is labeled 5, but it also looks like an 8.]{{\includegraphics[width=\linewidth]{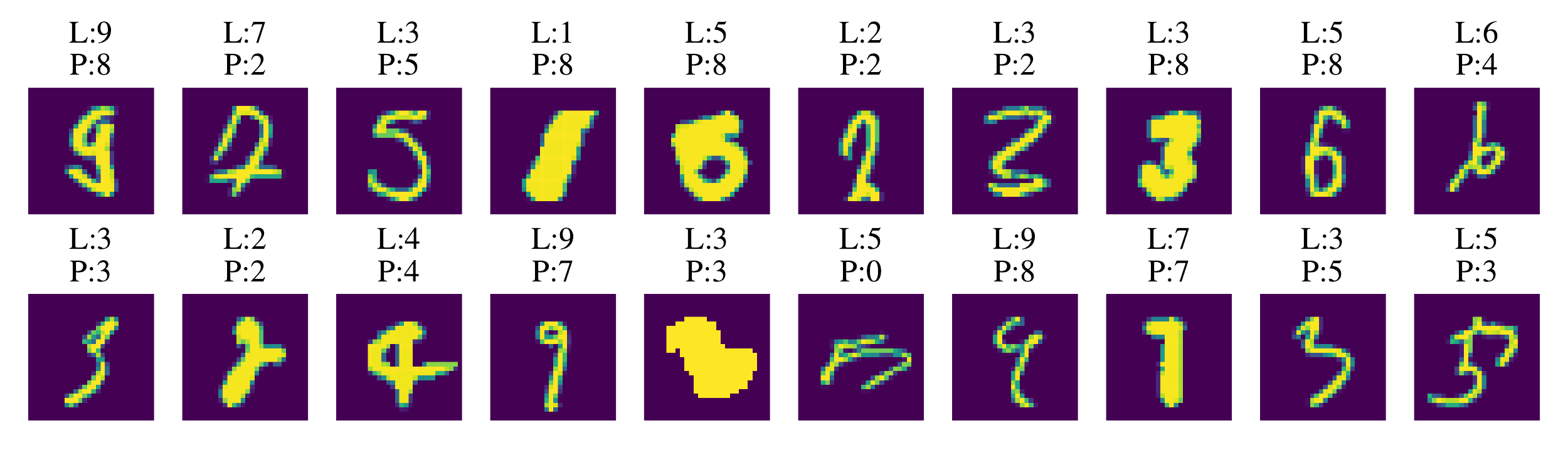} }}%

\centering
    \subfloat[\centering MNIST-LeNet Most-improved Examples. Similar to the most-worsened examples, such examples are atypical representation of labels, and may contain wrong or ambiguous labels.]{{\includegraphics[width=\linewidth]{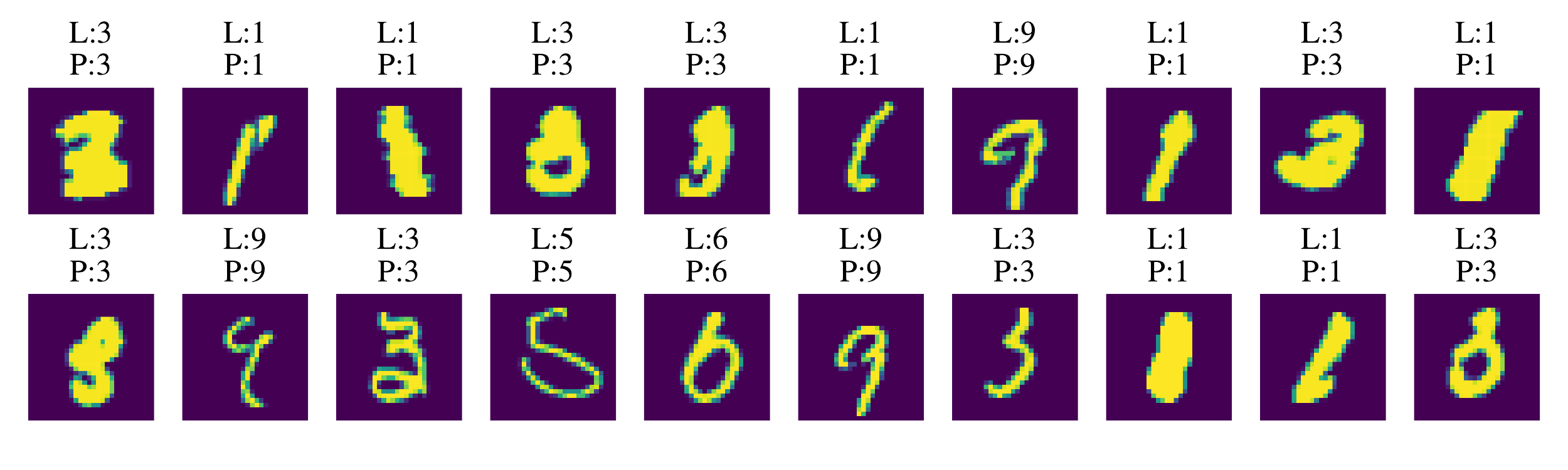} }}%

\centering
    \subfloat[\centering MNIST-LeNet Least-affected Examples.]{{\includegraphics[width=\linewidth]{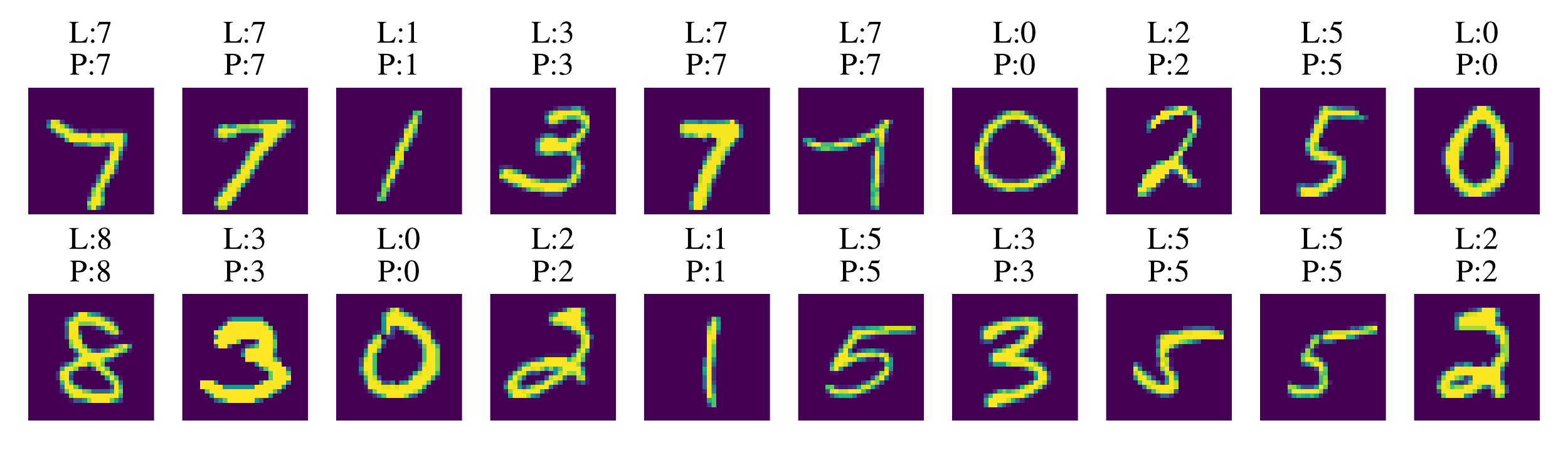} }}%

\centering
    \subfloat[\centering CIFAR-ResNet20 Most-worsened Examples.
    The 1st image on the 1st row, the 5th image on the 1st row and the 2nd image from the right on the 1st row contains multiple objects with the same label.
    The 1st image from the right on the 1st row is labeled truck, but cannot be discerned from an automobile when only its front is shown.]{{\includegraphics[width=\linewidth]{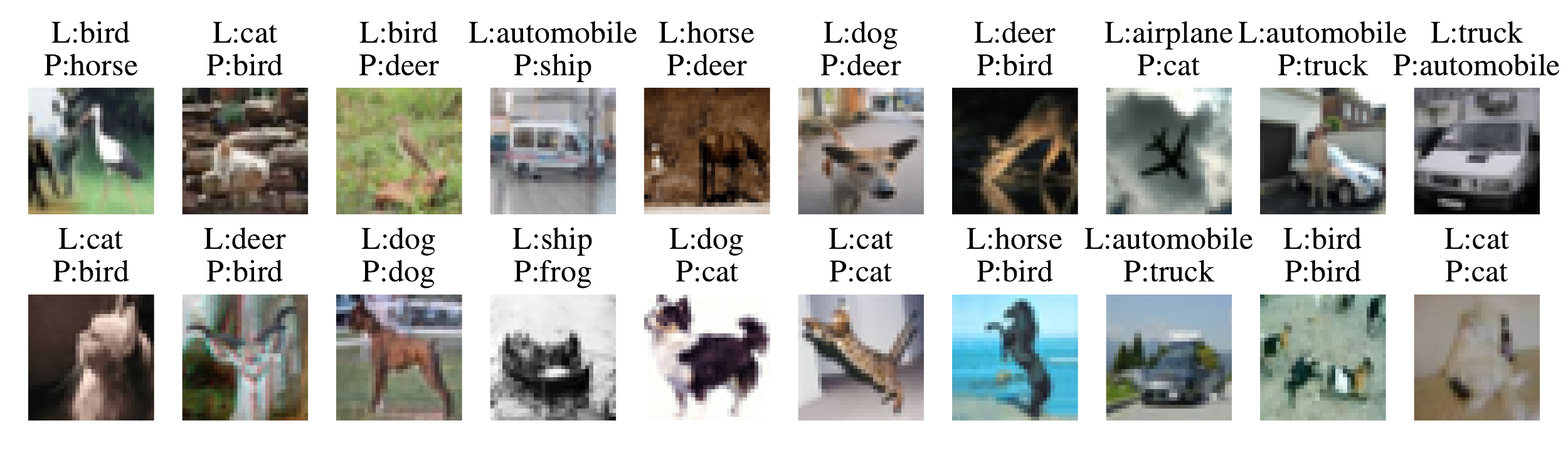} }}%

\caption{Most-worsened/improved examples are atypical representations of labels and may have wrong and ambiguous labels.
Least-affected examples are unambiguous and canonical representations of labels.
Plot title shows image label (denoted L) and model majority prediction (P).}
\label{fig:most-worsened-examples}
\end{figure*}

\begin{figure*}[h!]
\ContinuedFloat
\centering
    \subfloat[\centering CIFAR-ResNet20 Most-improved Examples. Similar to the most-worsened examples, such examples are atypical representation of labels and may contain wrong or ambiguous labels.]{{\includegraphics[width=\linewidth]{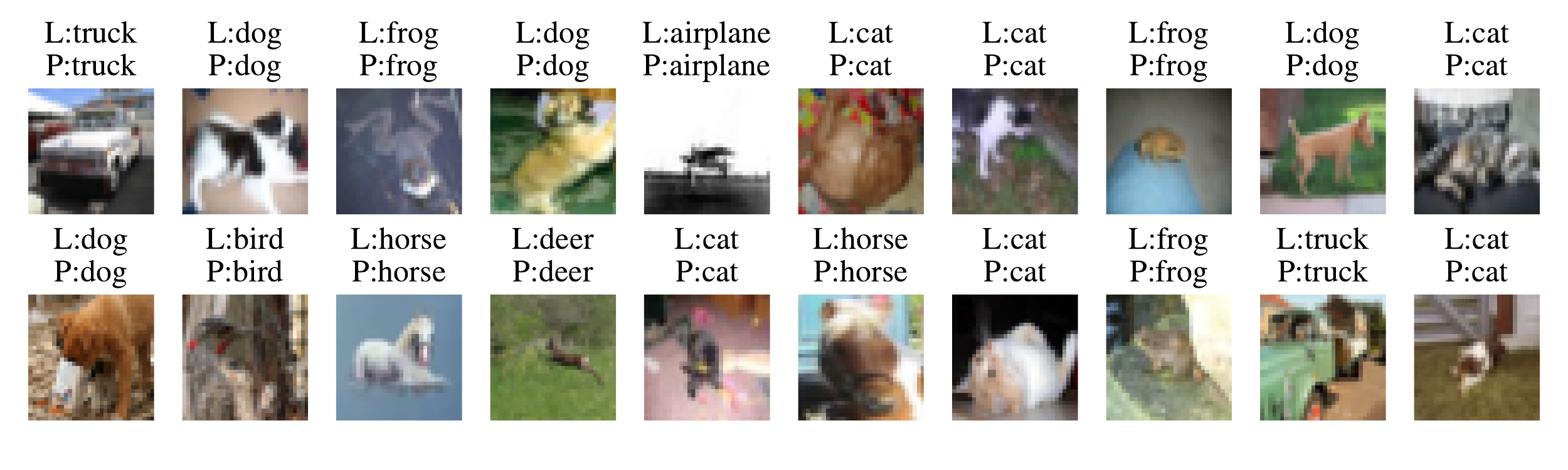} }}%

\centering
    \subfloat[\centering CIFAR-ResNet20 Least-affected Examples.]{{\includegraphics[width=\linewidth]{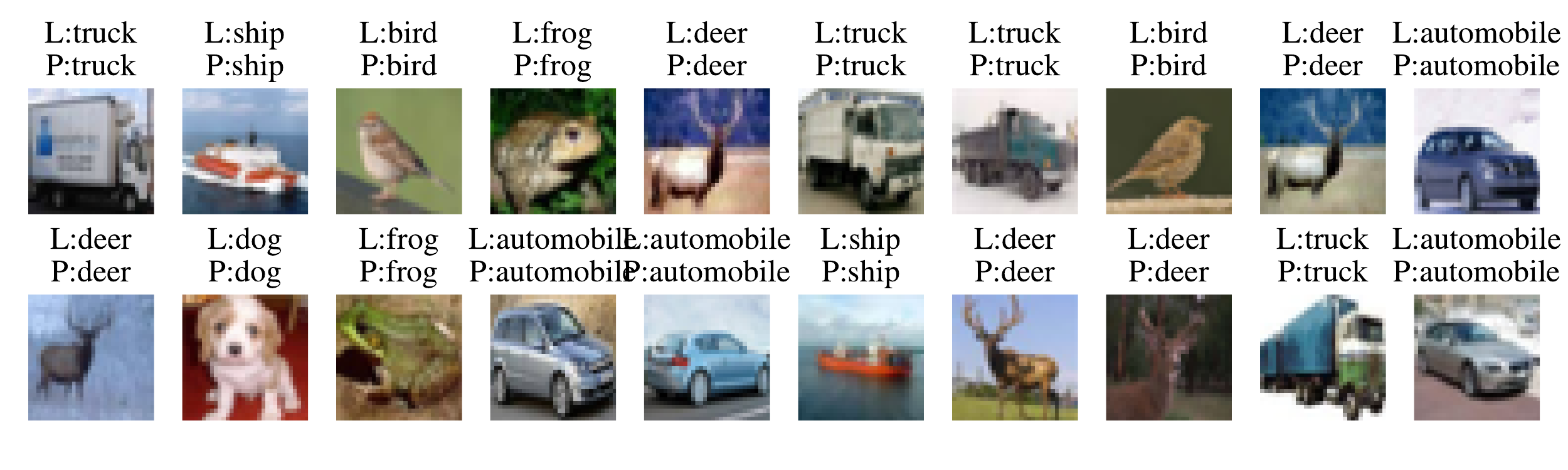} }}%

\centering
    \subfloat[\centering CIFAR-VGG-16 Most-affected Examples.
    The 5th image on the 1st row contains multiple objects with the same label.
    The 2nd image on the 1st row labeled bird does not contain enough information to be discerned from an airplane.
    The 2nd image from the right on the 1st row is quite blurred, but judging by the fact that horses are more likely to graze than dogs, the animal in this picture is likely to be a horse.
    ]{{\includegraphics[width=\linewidth]{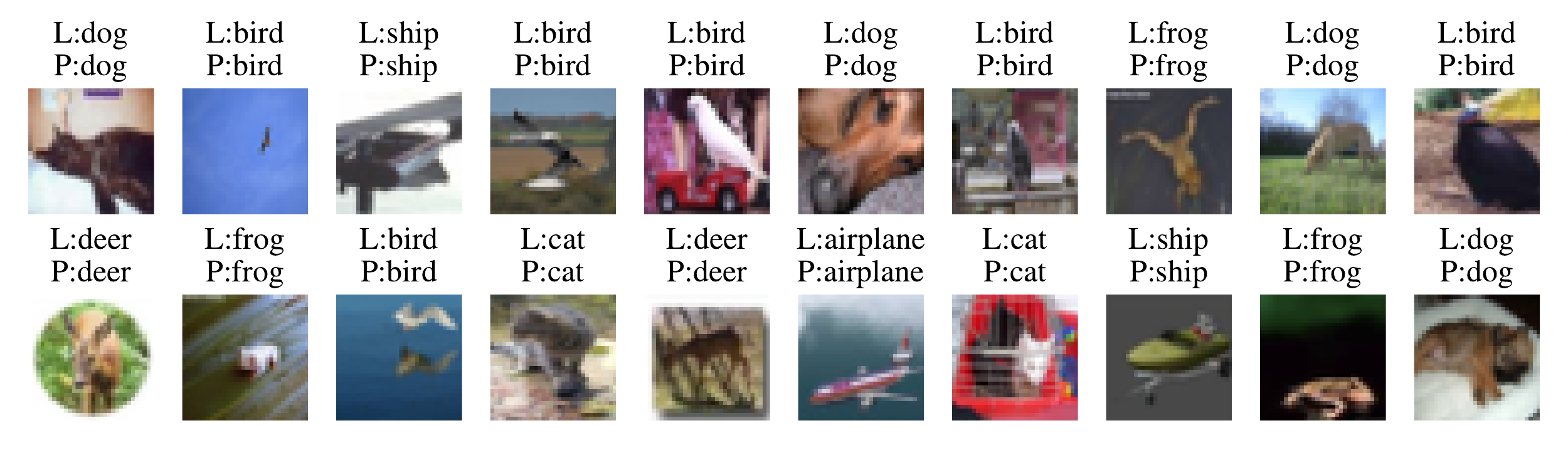} }}%

\centering
    \subfloat[\centering CIFAR-VGG-16 Most-improved Examples. Similar to the most-worsened examples, such examples are atypical representation of labels and may contain wrong or ambiguous labels..
    ]{{\includegraphics[width=\linewidth]{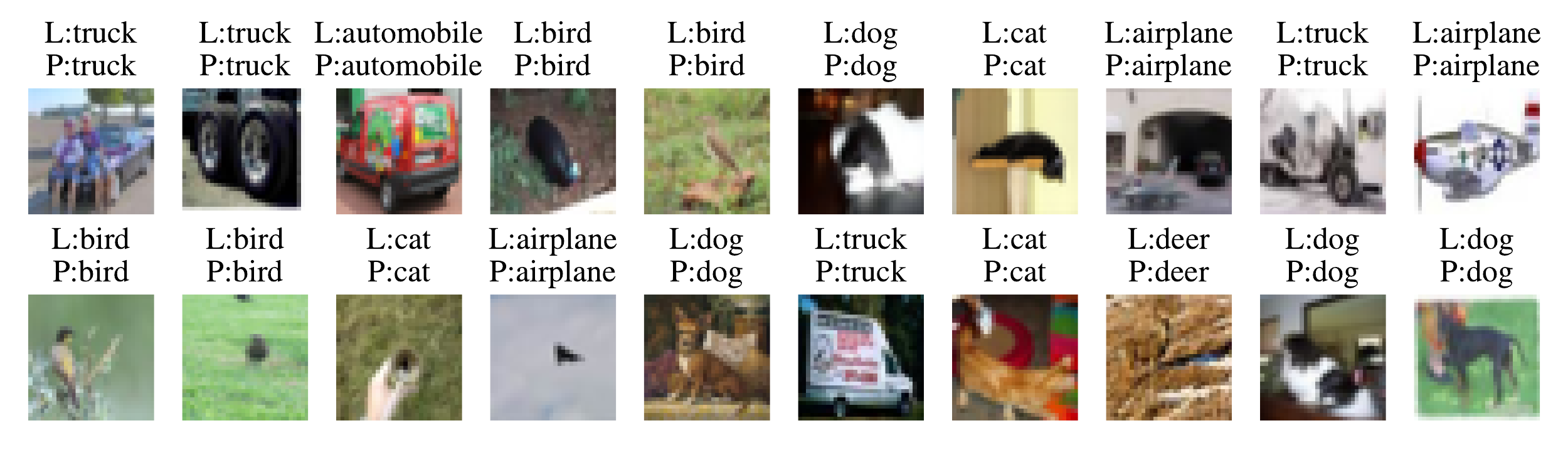} }}%

\caption{(Cont.) Most-worsened/improved examples are atypical representations of labels and may have wrong and ambiguous labels.
Least-affected examples are unambiguous and canonical representations of labels.
Plot title shows image label (denoted L) and model majority prediction (P).}
\label{fig:most-worsened-examples}
\end{figure*}

\begin{figure*}[h!]
\ContinuedFloat
\centering
    \subfloat[\centering CIFAR-VGG-16 Least-affected Examples.
    ]{{\includegraphics[width=\linewidth]{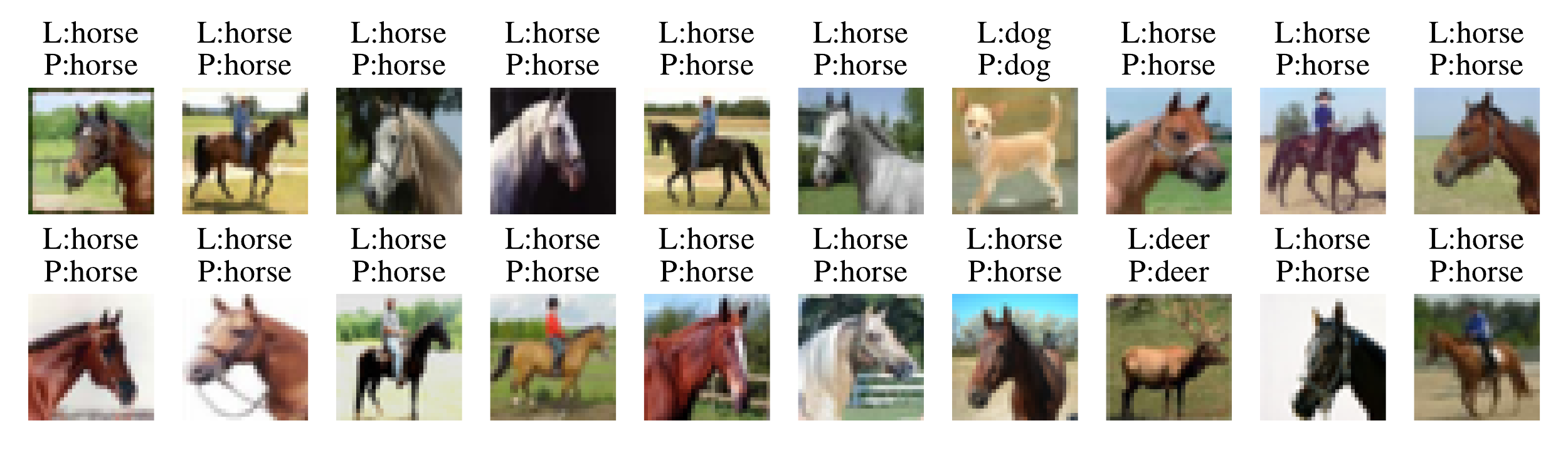} }}%

\centering
    \subfloat[\centering CIFAR-100-ResNet32 Most-affected Examples.
    The 2nd image on the first row labeled cloud has the wrong label.
    The 3rd image from the right on the 2nd row labeled table has more than one objects with the same label, it also has fruits on it.]{{\includegraphics[width=\linewidth]{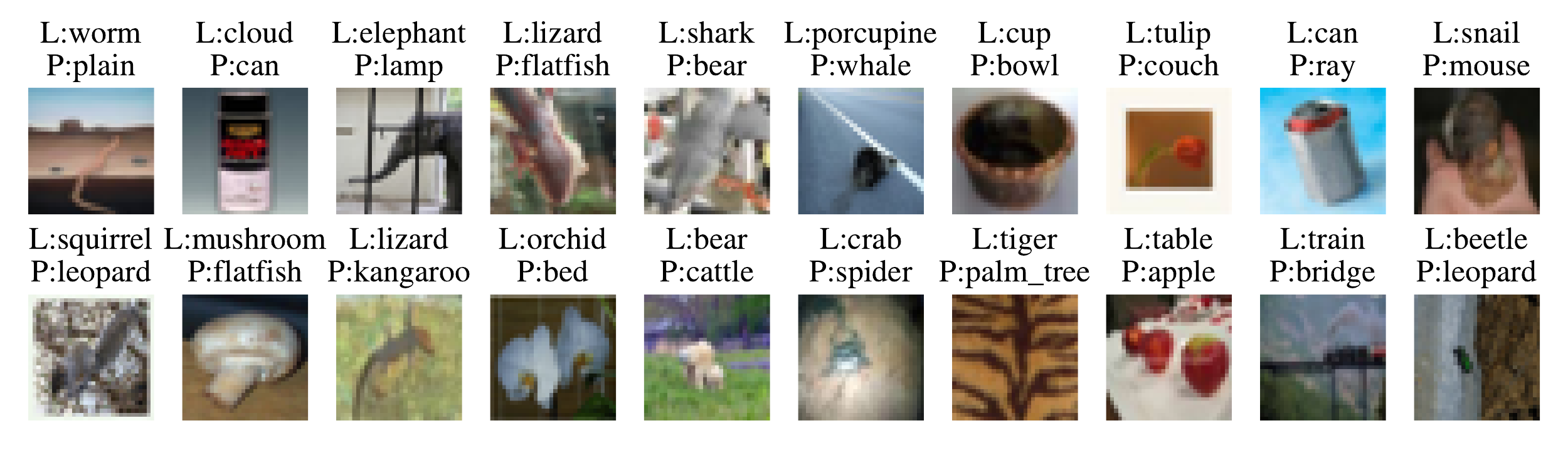} }}%

\centering
    \subfloat[\centering CIFAR-100-ResNet32 Most-improved Examples.
    Similar to the most-worsened examples, such examples are atypical representation of labels and may contain wrong or ambiguous labels.
    ]{{\includegraphics[width=\linewidth]{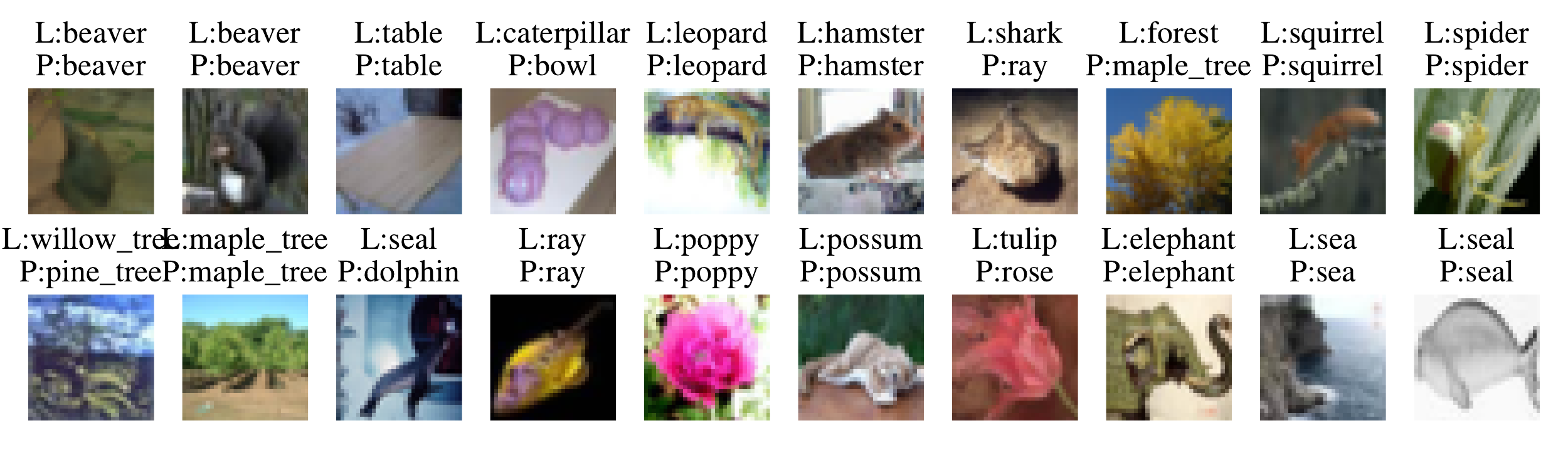} }}%

\centering
    \subfloat[\centering CIFAR-100-ResNet32 Least-affected Examples.
    ]{{\includegraphics[width=\linewidth]{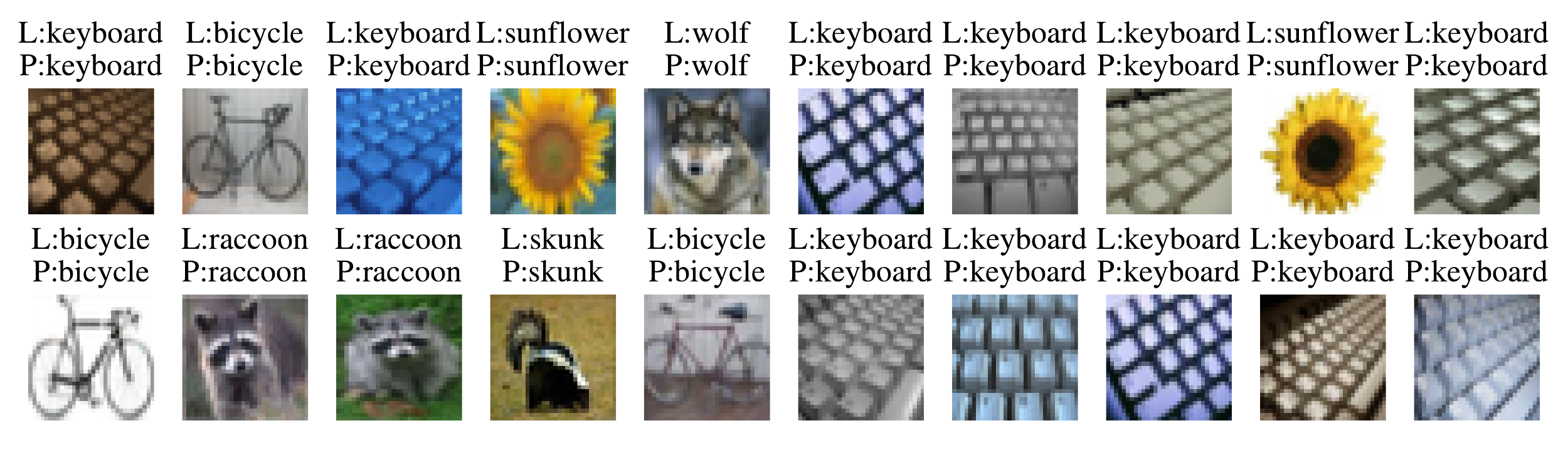} }}%

\caption{(Cont.) Most-worsened/improved examples are atypical representations of labels and may have wrong and ambiguous labels.
Least-affected examples are unambiguous and canonical representations of labels.
Plot title shows image label (denoted L) and model majority prediction (P).}
\label{fig:most-worsened-examples}
\end{figure*}

\begin{figure*}[h!]
\ContinuedFloat
\centering
    \subfloat[\centering ImageNet-ResNet50 Most-affected Examples.
    The 3rd image on the 1st row is not a limpkin, which is a bird species with long beak.
    The 1st image from the right on the 2nd row is not a tusker, but a wild boar.
    The 5th image from the right labeled guinea pig, the 2nd, 3rd and 4th images from the right on the 2nd row, among other images, show multiple objects with the same label.]{{\includegraphics[width=\linewidth]{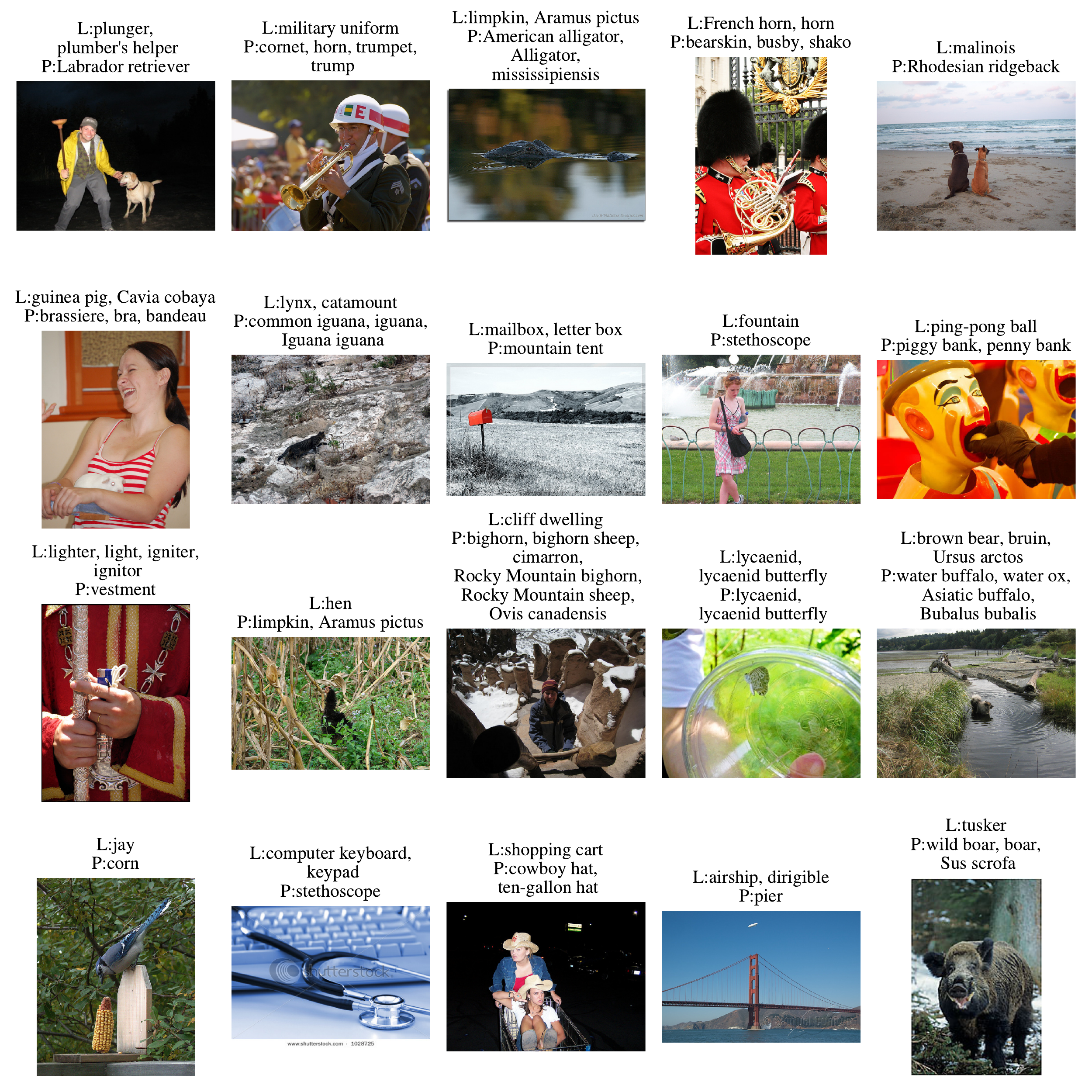} }}%
\caption{(Cont.) Most-worsened/improved examples are atypical representations of labels and may have wrong and ambiguous labels.
Least-affected examples are unambiguous and canonical representations of labels.
Plot title shows image label (denoted L) and model majority prediction (P).}
\label{fig:most-worsened-examples}
\end{figure*}

\begin{figure*}[h!]
\ContinuedFloat
    \subfloat[\centering ImageNet-ResNet50 Most-improved Examples.
    Similar to the most-worsened examples, such examples are atypical representation of labels and may contain wrong or ambiguous labels.]{{\includegraphics[width=\linewidth]{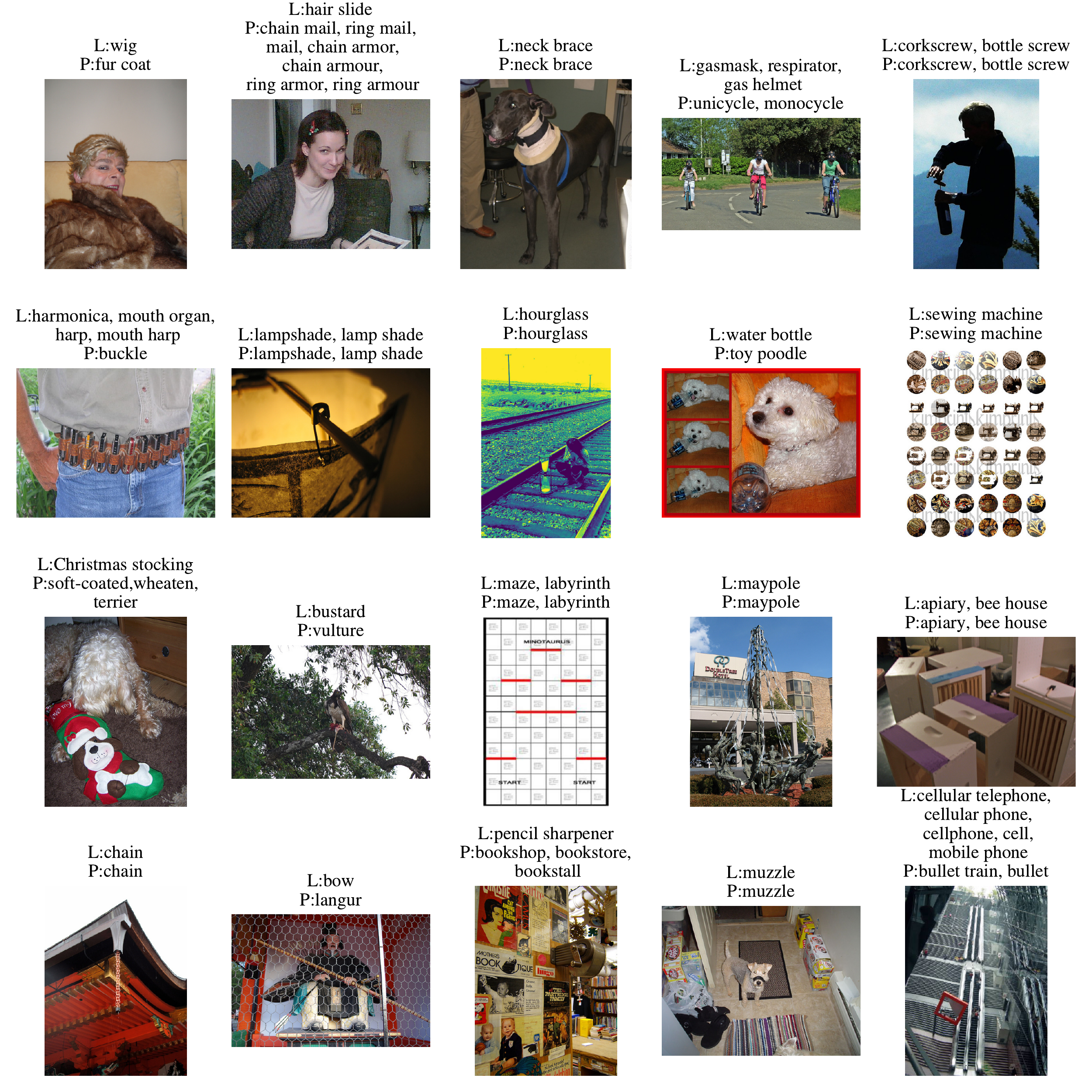} }}%
\caption{(Cont.) Most-worsened/improved examples are atypical representations of labels and may have wrong and ambiguous labels.
Least-affected examples are unambiguous and canonical representations of labels.
Plot title shows image label (denoted L) and model majority prediction (P).}
\label{fig:most-worsened-examples}
\end{figure*}

\begin{figure*}[h!]
\ContinuedFloat
    \subfloat[\centering ImageNet-ResNet50 Least-affected Examples.]{{\includegraphics[width=\linewidth]{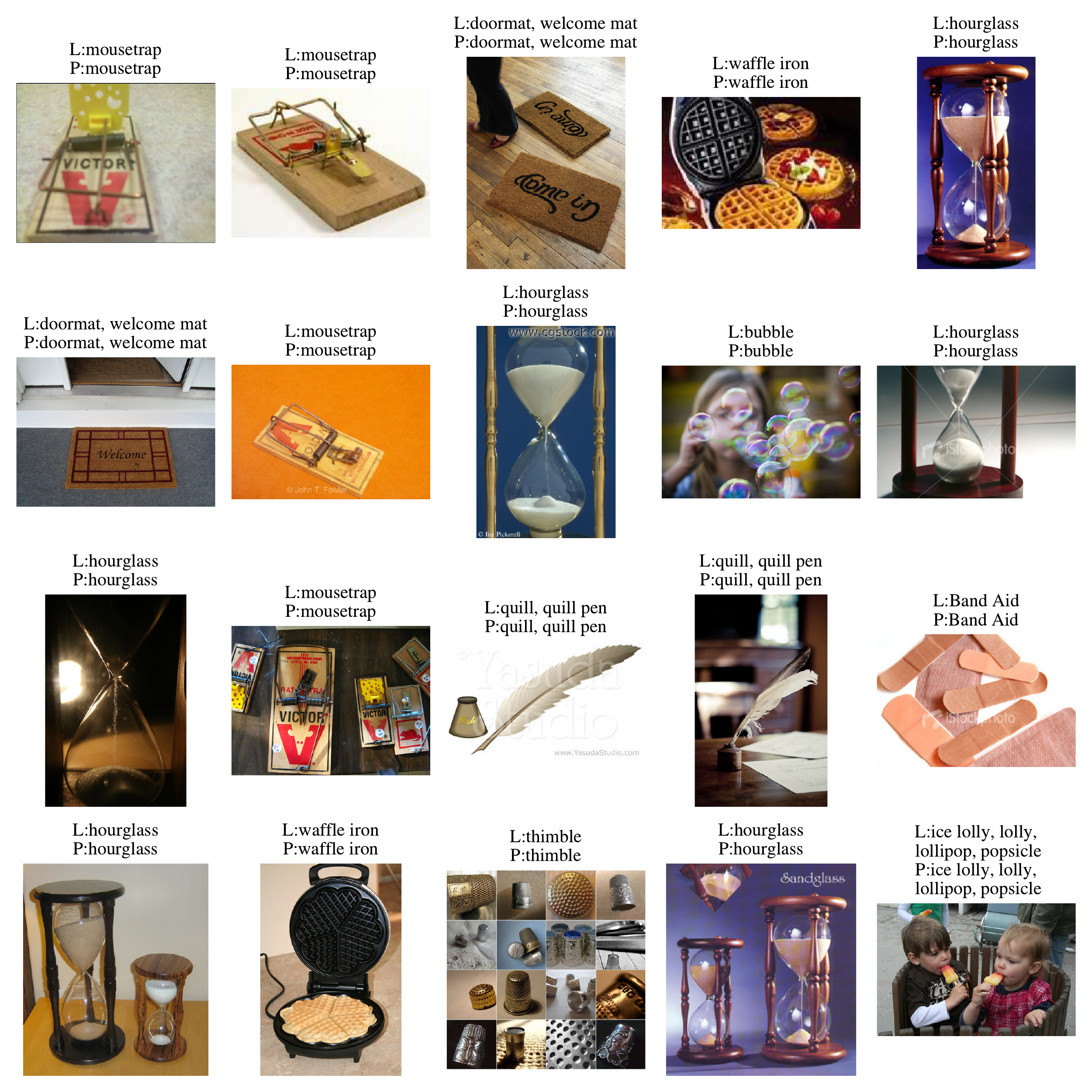} }}%
\caption{(Cont.) Most-worsened/improved examples are atypical representations of labels and may have wrong and ambiguous labels.
Least-affected examples are unambiguous and canonical representations of labels.
Plot title shows image label (denoted L) and model majority prediction (P).}
\label{fig:most-worsened-examples}
\end{figure*}

\clearpage

\subsection{Empirical Analysis}
\label{subsec:empirical-analysis}
We refer to the set of examples whose training loss is increased the most by pruning as \textit{top-worsened examples}.
In this subsection, we empirically evaluate the effect of the top-worsened examples on generalization.
We focus on two benchmarks -- CIFAR-10-ResNet20 and CIFAR-10-VGG-16, since we only observe generalization improvement with an overall \emph{increase} in training loss on the two aforementioned benchmarks.

\paragraph{Method.}
For each benchmark and model with sparsity $X\%$ that the pruning algorithm generates, we create two dataset subsamples of size $N$ to evaluate the effect of $P$ top-worsened examples on generalization, where $N \gg P$:
a). start with a randomly drawn $(N-P)$ examples, and add additional $P$ examples whose training loss is increased the most by pruning to sparsity $X$ (i.e., the top-worsened examples);
we denote this dataset subsample as $S_{TW}^X$;
b). start with the same $(N-P)$ examples, but add additional $P$ examples drawn randomly from the rest of the dataset;
we denote this dataset subsample as $S_{Rand}^X$.
We then train dense models on two subsamples $S_{TW}^X$ and $S_{Rand}^X$ to obtain the corresponding generalization $Y_{TW}^X$, $Y_{Rand}^X$.
If our hypothesis is correct -- that pruning improves generalization by increasing training loss, essentially ignoring, a small fraction of noisy examples detrimental to generalization -- for a range of sparsities $X$ where pruning improves generalization, we should observe that the generalization of dense models we train on the subsample containing top-worsened examples ($S_{TW}^X$) to be worse than the one we train on a randomly drawn subsample ($S_{Rand}^X$).
Following the precedent of  \citet{DBLP:journals/corr/abs-2107-07075}, we choose $N$ to be the size equivalent to $40\%$ of the training dataset and $P$ to be $1\%$ of the training dataset.

Notably, subsampling the dataset is necessary to reveal the effect of the small fraction of noisy examples empirically.
Their relatively rare occurrence makes their harmful effect on generalization difficult to detect empirically.
This is consistent with with observation and experimental setup used in \citet{DBLP:journals/corr/abs-2107-07075}, characterizing the harmful effect of a small fraction of training examples with high EL2N scores.
We similarly demonstrate the harmful effect of examples avoided by the sparse models.

\begin{figure}[h!]
\includegraphics[width=\linewidth]{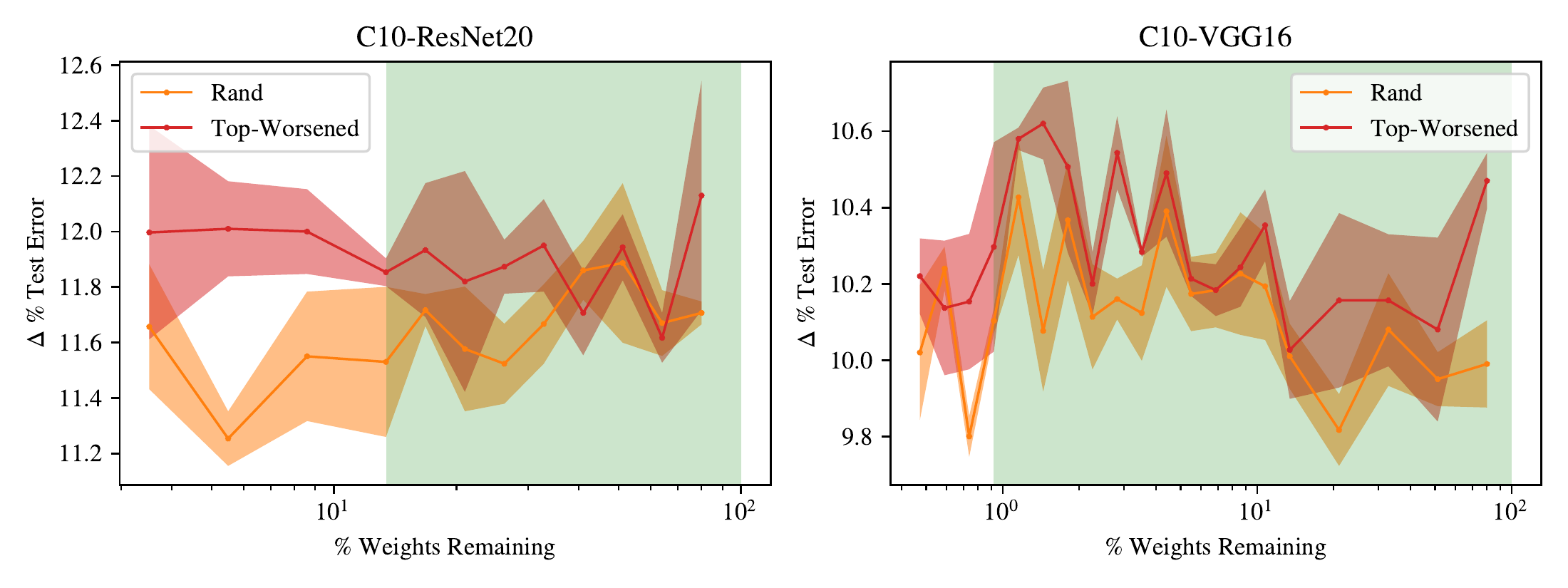}
\caption{Dense models we train on dataset subsamples including top-worsened examples (labeled Top-Worsened)  achieve worse generalization than dense models we train on dataset subsamples without top-worsened examples (labeled Rand).
We show test errors of dense models trained on two dataset subsamples as a function of sparsities at which per-example training loss is measured.
We color the range of sparsity levels where pruning improves generalization green.}%
\label{fig:most-pershing-examples}%
\end{figure}

\paragraph{Results.} We visualize the generalization difference $Y_{TW}^X$ - $Y_{Rand}^X$ of dense models we train on the two dataset subsamples as a function of sparsities $X$ in \Cref{fig:most-pershing-examples}.
Indeed, dense models we train on the subsample containing the top-worsened examples achieve worse generalization compared with dense models we train on a random subsample, most notably near and beyond the highest sparsity that still improves generalization.
Within the range of sparsity levels where pruning improves generalization, the test errors of dense ResNet20, VGG-16 models we train on the dataset subsample containing the top-worsened examples is worse than the models trained on a random subsample by 0 (matching) to 0.35\% and 0 (matching) to 0.54\% respectively.
We thus confirm that pruning improves generalization while avoiding fitting a small fraction of examples harmful to generalization.
\fTBD{Could iterate the experiments a bit more to get cleaner signal and comment quantitatively on the difference between $Y_rand$ and $Y_avoid$.}

\clearpage

\section{Effects of Learning Algorithms on Subgroup Training Loss}
\label{appx:aggregate-effect}

In \Cref{subsec:aggregate-effect}, we partition the training set into subgroups, each with a distinct EL2N score percentile range and present pruning's effect on the average training loss of example subgroups at two sparsities of interests.
In this section, we present pruning's effect on the average training loss of example subgroups at more sparsities.
We similarly present the effect of two other learning algorithms, extended dense training (\Cref{subsec:edt}) and width downscaling (\Cref{sec:size-reduction}), on subgroup training loss.

\paragraph{Method.}
Similar to \Cref{subsec:aggregate-effect}, to measure the change in training loss due to a particular learning algorithm, we partition the training set into $M$ subgroups, each with a different range of EL2N scores.
For each subgroup, we then compute the average change in training loss after we apply the learning algorithm, relative to the resulting training loss after training the dense model with standard hyperparameters as specified in \Cref{subsec:training-hyperparameters}, on examples in the subgroup.
A negative value indicates that the learning algorithm improves the subgroup training loss relative to training the dense model with standard hyperparameters.
We present the effect of pruning on subgroup training loss across a sequence of evenly spaced 5 sparsities in \Cref{fig:prunign-effect-on-model-fit-many-sparsity-levels-noisy}.
We present the effect of extended dense training and width downscaling on subgroup training loss across a sequence of 5 evenly spaced training time milestones and model downscaling factors in \Cref{fig:edt-effect-on-model-fit-many-sparsity-levels-noisy} and  \Cref{fig:width-downscaling-effect-on-model-fit-many-sparsity-levels-noisy}, respectively.
Consistent with \Cref{subsec:aggregate-effect}, we pick $M=20$ because it is the largest value of $M$ that enables us to clearly present per-subgroup training loss change.

\begin{figure*}[h!]
\subfloat[LeNet-MNIST, rows correspond to 0\%, 5\%, 10\% and 15\% random label noise]{
   \includegraphics[scale=0.65]{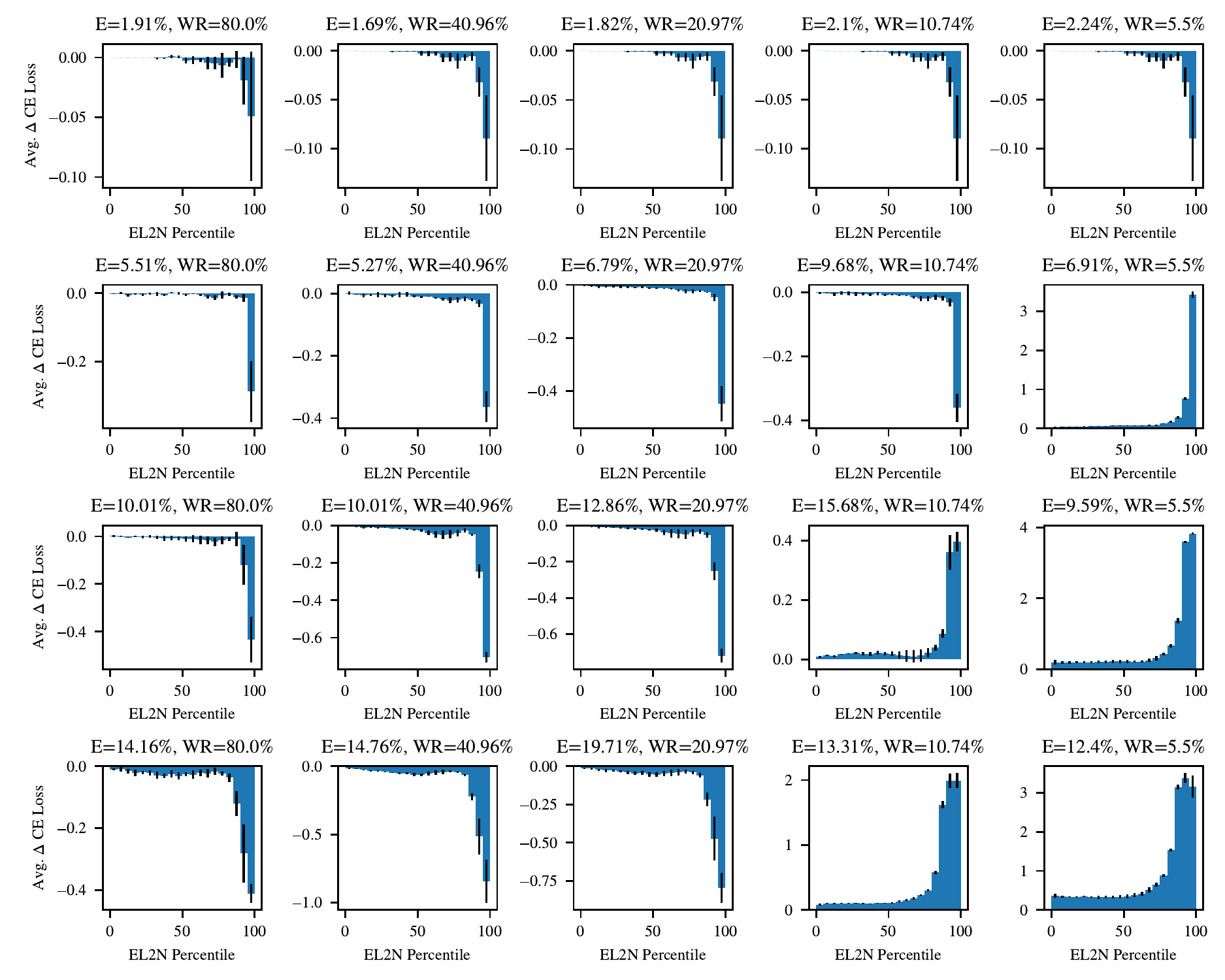}
}

\subfloat[CIFAR-10-ResNet20, rows correspond to 0\%, 15\%, 30\% and 60\% random label noise]{
	\includegraphics[scale=0.65]{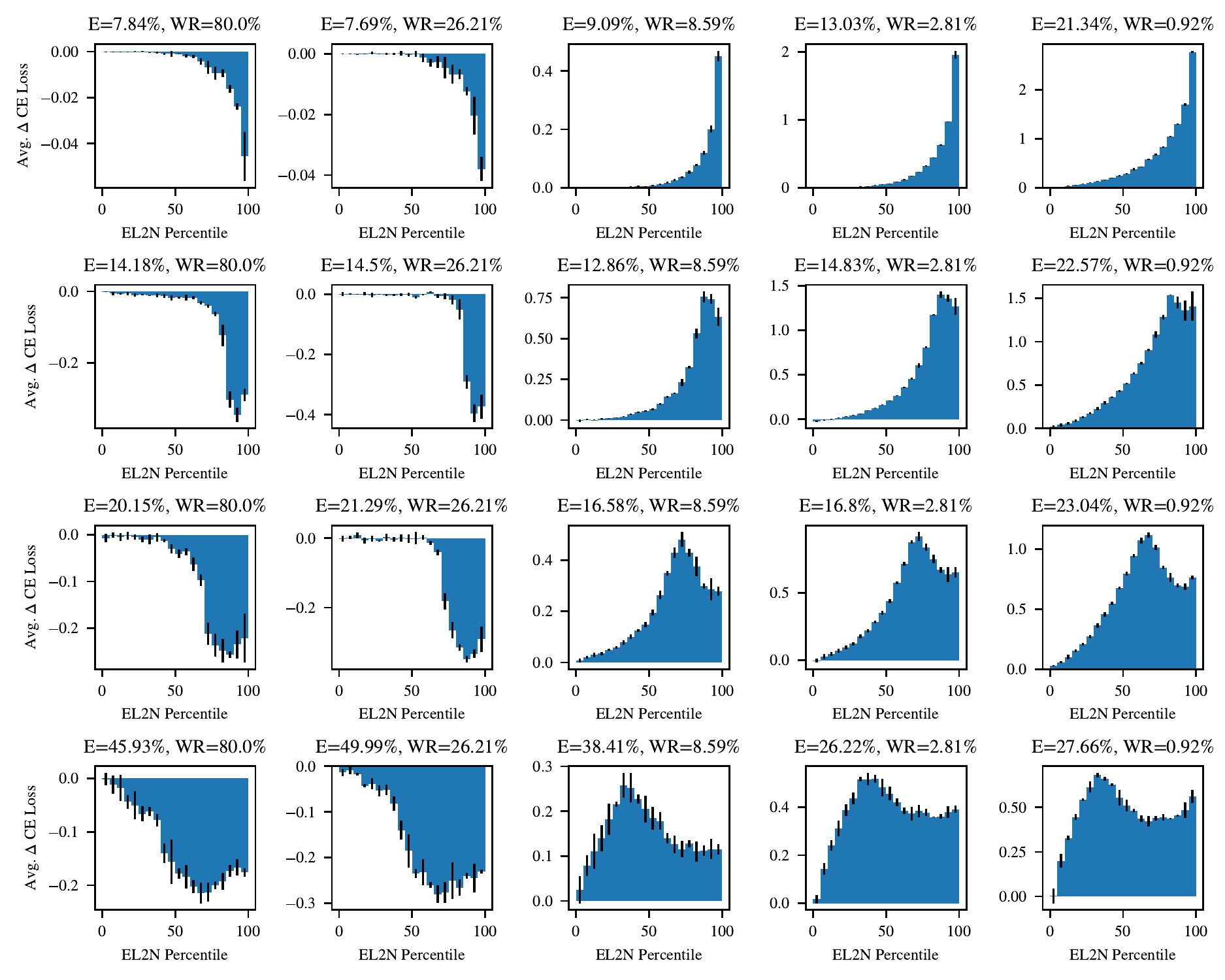}
}
\caption{Pruning's effect on average training loss on subgroups of examples with distinct EL2N score percentile range.
Title shows the test error (E) and weights remaining (WR) of the sparse model.
A negative value indicates that pruning improves training loss.}
\label{fig:prunign-effect-on-model-fit-many-sparsity-levels-noisy}
\end{figure*}

\begin{figure*}[h!]
\ContinuedFloat
\subfloat[CIFAR-10-VGG-16, rows correspond to 0\%, 15\%, 30\% and 60\% random label noise]{
	\includegraphics[scale=0.65]{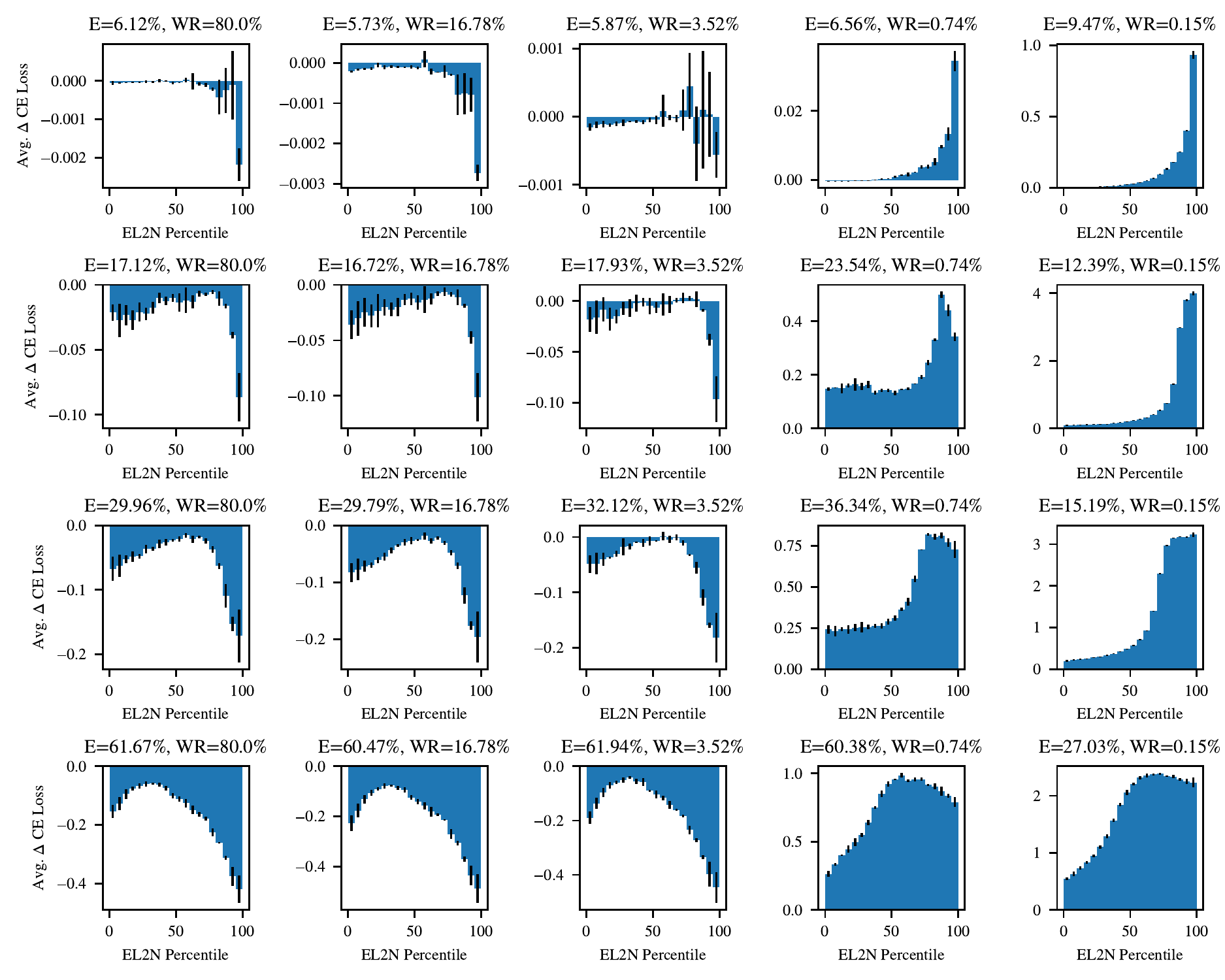}
}

\subfloat[CIFAR-100-ResNet32, rows correspond to 0\%, 15\%, 30\% and 60\% random label noise]{
	\includegraphics[scale=0.65]{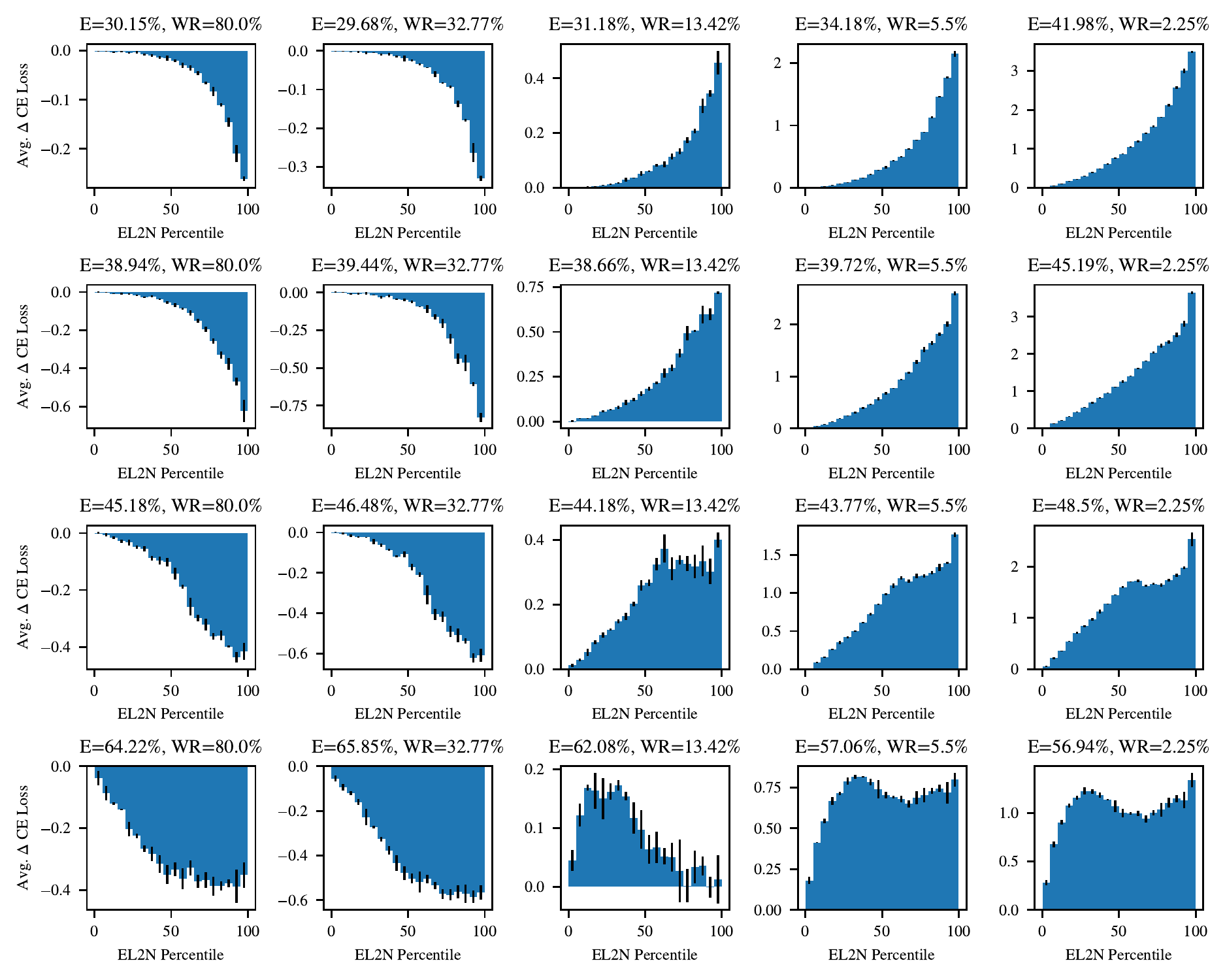}
}

\caption{(Cont.) Pruning's effect on average training loss on subgroups of examples with distinct EL2N score percentile range.
Title shows the test error (E) and weights remaining (WR) of the sparse model.
A negative value indicates that pruning improves training loss.}
\label{fig:prunign-effect-on-model-fit-many-sparsity-levels-noisy}
\end{figure*}

\begin{figure*}[h!]
\ContinuedFloat
\subfloat[ImageNet-ResNet50, with 0\%, 15\% random label noise]{
	\includegraphics[scale=0.65]{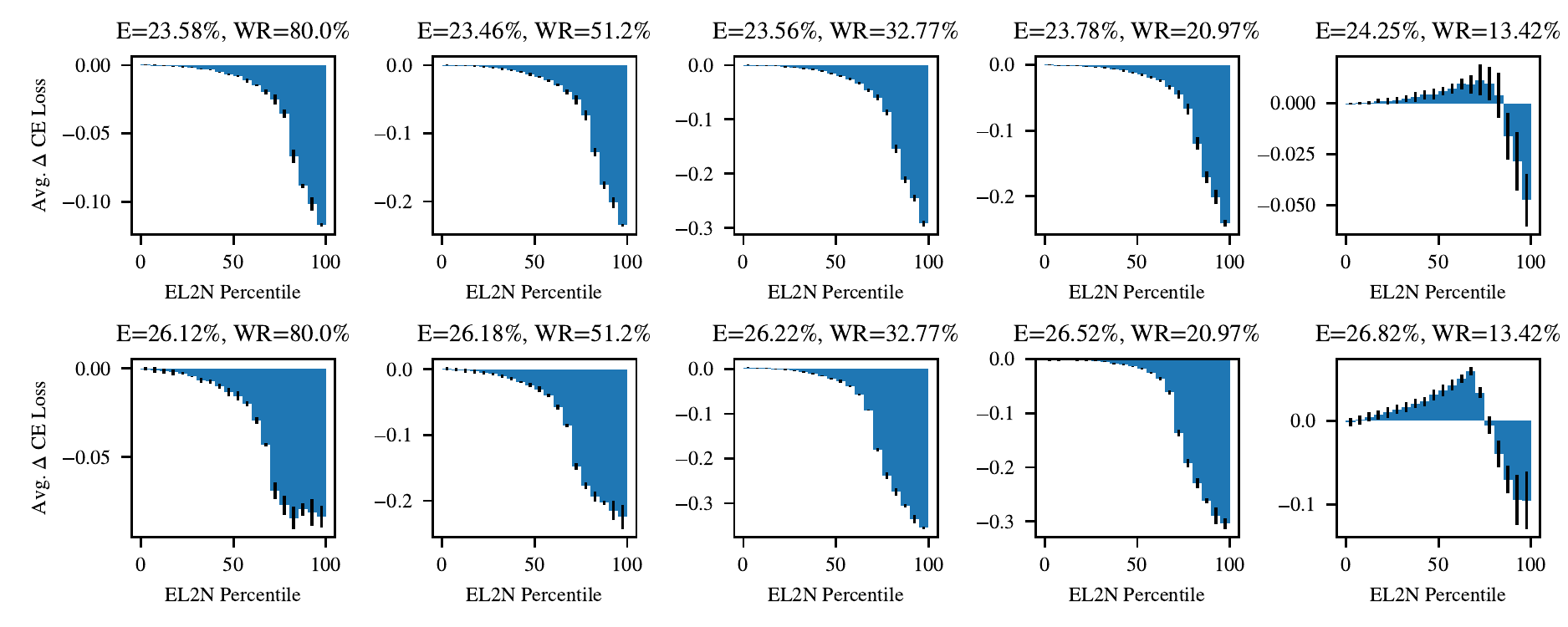}
}
\caption{(Cont.) Pruning's effect on average training loss on subgroups of examples with distinct EL2N score percentile range.
Title shows the test error (E) and weights remaining (WR) of the sparse model.
A negative value indicates that pruning improves training loss.}
\label{fig:prunign-effect-on-model-fit-many-sparsity-levels-noisy}
\end{figure*}

\paragraph{Pruning results.}
\Cref{fig:prunign-effect-on-model-fit-many-sparsity-levels-noisy} shows that, at low sparsities, pruning's effect is to improve training loss on almost all example subgroups.
For example, on CIFAR-10-ResNet20 benchmark with 0\% random label noise, pruning to 80\% to 26.21\% weights remaining reduces the average training loss on example subgroups.
At these sparsities, pruning's effect is to improve training.
Pruning to higher sparsities, pruning's regularization effect dominates -- pruning increases training loss across all example subgroups.
For example, on CIFAR-10-ResNet20 benchmark with 8.59\% to 0.92\% weights remaining, pruning's effect is to increase the average training loss on almost all example subgroups.

\paragraph{Extended dense training results.}
\Cref{fig:edt-effect-on-model-fit-many-sparsity-levels-noisy} shows that the effect of extended dense training is to improve training loss on almost all example subgroups.
This effect is similar to pruning to low sparsities (e.g., 80\% to 26.21\% weights remaining for CIFAR-10-ResNet20), as shown in \Cref{fig:prunign-effect-on-model-fit-many-sparsity-levels-noisy}.
However, unlike pruning to high sparsities (e.g, 8.59\% to 0.92\% weights remaining for CIFAR-10-ResNet20), we do not observe any regularization effect of extended dense training.

\begin{figure*}[h!]
\subfloat[LeNet-MNIST, rows correspond to 0\%, 5\%, 10\% and 15\% random label noise]{
   \includegraphics[scale=0.65]{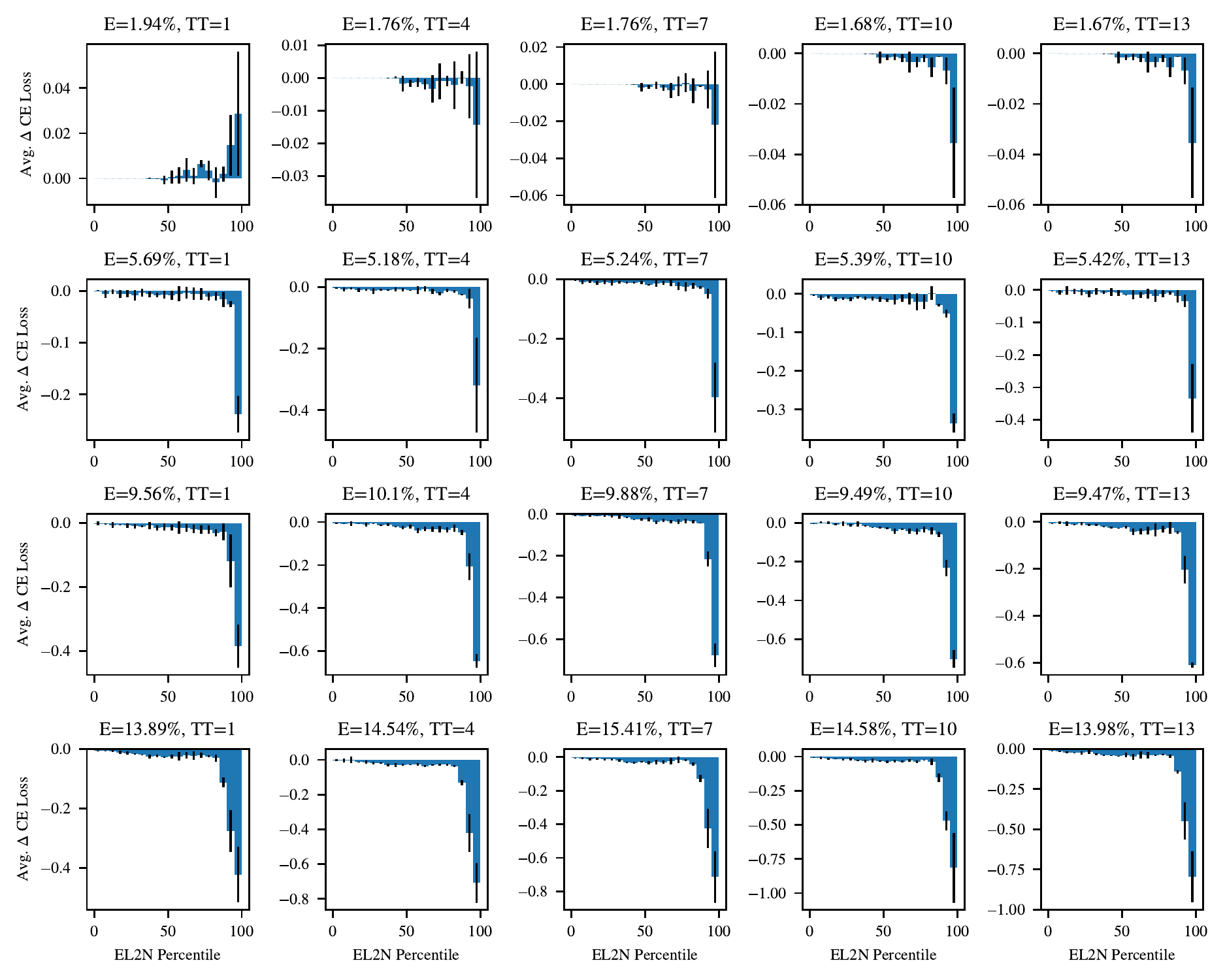}
}

\subfloat[CIFAR-10-ResNet20, rows correspond to 0\%, 15\%, 30\% and 60\% random label noise]{
	\includegraphics[scale=0.65]{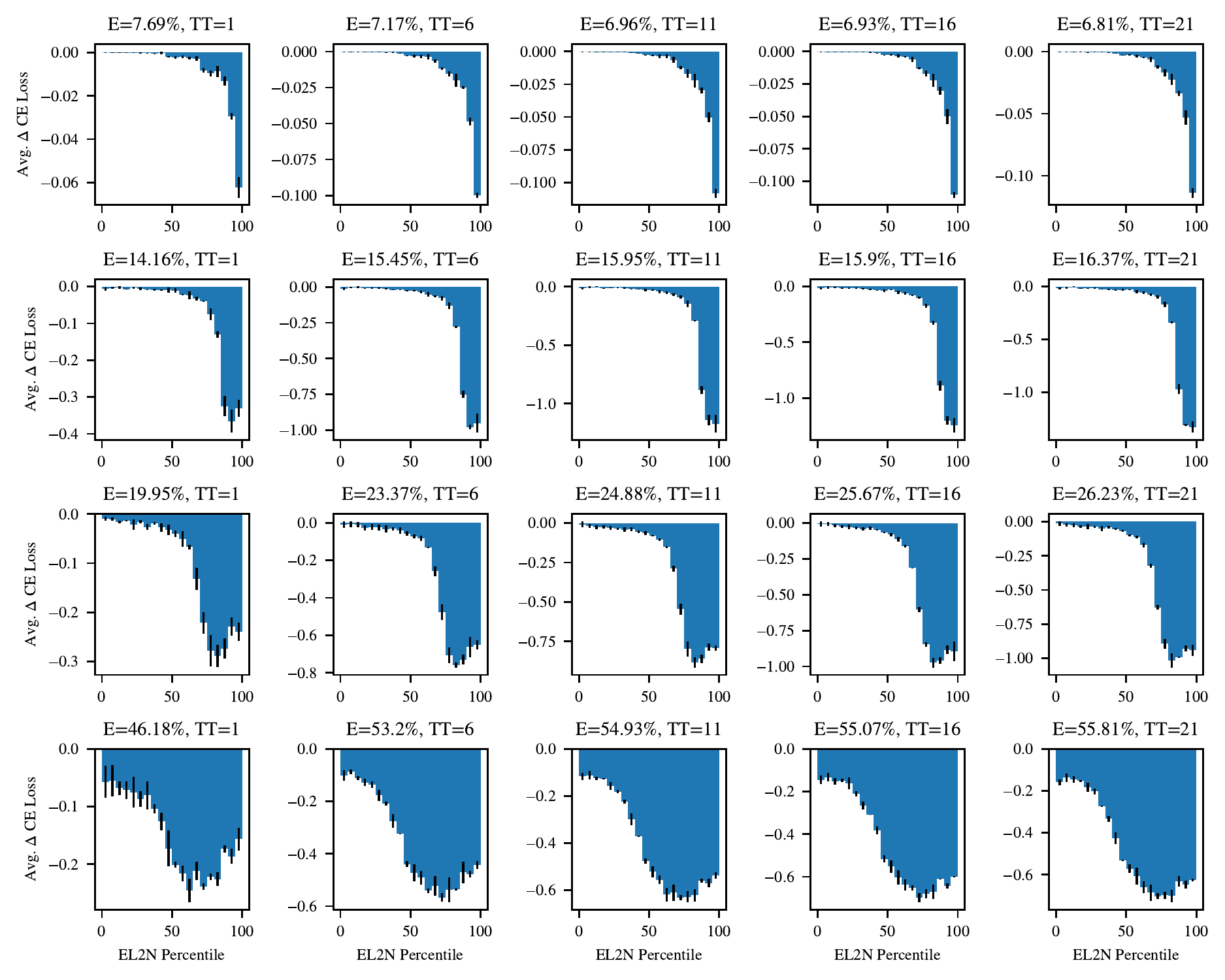}
}

\caption{Extended dense training improves subgroup training loss.
Title shows the test error (E) and training time (TT) in number of pruning iterations' worth of training epochs.
A negative value indicates that extended dense training improves training loss.}
\label{fig:edt-effect-on-model-fit-many-sparsity-levels-noisy}
\end{figure*}

\begin{figure*}[h!]
\ContinuedFloat
\subfloat[CIFAR-10-VGG-16, rows correspond to 0\%, 15\%, 30\% and 60\% random label noise]{
	\includegraphics[scale=0.65]{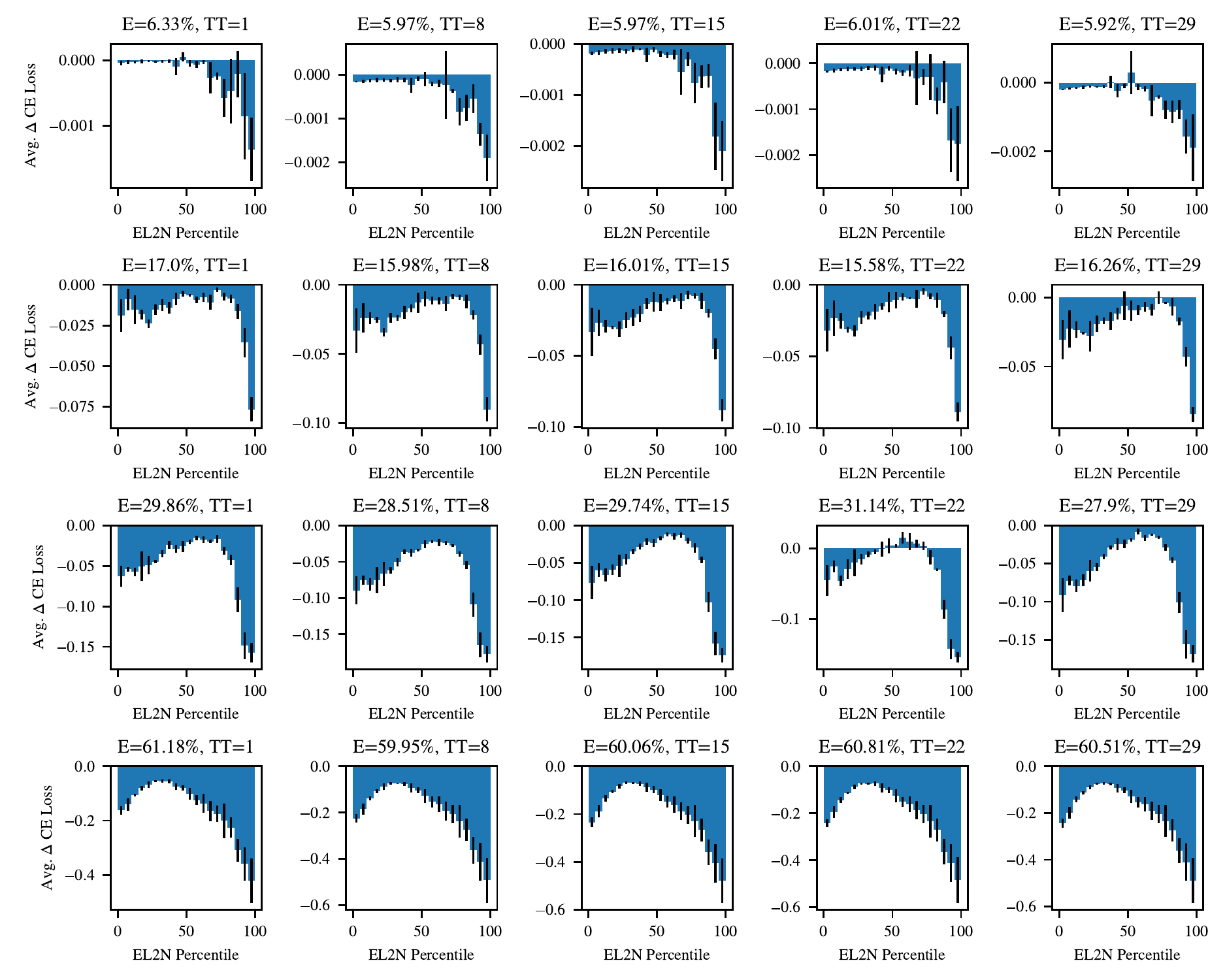}
}

\subfloat[CIFAR-100-ResNet32, rows correspond to 0\%, 15\%, 30\% and 60\% random label noise]{
	\includegraphics[scale=0.65]{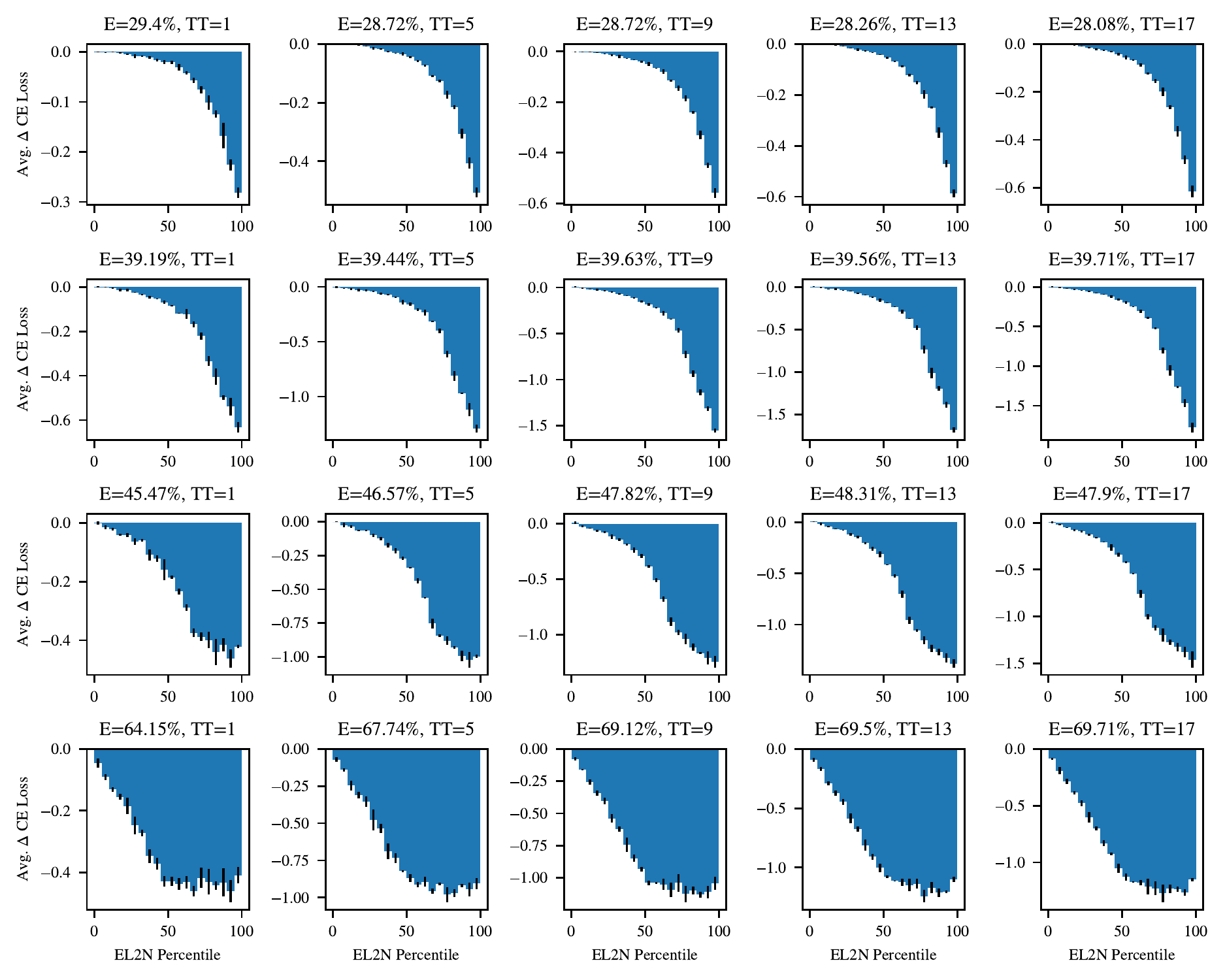}
}

\caption{(Cont.) Extended dense training improves subgroup training loss.
Title shows the test error (E) and training time (TT) in number of pruning iterations' worth of training epochs.
A negative value indicates that extended dense training improves training loss.}
\label{fig:edt-effect-on-model-fit-many-sparsity-levels-noisy}
\end{figure*}

\begin{figure*}[h!]
\ContinuedFloat
\subfloat[ImageNet-ResNet50, with 0\%, 15\% random label noise]{
	\includegraphics[scale=0.65]{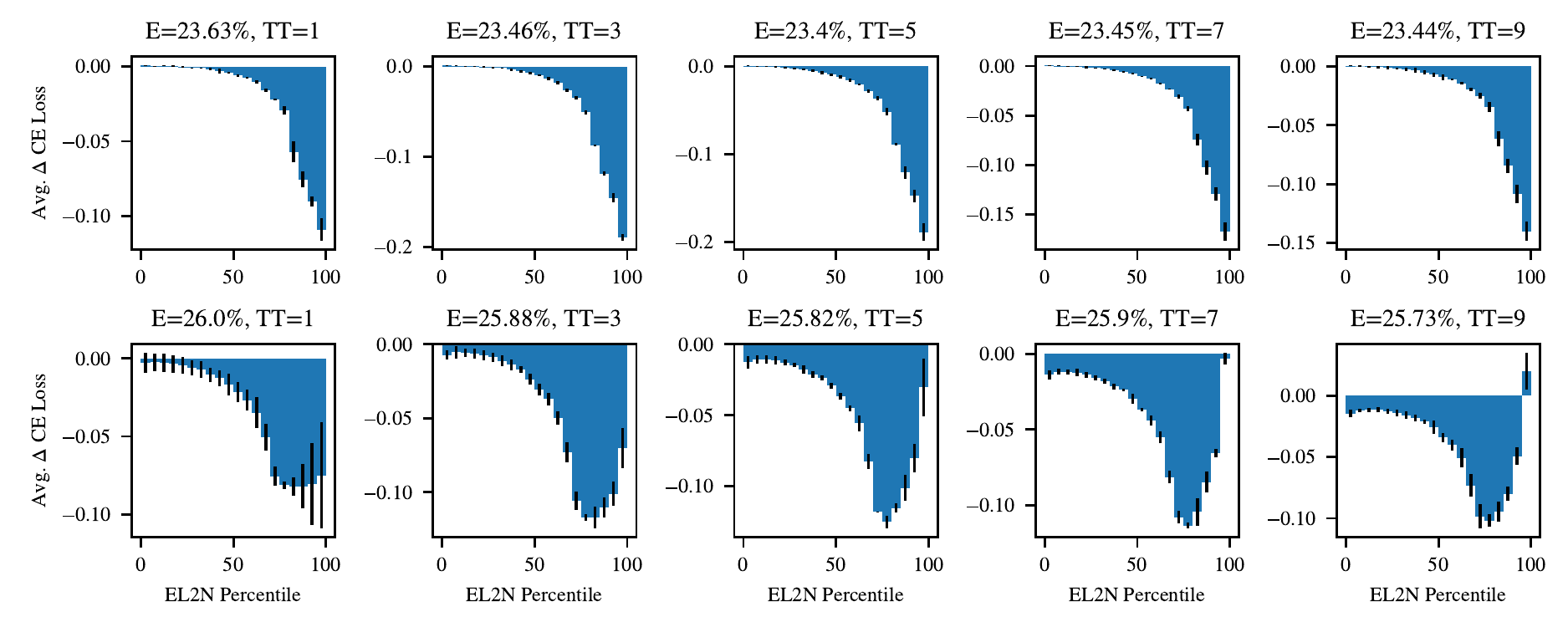}
}
\caption{(Cont.) Extended dense training improves subgroup training loss.
Title shows the test error (E) and training time (TT) in number of pruning iterations' worth of training epochs.
A negative value indicates that extended dense training improves training loss.}
\label{fig:edt-effect-on-model-fit-many-sparsity-levels-noisy}
\end{figure*}

\paragraph{Width downscaling results.}
\Cref{fig:width-downscaling-effect-on-model-fit-many-sparsity-levels-noisy} shows that width downscaling has a regularization effect, as it increases training loss on almost all example subgroups.
This effect is similar to pruning to high sparsities (e.g., 8.59\% to 0.92\% weights remaining for CIFAR-10-ResNet20), as shown in \Cref{fig:prunign-effect-on-model-fit-many-sparsity-levels-noisy}.
However, unlike pruning to low sparsities (e.g, 80\% to 26.21\% weights remaining for CIFAR-10-ResNet20), we do not observe any training loss improvement at any model width we test.

\begin{figure*}[h!]
\subfloat[LeNet-MNIST, rows correspond to 0\%, 5\%, 10\% and 15\% random label noise]{
   \includegraphics[scale=0.65]{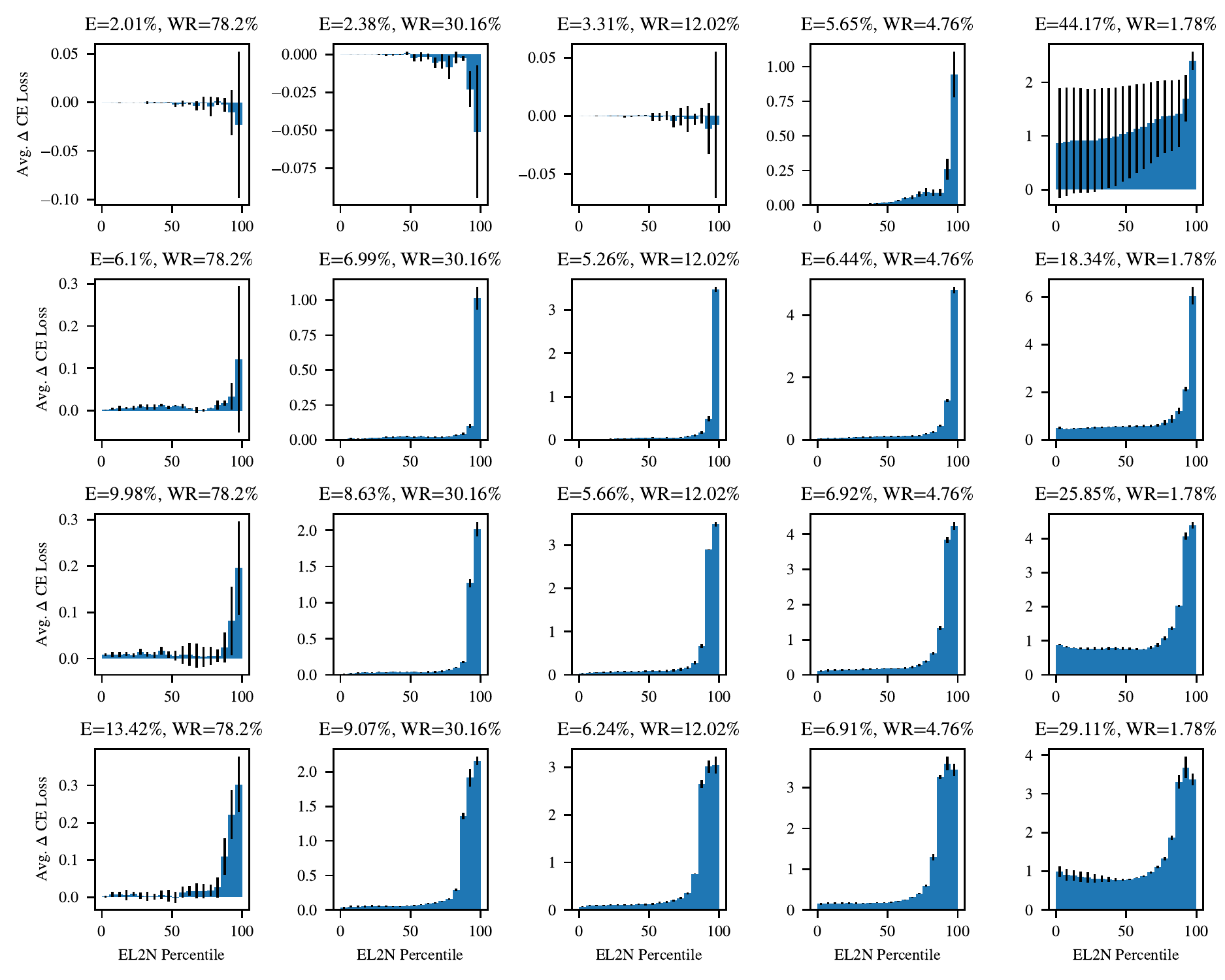}
}

\subfloat[CIFAR-10-ResNet20, rows correspond to 0\%, 15\%, 30\% and 60\% random label noise]{
	\includegraphics[scale=0.65]{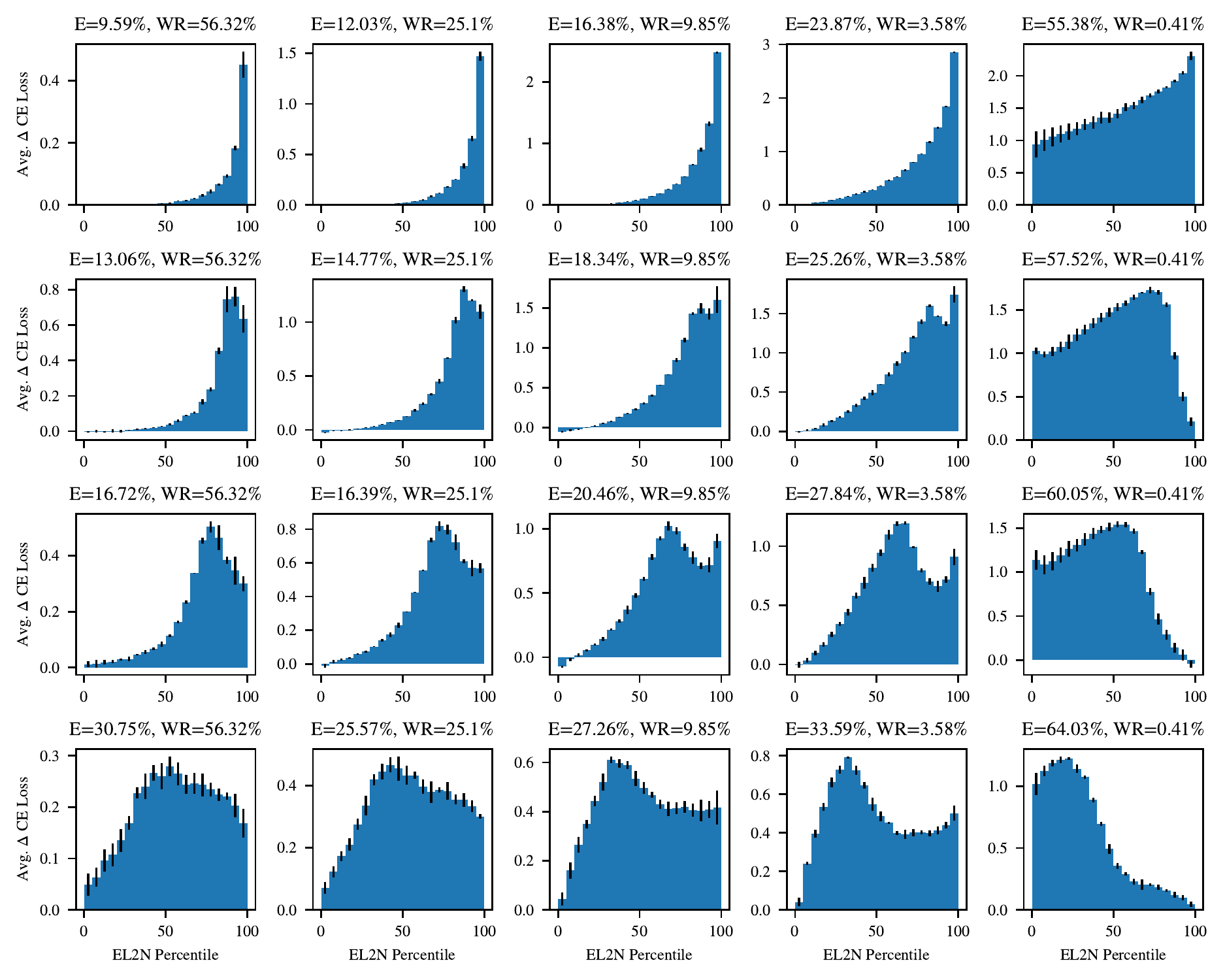}
}

\caption{Width downscaling increases subgroup training loss.
Title shows the test error (E) and weights remaining (WR), as a portion of the number of parameters in the model with the original and unscaled width.
A positive value indicates that width downscaling increases training loss.}
\label{fig:width-downscaling-effect-on-model-fit-many-sparsity-levels-noisy}
\end{figure*}

\begin{figure*}[h!]
\ContinuedFloat
\subfloat[CIFAR-10-VGG-16, rows correspond to 0\%, 15\%, 30\% and 60\% random label noise]{
	\includegraphics[scale=0.65]{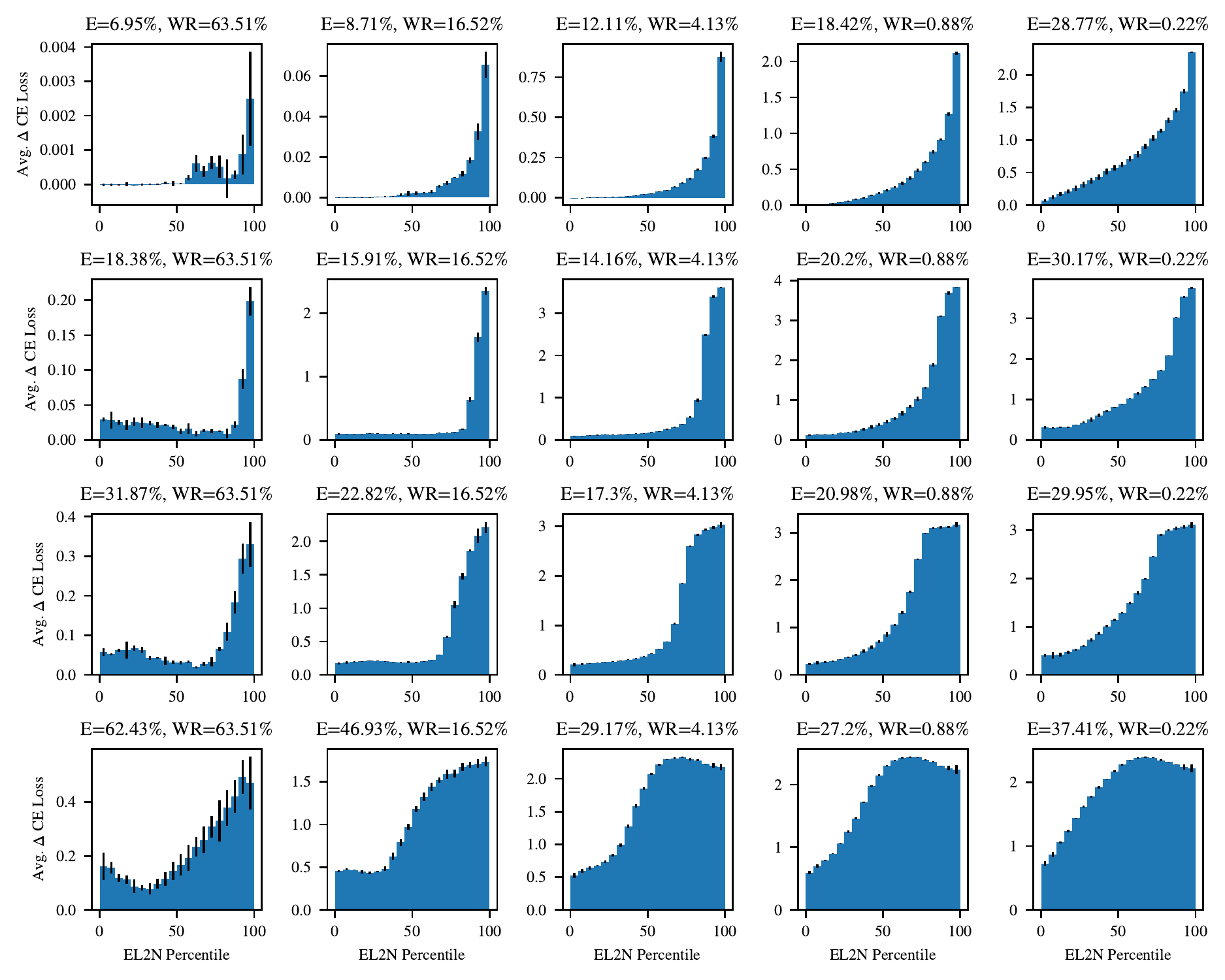}
}

\subfloat[CIFAR-100-ResNet32, rows correspond to 0\%, 15\%, 30\% and 60\% random label noise]{
	\includegraphics[scale=0.65]{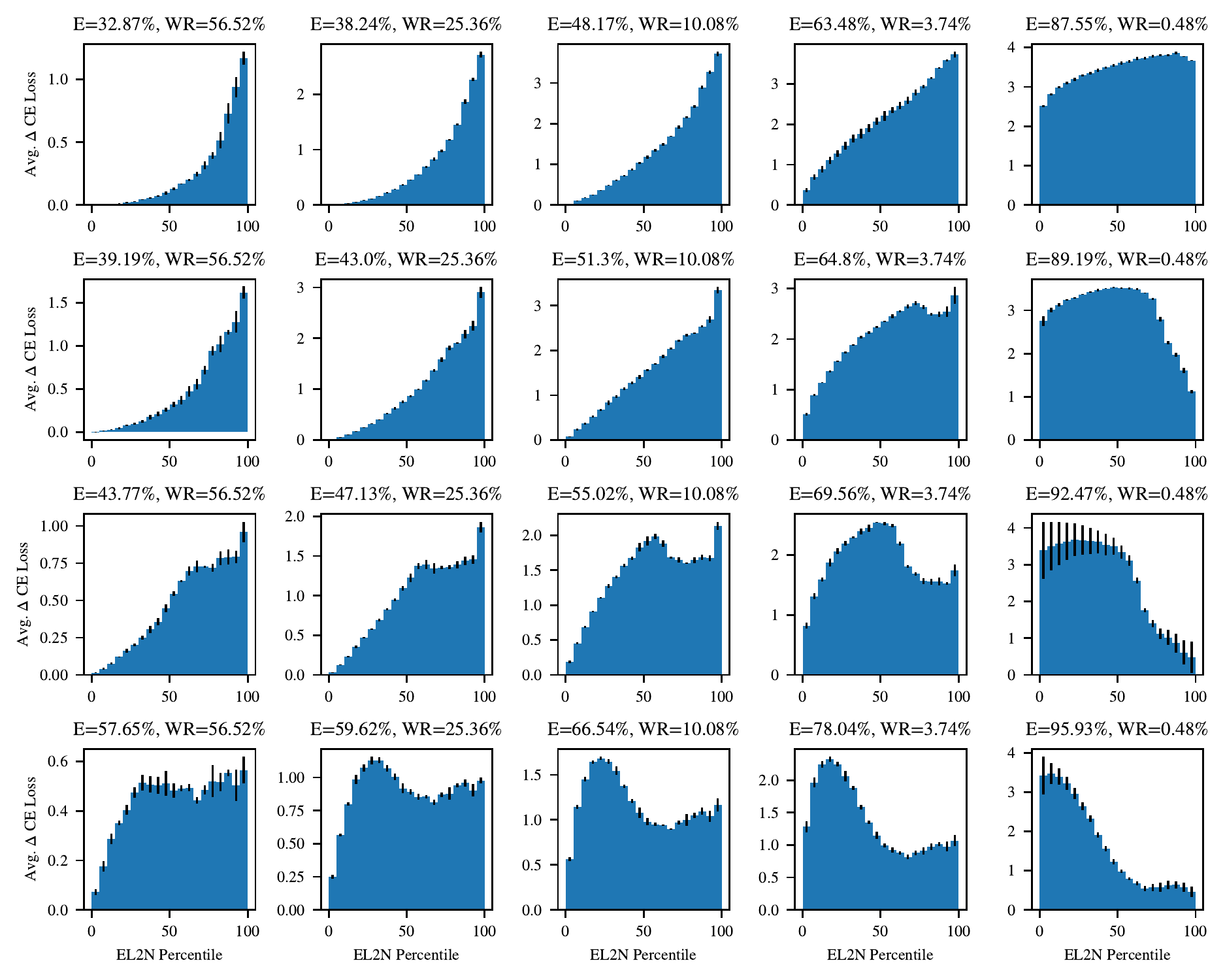}
}

\caption{(Cont.) Width downscaling increases subgroup training loss.
Title shows the test error (E) and weights remaining (WR), as a portion of the number of weights in the model with the original and unscaled width.
A positive value indicates that width downscaling increases training loss.}
\label{fig:width-downscaling-effect-on-model-fit-many-sparsity-levels-noisy}
\end{figure*}

\begin{figure*}[h!]
\ContinuedFloat
\subfloat[ImageNet-ResNet50, with 0\%, 15\% random label noise]{
	\includegraphics[scale=0.65]{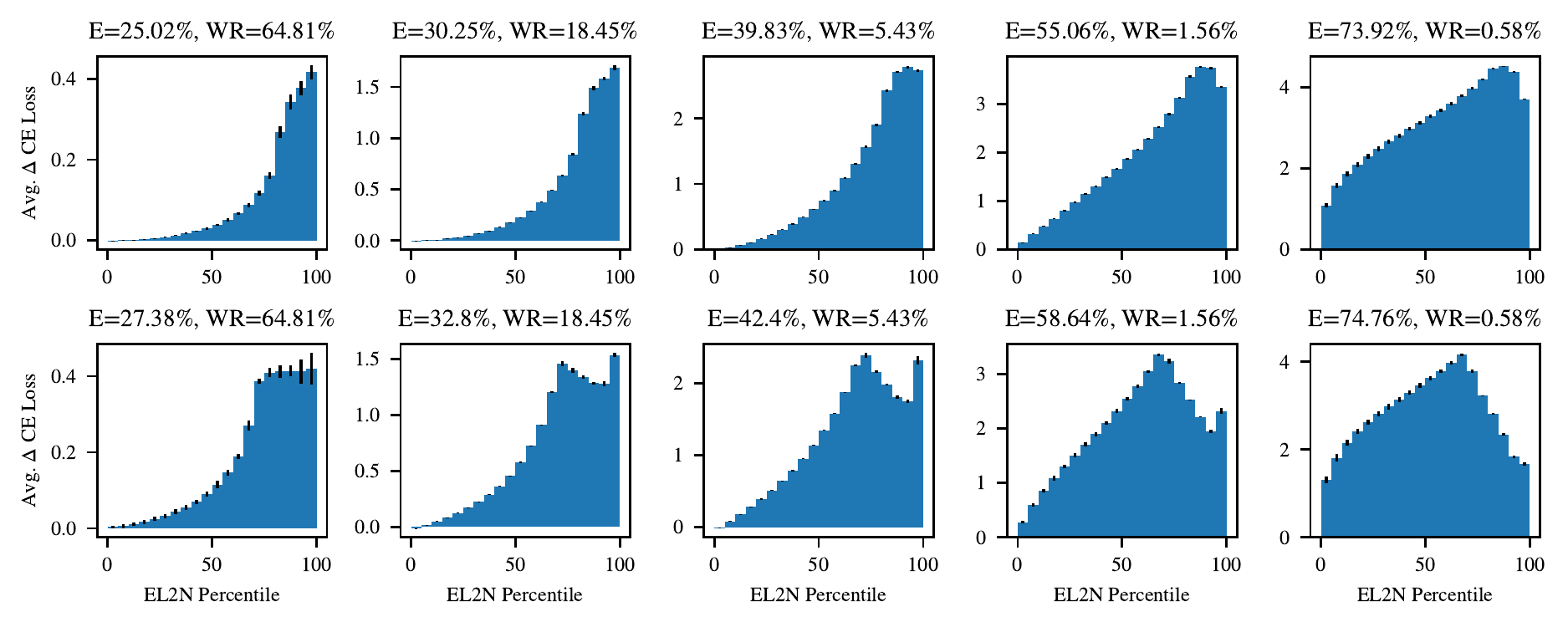}
}
\caption{(Cont.) Width downscaling increases subgroup training loss.
Title shows the test error (E) and weights remaining (WR), as a portion of the number of weights in the model with the original and unscaled width.
A positive value indicates that width downscaling increases training loss.}
\label{fig:width-downscaling-effect-on-model-fit-many-sparsity-levels-noisy}
\end{figure*}

\clearpage

\section{Data for Experiments in Paper}
\label{appx:raw-data}

\begin{table}[]
\centering
\begin{tabular}{@{}cccc@{}}
\toprule
Models                         & Method & Noise Level    & Test Errors                            \\ \midrule
\multirow{2}{*}{M-LeNet}       & LR     & 5\%/10\%/15\%  & 5.21±0.08\%/7.57±0.35\%/9.96±0.62\%    \\
                               & EDT    & 5\%/10\%/15\%  & 4.84±0.33\%/8.96±0.13\%/13.04±0.71\%   \\ \midrule
\multirow{2}{*}{C10-ResNet20}  & LR     & 15\%/30\%/60\% & 13.04±0.43\%/16.03±0.13\%/25.39±0.11\% \\
                               & EDT    & 15\%/30\%/60\% & 13.45±0.16\%/18.40±0.24\%/39.87±0.58\% \\ \midrule
\multirow{2}{*}{C10-VGG-16}     & LR     & 15\%/30\%/60\% & 12.47±0.29\%/15.47±0.13\%/25.29±0.49\% \\
                               & EDT    & 15\%/30\%/60\% & 16.04±0.22\%/28.22±0.37\%/59.46±0.53\% \\ \midrule
\multirow{2}{*}{C100-ResNet32} & LR     & 15\%/30\%/60\% & 38.81±0.70\%/43.86±0.29\%/55.83±0.45\% \\
                               & EDT    & 15\%/30\%/60\% & 39.51±0.40\%/44.10±0.49\%/60.62±0.74\% \\ \midrule
\multirow{2}{*}{I-ResNet50}    & LR     & 15\%           & 26.15±0.09\%                           \\
                               & EDT    & 15\%           & 25.95±0.06\%                           \\ \bottomrule
\end{tabular}
\caption{Comparing Learning Rate Rewinding with Extended Dense Training. LR=Learning Rate Rewinding, EDT=Extended Dense Training.}
\label{tbl:pruning-vs-dense-training}
\end{table}

\begin{table}[h!]
\centering
\begin{tabular}{@{}ccccc@{}}
\toprule
Model                          & Noise Level                              & Rewind   & Type & Test Error                                  \\ \midrule
\multirow{3}{*}{M-LeNet}       & \multirow{3}{*}{0\%/5\%/10\%/15\%}  & Weight & Sparse      & 1.5±0.1\%/3.7±0.3\%/4.2±0.1\%/4.4±0.2\%     \\
                               &                                     & LR     & Sparse      & 1.8±0.1\%/5.2±0.1\%/7.6±0.3\%/10.0±0.6\%    \\
                               &                                     & N/A    & Dense       & 2.1±0.2\%/5.3±0.2\%/9.1±0.2\%/13.4±0.3\%    \\ \midrule
\multirow{3}{*}{C10-ResNet20}  & \multirow{3}{*}{0\%/15\%/30\%/60\%} & Weight & Sparse      & 8.1±0.3\%/13.1±0.2\%/15.4±0.2\%/24.5±0.5\%  \\
                               &                                     & LR     & Sparse      & 7.7±0.1\%/13.0±0.4\%/16.0±0.1\%/25.4±0.1\%  \\
                               &                                     & N/A    & Dense       & 8.2±0.1\%/13.6±0.3\%/18.4±0.4\%/40.3±0.4\%  \\ \midrule
\multirow{3}{*}{C10-VGG-16}     & \multirow{3}{*}{0\%/15\%/30\%/60\%} & Weight & Sparse      & 6.2±0.1\%/12.0±0.3\%/14.8±0.4\%/23.8±0.2\%  \\
                               &                                     & LR     & Sparse      & 5.9±0.1\%/12.5±0.3\%/15.5±0.1\%/25.5±0.3\%  \\
                               &                                     & N/A    & Dense       & 6.6±0.0\%/17.5±0.2\%/31.5±0.6\%/62.7±0.6\%  \\ \midrule
\multirow{3}{*}{C100-ResNet32} & \multirow{3}{*}{0\%/15\%/30\%/60\%} & Weight & Sparse      & 30.6±0.4\%/38.7±0.6\%/42.9±0.1\%/54.7±0.5\% \\
                               &                                     & LR     & Sparse      & 29.9±0.3\%/38.8±0.7\%/43.9±0.3\%/55.8±0.5\% \\
                               &                                     & N/A    & Dense       & 30.3±0.2\%/38.9±0.3\%/44.0±0.6\%/61.0±0.4\% \\ \bottomrule
\end{tabular}
\caption{Pruning's Impact on Model Generalization. Shorthand for dataset names: M=MNIST, C10=CIFAR-10, C100=CIFAR-100, I=ImageNet.}
\label{tbl:pruning-impact-on-model-generalization}
\end{table}

\begin{table}[]
\centering
\begin{tabular}{@{}cccc@{}}
\toprule
Model                          & Method & Noise Level        & Test Error                                          \\ \midrule
\multirow{2}{*}{M-LeNet}       & WD     & 0\%/5\%/10\%/15\%  & 1.90±0.09\%/5.63±0.18\%/5.45±0.38\%/6.08±0.33\%     \\
                               & LR     & 0\%/5\%/10\%/15\%  & 1.82±0.11\%/5.21±0.08\%/7.57±0.35\%/9.96±0.62\%     \\ \midrule
\multirow{2}{*}{C10-ResNet20}  & WD     & 0\%/15\%/30\%/60\% & 8.42±0.11\%/13.06±0.16\%/16.01±0.55\%/25.29±0.59\%  \\
                               & LR     & 0\%/15\%/30\%/60\% & 7.71±0.09\%/13.04±0.43\%/16.03±0.13\%/25.39±0.11\%  \\ \midrule
\multirow{2}{*}{C10-VGG-16}     & WD     & 0\%/15\%/30\%/60\% & 6.67±0.04\%/14.32±0.24\%/17.43±0.10\%/26.76±0.37\%  \\
                               & LR     & 0\%/15\%/30\%/60\% & 5.86±0.08\%/12.47±0.29\%/15.47±0.13\%/25.29±0.49\%  \\ \midrule
\multirow{2}{*}{C100-ResNet32} & WD     & 0\%/15\%/30\%/60\% & 30.43±0.25\%/38.88±0.13\%/44.41±0.16\%/57.92±0.29\% \\
                               & LR     & 0\%/15\%/30\%/60\% & 29.93±0.34\%/38.81±0.70\%/43.86±0.29\%/55.83±0.45\% \\ \midrule
\multirow{2}{*}{I-ResNet50}    & WD     & 0\%/15\%           & 23.89±0.10\%/24.75±0.18\%                           \\
                               & LR     & 0\%/15\%           & 23.42±0.14\%/26.15±0.09\%                           \\ \bottomrule
\end{tabular}
\caption{
Comparing Pruning with Width Downscaling.
Pruning matches or exceeds the generalization performance of width downscaling, except on MNIST-LeNet benchmark.
Shorthand for dataset names: M=MNIST, C10=CIFAR-10, C100=CIFAR-100, I=ImageNet.
}

\label{tbl:cmp-pruning-with-width-scaling}
\end{table}

\begin{table}[]
\centering
\small
\begin{tabular}{@{}ccccc@{}}
\toprule
Model                                                                        & Method & Noise Level    & Test Error  (Optimal)                  & Test Error (0\% Sparsity)              \\ \midrule
\multirow{2}{*}{\begin{tabular}[c]{@{}c@{}}MNIST\\ LeNet\end{tabular}}       & LR     & 5\%/10\%/15\%  & 5.22±0.06\%/7.57±0.35\%/10.03±0.60\%   & 5.35±0.18\%/9.08±0.21\%/13.40±0.32\%   \\
                                                                             & DS     & 5\%/10\%/15\%  & 2.55±0.12\%/2.74±0.13\%/3.30±0.11\%    & 5.23±0.35\%/9.11±0.09\%/12.03±0.93\%   \\ \midrule
\multirow{2}{*}{\begin{tabular}[c]{@{}c@{}}CIFAR-10\\ ResNet20\end{tabular}}  & LR     & 15\%/30\%/60\% & 12.63±0.12\%/15.88±0.26\%/25.39±0.11\% & 13.57±0.26\%/18.35±0.42\%/40.26±0.43\% \\
                                                                             & DS     & 15\%/30\%/60\% & 9.76±0.11\%/11.38±0.10\%/18.66±0.26\%  & 11.20±0.29\%/14.51±0.40\%/32.77±0.48\% \\ \midrule
\multirow{2}{*}{\begin{tabular}[c]{@{}c@{}}CIFAR-10\\ VGG-16\end{tabular}}     & LR     & 15\%/30\%/60\% & 12.39±0.27\%/15.19±0.26\%/25.49±0.29\% & 17.51±0.18\%/31.46±0.60\%/62.68±0.63\% \\
                                                                             & DS     & 15\%/30\%/60\% & 8.84±0.08\%/10.09±0.09\%/17.80±0.37\%  & 18.07±0.57\%/32.25±0.40\%/61.21±0.88\% \\ \midrule
\multirow{2}{*}{\begin{tabular}[c]{@{}c@{}}C100\\ ResNet32\end{tabular}}     & LR     & 15\%/30\%/60\% & 38.81±0.70\%/43.29±0.40\%/55.81±0.46\% & 38.85±0.29\%/43.97±0.61\%/61.03±0.36\% \\
                                                                             & DS     & 15\%/30\%/60\% & 34.24±0.25\%/37.17±0.23\%/47.38±1.10\% & 34.35±0.26\%/37.89±0.33\%/52.66±0.42\% \\ \midrule
\multirow{2}{*}{\begin{tabular}[c]{@{}c@{}}ImageNet\\ ResNet50\end{tabular}} & LR     & 15\%           & 26.12±0.08\%                           & 26.44±0.06\%                           \\
                                                                             & DS     & 15\%           & 24.75±0.09\%                           & 24.77±0.07\%                           \\ \bottomrule
\end{tabular}
\caption{
Comparing Pruning with Training Dense Models Exclusively on Dataset Subsets Predicted Correctly by Sparse Models (referred to as ``Dense Subset" models).
``Dense Subset" models matches or exceeds the generalization performance achieved by pruning.
Shorthand for dataset names: LR = Learning Rate Rewinding, DS = ``Dense Subset".
}
\label{tbl:cmp-pruning-with-dense-subset}
\end{table}

\begin{table}[h!]
\centering
\begin{tabular}{@{}cccc@{}}
\toprule
Model                          & Method & Noise Level        & Test Error                                          \\ \midrule
\multirow{2}{*}{M-LeNet}       & WR     & 0\%/5\%/10\%/15\%  & 1.50±0.08\%/3.74±0.30\%/4.16±0.07\%/4.41±0.16\%     \\
                               & LR     & 0\%/5\%/10\%/15\%  & 1.82±0.11\%/5.21±0.08\%/7.57±0.35\%/9.96±0.62\%     \\ \midrule
\multirow{2}{*}{C10-ResNet20}  & WR     & 0\%/15\%/30\%/60\% & 8.07±0.29\%/13.09±0.24\%/15.43±0.25\%/24.47±0.49\%  \\
                               & LR     & 0\%/15\%/30\%/60\% & 7.71±0.09\%/13.04±0.43\%/16.03±0.13\%/25.39±0.11\%  \\ \midrule
\multirow{2}{*}{C10-VGG-16}     & WR     & 0\%/15\%/30\%/60\% & 6.19±0.10\%/11.97±0.33\%/14.81±0.39\%/23.80±0.15\%  \\
                               & LR     & 0\%/15\%/30\%/60\% & 5.86±0.08\%/12.47±0.29\%/15.47±0.13\%/27.90±1.41\%  \\ \midrule
\multirow{2}{*}{C100-ResNet32} & WR     & 0\%/15\%/30\%/60\% & 30.58±0.36\%/38.74±0.59\%/42.89±0.09\%/54.67±0.50\% \\
                               & LR     & 0\%/15\%/30\%/60\% & 29.93±0.34\%/38.81±0.70\%/43.86±0.29\%/55.83±0.45\% \\ \midrule
\multirow{2}{*}{I-ResNet50}    & WR     & 0\%/15\%           & 23.51±0.08\%/26.14±0.23\%                           \\
                               & LR     & 0\%/15\%           & 23.42±0.14\%/26.15±0.09\%                           \\ \bottomrule
\end{tabular}
\caption{The Effect of Weight Resetting on Pruning's Generalization Improvement. Shorthand for dataset names: M=MNIST, C10=CIFAR-10, C100=CIFAR-100, I=ImageNet.}
\label{tbl:effect-of-weight-resetting}
\end{table}

\begin{table}[]
\centering
\begin{tabular}{@{}cccc@{}}
\toprule
Model                          & Method    & Noise Level        & Test Error                                          \\ \midrule
\multirow{3}{*}{M-LeNet}       & Magnitude & 0\%/5\%/10\%/15\%  & 1.82±0.11\%/5.21±0.08\%/7.57±0.35\%/9.96±0.62\%     \\
                               & SynFlow   & 0\%/5\%/10\%/15\%  & 1.65±0.05\%/5.20±0.15\%/9.59±0.20\%/11.72±0.64\%    \\
                               & Random    & 0\%/5\%/10\%/15\%  & 1.86±0.19\%/5.11±0.27\%/6.88±0.23\%/7.50±0.07\%     \\ \midrule
\multirow{3}{*}{C10-ResNet20}  & Magnitude & 0\%/15\%/30\%/60\% & 7.71±0.09\%/13.04±0.43\%/16.03±0.13\%/25.39±0.11\%  \\
                               & SynFlow   & 0\%/15\%/30\%/60\% & 7.83±0.28\%/12.61±0.41\%/15.27±0.02\%/25.09±0.20\%  \\
                               & Random    & 0\%/15\%/30\%/60\% & 8.14±0.29\%/12.09±0.08\%/14.53±0.09\%/24.52±0.42\%  \\ \midrule
\multirow{3}{*}{C10-VGG-16}     & Magnitude & 0\%/15\%/30\%/60\% & 5.86±0.08\%/12.47±0.29\%/15.47±0.13\%/25.29±0.49\%  \\
                               & SynFlow   & 0\%/15\%/30\%/60\% & 6.14±0.09\%/12.35±0.18\%/15.04±0.21\%/30.67±10.03\% \\
                               & Random    & 0\%/15\%/30\%/60\% & 6.40±0.17\%/14.65±0.39\%/17.77±0.42\%/27.30±0.49\%  \\ \midrule
\multirow{3}{*}{C100-ResNet32} & Magnitude & 0\%/15\%/30\%/60\% & 29.93±0.34\%/38.81±0.70\%/43.86±0.29\%/55.83±0.45\% \\
                               & SynFlow   & 0\%/15\%/30\%/60\% & 30.24±0.24\%/38.13±0.33\%/43.12±0.57\%/55.55±0.17\% \\
                               & Random    & 0\%/15\%/30\%/60\% & 30.15±0.14\%/37.18±0.32\%/41.44±0.30\%/53.60±0.07\% \\ \midrule
\multirow{3}{*}{I-ResNet50}    & Magnitude & 0\%/15\%           & 23.42±0.14\%/26.15±0.09\%                           \\
                               & SynFlow   & 0\%/15\%           & 23.90±0.08\%/26.37±0.10\%                           \\
                               & Random    & 0\%/15\%           & 23.86±0.10\%/26.49±0.10\%                           \\ \bottomrule
\end{tabular}
\caption{The Effect of Weight Selection Heuristics on Pruning's Generalization Improvement. Shorthand for dataset names: M=MNIST, C10=CIFAR-10, C100=CIFAR-100, I=ImageNet.}
\label{tbl:weight-selection-heuristics}
\end{table}

\begin{table}[]
\centering
\begin{tabular}{@{}cccc@{}}
\toprule
Benchmark                      & Noise Level           & Partition & Dense/Sparse (Diff) CE Loss \\ \midrule
\multirow{6}{*}{M-LeNet}       & \multirow{2}{*}{5\%}  & Noisy     & 0.46±0.06/0.2±0.13(-0.26)  \\
                               &                       & Original     & 0.02±0.0/0.01±0.0(-0.01)   \\ \cmidrule(l){2-4}
                               & \multirow{2}{*}{10\%} & Noisy     & 0.53±0.04/3.83±0.06(+3.3)   \\
                               &                       & Original     & 0.03±0.01/0.24±0.01(+0.21)  \\ \cmidrule(l){2-4}
                               & \multirow{2}{*}{15\%} & Noisy     & 0.59±0.0/3.5±0.17(+2.91)    \\
                               &                       & Original     & 0.05±0.01/0.36±0.04(+0.31)  \\ \midrule
\multirow{6}{*}{C10-ResNet20}  & \multirow{2}{*}{15\%} & Noisy     & 2.63±0.06/3.18±0.27(+0.55)  \\
                               &                       & Original     & 0.18±0.0/0.26±0.03(+0.08)   \\ \cmidrule(l){2-4}
                               & \multirow{2}{*}{30\%} & Noisy     & 2.52±0.01/3.1±0.03(+0.58)   \\
                               &                       & Original     & 0.34±0.0/0.55±0.02(+0.21)   \\ \cmidrule(l){2-4}
                               & \multirow{2}{*}{60\%} & Noisy     & 2.21±0.01/2.61±0.0(+0.4)    \\
                               &                       & Original     & 0.77±0.01/1.16±0.0(+0.39)   \\ \midrule
\multirow{6}{*}{C10-VGG-16}     & \multirow{2}{*}{15\%} & Noisy     & 0.06±0.01/3.28±0.39(+3.22)  \\
                               &                       & Original     & 0.03±0.01/0.31±0.02(+0.28)  \\ \cmidrule(l){2-4}
                               & \multirow{2}{*}{30\%} & Noisy     & 0.12±0.01/2.99±0.1(+2.87)   \\
                               &                       & Original     & 0.07±0.0/0.53±0.01(+0.46)   \\ \cmidrule(l){2-4}
                               & \multirow{2}{*}{60\%} & Noisy     & 0.27±0.01/2.55±0.01(+2.28)  \\
                               &                       & Original     & 0.14±0.01/1.17±0.04(+1.03)  \\ \midrule
\multirow{6}{*}{C100-ResNet32} & \multirow{2}{*}{15\%} & Noisy     & 4.19±0.03/5.71±0.04(+1.52)  \\
                               &                       & Original     & 0.43±0.0/0.81±0.01(+0.38)   \\ \cmidrule(l){2-4}
                               & \multirow{2}{*}{30\%} & Noisy     & 4.25±0.04/5.13±0.34(+0.88)  \\
                               &                       & Original     & 0.66±0.01/1.05±0.16(+0.39)  \\ \cmidrule(l){2-4}
                               & \multirow{2}{*}{60\%} & Noisy     & 4.02±0.02/4.94±0.04(+0.92)  \\
                               &                       & Original     & 1.24±0.02/2.13±0.03(+0.89)  \\ \midrule
\multirow{2}{*}{I-ResNet50}    & \multirow{2}{*}{15\%} & Noisy     & 7.53±0.03/7.34±0.04(-0.19) \\
                               &                       & Original     & 0.56±0.0/0.53±0.0(-0.03)   \\ \bottomrule
\end{tabular}
\caption{Pruning's Impact on Training Loss Incurred on Noisy/Original partitions of Dataset.
Shorthand for dataset names: M=MNIST, C10=CIFAR-10, C100=CIFAR-100, I=ImageNet.}
\label{tbl:pruning-impact-on-noisy-clean-memorization}
\end{table}

\clearpage

\end{document}